# Understanding Generalization and Optimization Performance of Deep CNNs

**Pan Zhou** [1] **Jiashi Feng** [1]

## Abstract

This work aims to provide understandings on the remarkable success of deep convolutional neural networks (CNNs) by theoretically analyzing their generalization performance and establishing optimization guarantees for gradient descent based training algorithms. Specifically, for a CNN model consisting of $l$ convolutional layers and one fully connected layer, we prove that its generalization error is bounded by $\mathcal{O}(\sqrt{\theta\widetilde{\varrho}/n})$ where $\theta$ denotes freedom degree of the network parameters and $\widetilde{\varrho} = \mathcal{O}(\log(\prod_{i=1}^{l} b_i(k_i - s_i + 1)/p) + \log(b_{l+1}))$ encapsulates architecture parameters including the kernel size $k_i$, stride $s_i$, pooling size $p$ and parameter magnitude $b_i$. To our best knowledge, this is the first generalization bound that only depends on $\mathcal{O}(\log(\prod_{i=1}^{l+1} b_i))$, tighter than existing ones that all involve an exponential term like $\mathcal{O}(\prod_{i=1}^{l+1} b_i)$. Besides, we prove that for an arbitrary gradient descent algorithm, the computed approximate stationary point by minimizing empirical risk is also an approximate stationary point to the population risk. This well explains why gradient descent training algorithms usually perform sufficiently well in practice. Furthermore, we prove the one-to-one correspondence and convergence guarantees for the non-degenerate stationary points between the empirical and population risks. It implies that the computed local minimum for the empirical risk is also close to a local minimum for the population risk, thus ensuring the good generalization performance of CNNs.

## 1. Introduction

Deep convolutional neural networks (CNNs) have been successfully applied to various fields, such as image classification (Szegedy et al., 2015; He et al., 2016; Wang et al., 2017), speech recognition (Sainath et al., 2013; Abdel-Hamid et al., 2014), and classic games (Silver et al., 2016; Brown & Sandholm, 2017). However, theoretical analyses and understandings on deep CNNs still largely lag their practical applications. Recently, although many works establish theoretical understandings on deep feedforward neural networks (DNNs) from various aspects, *e.g.* (Neyshabur et al., 2015; Kawaguchi, 2016; Zhou & Feng, 2018; Tian, 2017; Lee et al., 2017), only a few (Sun et al., 2016; Kawaguchi et al., 2017; Du et al., 2017a;b) provide explanations on deep CNNs due to their more complex architectures and operations. Besides, these existing works all suffer certain discrepancy between their theories and practice. For example, the developed generalization error bound either exponentially grows along with the depth of a CNN model (Sun et al., 2016) or is data-dependent (Kawaguchi et al., 2017), and the convergence guarantees for optimization algorithms over CNNs are achieved by assuming an over-simplified CNN model consisting of *only one* non-overlapping convolutional layer (Du et al., 2017a;b).

As an attempt to explain the practical success of deep CNNs and mitigate the gap between theory and practice, this work aims to provide tighter data-independent generalization error bound and algorithmic optimization guarantees for the commonly used deep CNN models in practice. Specifically, we theoretically analyze the deep CNNs from following two aspects: (1) how their generalization performance varies with different network architecture choices and (2) why gradient descent based algorithms such as stochastic gradient descend (SGD) (Robbins & Monro, 1951), adam (Kingma & Ba, 2015) and RMSProp (Tieleman & Hinton, 2012), on minimizing empirical risk usually offer models with satisfactory performance. Moreover, we theoretically demonstrate the benefits of (stride) convolution and pooling operations, which are unique for CNNs, to the generalization performance, compared with feedforward networks.

Formally, we consider a CNN model $g(\boldsymbol{w}; \boldsymbol{D})$ parameterized by $\boldsymbol{w} \in \mathbb{R}^d$, consisting of $l$ convolutional layers and one subsequent fully connected layer. It maps the input $\boldsymbol{D} \in \mathbb{R}^{r_0 \times c_0}$ to an output vector $\boldsymbol{v} \in \mathbb{R}^{d_{l+1}}$. Its $i$-th convolutional layer takes $\boldsymbol{Z}_{(i-1)} \in \mathbb{R}^{\tilde{r}_{i-1} \times \tilde{c}_{i-1} \times d_{i-1}}$ as input and outputs $\boldsymbol{Z}_{(i)} \in \mathbb{R}^{\tilde{r}_i \times \tilde{c}_i \times d_i}$ through spatial convolution, non-linear activation and pooling operations sequentially.

[1]Department of Electrical & Computer Engineering (ECE), National University of Singapore, Singapore. Correspondence to: Pan Zhou <pzhou@u.nus.edu>, Jiashi Feng <elefjia@nus.edu.sg>.

Here $\tilde{r}_i \times \tilde{c}_i$ and $d_i$ respectively denote resolution and the number of feature maps. Specifically, the computation with the $i$-th convolutional layer is described as

$$\boldsymbol{X}_{(i)}(:,:,j) = \boldsymbol{Z}_{(i-1)} \circledast \boldsymbol{W}_{(i)}^{j} \in \mathbb{R}^{r_i \times c_i}, \forall j = 1, \cdots, d_i,$$
$$\boldsymbol{Y}_{(i)} = \sigma_1(\boldsymbol{X}_{(i)}) \in \mathbb{R}^{r_i \times c_i \times d_i},$$
$$\boldsymbol{Z}_{(i)} = \texttt{pool}\left(\boldsymbol{Y}_{(i)}\right) \in \mathbb{R}^{\tilde{r}_i \times \tilde{c}_i \times d_i},$$

where $\boldsymbol{X}_{(i)}(:,:,j)$ denotes the $j$-th feature map output by the $i$-th layer; $\boldsymbol{W}_{(i)}^{j} \in \mathbb{R}^{k_i \times k_i \times d_{i-1}}$ denotes the $j$-th convolutional kernel of size $k_i \times k_i$ and there are in total $d_i$ kernels in the $i$-th layer; $\circledast$, $\texttt{pool}(\cdot)$ and $\sigma_1(\cdot)$ respectively denote the convolutional operation with stride $s_i$, pooling operation with window size $p \times p$ without overlap and the sigmoid function. In particular, $\boldsymbol{Z}_{(0)} = \boldsymbol{D}$ is the input sample. The last layer is a fully connected one and formulated as

$$\boldsymbol{u} = \boldsymbol{W}_{(l+1)} \boldsymbol{z}_{(l)} \in \mathbb{R}^{d_{l+1}} \quad \text{and} \quad \boldsymbol{v} = \sigma_2(\boldsymbol{u}) \in \mathbb{R}^{d_{l+1}},$$

where $\boldsymbol{z}_{(l)} \in \mathbb{R}^{\tilde{r}_l \tilde{c}_l d_l}$ is vectorization of the output $\boldsymbol{Z}_{(l)}$ of the last convolutional layer; $\boldsymbol{W}_{(l+1)} \in \mathbb{R}^{d_{l+1} \times \tilde{r}_l \tilde{c}_l d_l}$ denotes the connection weight matrix; $\sigma_2(\cdot)$ is a softmax activation function (for classification) and $d_{l+1}$ is the class number.

In practice, a deep CNN model is trained by minimizing the following empirical risk in terms of squared loss on the training data pairs $(\boldsymbol{D}^{(i)}, \boldsymbol{y}^{(i)})$ drawn from an unknown distribution $\boldsymbol{\mathcal{D}}$,

$$\widetilde{\boldsymbol{Q}}_n(\boldsymbol{w}) \triangleq \frac{1}{n} \sum_{i=1}^{n} f(g(\boldsymbol{w}; \boldsymbol{D}^{(i)}), \boldsymbol{y}^{(i)}), \tag{1}$$

where $f(g(\boldsymbol{w}; \boldsymbol{D}), \boldsymbol{y}) = \frac{1}{2}\|\boldsymbol{v} - \boldsymbol{y}\|_2^2$ is the squared loss function. One can obtain the model parameter $\widetilde{\boldsymbol{w}}$ via SGD or its variants like adam and RMSProp. However, this empirical solution is different from the desired optimum $\boldsymbol{w}^*$ that minimizes the population risk:

$$\boldsymbol{Q}(\boldsymbol{w}) \triangleq \mathbb{E}_{(\boldsymbol{D}, \boldsymbol{y}) \sim \boldsymbol{\mathcal{D}}}\, f(g(\boldsymbol{w}; \boldsymbol{D}), \boldsymbol{y}).$$

This raises an important question: why CNNs trained by minimizing the empirical risk usually perform well in practice, considering the high model complexity and nonconvexity? This work answers this question by (1) establishing the generalization performance guarantee for CNNs and (2) expounding why the computed solution $\widetilde{\boldsymbol{w}}$ by gradient descent based algorithms for minimizing the empirical risk usually performs sufficiently well in practice.

To be specific, we present three new theoretical guarantees for CNNs. First, we prove that the generalization error of deep CNNs decreases at the rate of $\mathcal{O}(\sqrt{\theta \widetilde{\varrho}/(2n)})$ where $\theta$ denotes parameter freedom degree[1], and $\widetilde{\varrho}$ depends on the network architecture parameters including the convolutional kernel size $k_i$, stride $s_i$, pooling size $p$, channel number $d_i$ and parameter magnitudes. It is worth mentioning that our generalization error bound is the first one that does not exponentially grow with depth.

Secondly, we prove that for any gradient descent based optimization algorithm, *e.g.* SGD, RMSProp or adam, if its output $\widetilde{\boldsymbol{w}}$ is an approximate stationary point of the empirical risk $\widetilde{\boldsymbol{Q}}_n(\boldsymbol{w})$, $\widetilde{\boldsymbol{w}}$ is also an approximate stationary point of the population risk $\boldsymbol{Q}(\boldsymbol{w})$. This result is important as it explains why CNNs trained by minimizing the empirical risk have good generalization performance on test samples. We achieve such results by analyzing the convergence behavior of the empirical gradient to its population counterpart.

Finally, we go further and quantitatively bound the distance between $\widetilde{\boldsymbol{w}}$ and $\boldsymbol{w}^*$. We prove that when the samples are sufficient, a non-degenerate stationary point $\boldsymbol{w}_n$ of $\widetilde{\boldsymbol{Q}}_n(\boldsymbol{w})$ uniquely corresponds to a non-degenerate stationary point $\boldsymbol{w}^*$ of the population risk $\boldsymbol{Q}(\boldsymbol{w})$, with a distance shrinking at the rate of $\mathcal{O}((\beta/\zeta)\sqrt{d\widetilde{\varrho}/n})$ where $\beta$ also depends on the CNN architecture parameters (see Thereom 2). Here $\zeta$ accounts for the geometric topology of non-degenerate stationary points. Besides, the corresponding pair $(\boldsymbol{w}_n, \boldsymbol{w}^*)$ shares the same geometrical property—if one in $(\boldsymbol{w}_n, \boldsymbol{w}^*)$ is a local minimum or saddle point, so is the other one. All these results guarantee that for an arbitrary algorithm provided with sufficient samples, if the computed $\widetilde{\boldsymbol{w}}$ is close to the stationary point $\boldsymbol{w}_n$, then $\widetilde{\boldsymbol{w}}$ is also close to the optimum $\boldsymbol{w}^*$ and they share the same geometrical property.

To sum up, we make multiple contributions to understand deep CNNs theoretically. To our best knowledge, this work presents the first theoretical guarantees on both generalization error bound without exponential growth over network depth and optimization performance for *deep* CNNs. We substantially extend prior works on CNNs (Du et al., 2017a;b) from the over-simplified single-layer models to the multi-layer ones, which is of more practical significance. Our generalization error bound is much tighter than the one derived from Rademacher complexity (Sun et al., 2016) and is also independent of data and specific training procedure, which distinguishes it from (Kawaguchi et al., 2017).

## 2. Related Works

Recently, many works have been devoted to explaining the remarkable success of deep neural networks. However, most works only focus on analyzing fully feedforward networks from aspects like generalization performance (Bartlett & Maass, 2003; Neyshabur et al., 2015), loss surface (Saxe et al., 2014; Dauphin et al., 2014; Choromanska et al., 2015; Kawaguchi, 2016; Nguyen & Hein, 2017; Zhou & Feng, 2018), optimization algorithm convergence (Tian, 2017; Li

[1] We use the terminology of "parameter freedom degree" here for characterizing redundancy of parameters. For example, for a rank-$r$ matrix $A \in \mathbb{R}^{m_1 \times m_2}$, the parameter freedom degree in this work is $r(m_1 + m_2 + 1)$ instead of the commonly used one $r(m_1 + m_2 - r)$.

& Yuan, 2017) and expression ability (Eldan & Shamir, 2016; Soudry & Hoffer, 2017; Lee et al., 2017).

The literature targeting at analyzing CNNs is very limited, mainly because CNNs have much more complex architectures and computation. Among the few existing works, Du et al. (2017b) presented results for a simple and shallow CNN consisting of only one non-overlapping convolutional layer and ReLU activations, showing that gradient descent (GD) algorithms with weight normalization can converge to the global minimum. Similarly, Du et al. (2017a) also analyzed optimization performance of GD and SGD with non-Gaussian inputs for CNNs with only one non-overlapping convolutional layer. By utilizing the kernel method, Zhang et al. (2017) transformed a CNN model into a single-layer convex model which has almost the same loss as the original CNN with high probability and proved that the transformed model has higher learning efficiency.

Regarding generalization performance of CNNs, Sun et al. (2016) provided the Rademacher complexity of a deep CNN model which is then used to establish the generalization error bound. But the Rademacher complexity exponentially depends on the magnitude of total parameters per layer, leading to loose results. In contrast, the generalization error bound established in this work is much tighter, as discussed in details in Sec. 3. Kawaguchi et al. (2017) introduced two generalization error bounds of CNN, but both depend on a specific dataset as they involve the validation error or the intermediate outputs for the network model on a provided dataset. They also presented dataset-independent generalization error bound, but with a specific two-phase training procedure required, where the second phase need fix the states of ReLU activation functions. However, such two-phase training procedure is not used in practice.

There are also some works focusing on convergence behavior of nonconvex empirical risk of a single-layer model to the population risk. Our proof techniques essentially differ from theirs. For example, (Gonen & Shalev-Shwartz, 2017) proved that the empirical risk converges to the population risk for those nonconvex problems with no degenerated saddle points. Unfortunately, due to existence of degenerated saddle points in deep networks (Dauphin et al., 2014; Kawaguchi, 2016), their results are not applicable here. A very recent work (Mei et al., 2017) focuses on single-layer nonconvex problems but requires the gradient and Hessian of the empirical risk to be strong sub-Gaussian and sub-exponential respectively. Besides, it assumes a linearity property for the gradient which hardly holds in practice. Comparatively, our assumptions are much milder. We only assume magnitude of the parameters to be bounded. Furthermore, we also explore the parameter structures of optimized CNNs, *i.e.* the low-rankness property, and derive bounds matching empirical observations better. Finally, we analyze the convergence rate of the empirical risk and generalization error of CNN which is absent in (Mei et al., 2017).

Our work is also critically different from the recent work (Zhou & Feng, 2018) although we adopt a similar analysis road map with it. Zhou & Feng (2018) analyzed DNNs while this work focuses on CNNs with more complex architectures and operations which are more challenging and requires different analysis techniques. Besides, this work provides stronger results in the sense of several tighter bounds with much milder assumptions. (1) For nonlinear DNNs, Zhou & Feng (2018) assumed the input data to be Gaussian, while this work gets rid of such a restrictive assumption. (2) The generalization error bound $\mathcal{O}(\widehat{r}^{l+1}\sqrt{d/n})$ in (Zhou & Feng, 2018) exponentially depends on the upper magnitude bound $\widehat{r}$ of the weight matrix per layer and linearly depends on the total parameter number $d$, while ours is $\mathcal{O}(\sqrt{\theta\widetilde{\varrho}/n})$ which only depends on the logarithm term $\widetilde{\varrho}=\log(\prod_{i=1}^{l+1} b_i)$ and the freedom degree $\theta$ of the network parameters, where $b_i$ and $b_{l+1}$ respectively denote the upper magnitude bounds of each kernel per layer and the weight matrix of the fully connected layer. Note, the exponential term $\mathcal{O}(\widehat{r}^{l+1})$ in (Zhou & Feng, 2018) cannot be further improved due to their proof techniques. The results on empirical gradient and stationary point pairs in (Zhou & Feng, 2018) rely on $\mathcal{O}(\widehat{r}^{2(l+1)})$, while ours is $\mathcal{O}(\prod_{i=1}^{l+1} b_i)$ which only depends on $b_i$ instead of $b_i^2$. (3) This work explores the parameter structures, *i.e.* the low-rankness property, and derives tighter bounds as the parameter freedom degree $\theta$ is usually smaller than the total parameter number $d$.

## 3. Generalization Performance of Deep CNNs

In this section, we present the generalization error bound for deep CNNs and reveal effects of different architecture parameters on their generalization performance, providing some principles on model architecture design. We derive these results by establishing *uniform* convergence of the empirical risk $\widetilde{\boldsymbol{Q}}_n(\boldsymbol{w})$ to its population one $\boldsymbol{Q}(\boldsymbol{w})$.

We start with explaining our assumptions. Similar to (Xu & Mannor, 2012; Tian, 2017; Zhou & Feng, 2018), we assume that the parameters of the CNN have bounded magnitude. But we get rid of the Gaussian assumptions on the input data, meaning our assumption is milder than the ones in (Tian, 2017; Soudry & Hoffer, 2017; Zhou & Feng, 2018).

**Assumption 1.** *The magnitudes of the $j$-th kernel $\boldsymbol{W}_{(i)}^{j}\in\mathbb{R}^{k_i\times k_i\times d_{i-1}}$ in the $i$-th convolutional layer and the weight matrix $\boldsymbol{W}_{(l+1)}\in\mathbb{R}^{d_{l+1}\times\tilde{r}_l\tilde{c}_l d_l}$ in the the fully connected layer are respectively bounded as follows*

$$\|\boldsymbol{W}_{(i)}^{j}\|_F\leq b_i\ (1\leq j\leq d_i;1\leq i\leq l),\quad \|\boldsymbol{W}_{(l+1)}\|_F\leq b_{l+1},$$

*where $b_i$ $(1\leq i\leq l)$ and $b_{l+1}$ are positive constants.*

We also assume that the entry value of the target output $\boldsymbol{y}$

always falls in $[0, 1]$, which can be achieved by scaling the entry value in $\boldsymbol{y}$ conveniently.

In this work, we also consider possible emerging structure of the learned parameters after training—the parameters usually present redundancy and low-rank structures (Lebedev et al., 2014; Jaderberg et al., 2014) due to high model complexity. So we incorporate low-rankness of the parameters or more concretely the parameter matrix consisting of kernels per layer, into our analysis. Denoting by $\mathsf{vec}(\boldsymbol{A})$ the vectorization of a matrix $\boldsymbol{A}$, we have Assumption 2.

**Assumption 2.** *Assume the matrices $\widetilde{\boldsymbol{W}}_{(i)}$ and $\boldsymbol{W}_{(l+1)}$ obey*

$$\mathsf{rank}(\widetilde{\boldsymbol{W}}_{(i)}) \leq a_i \ (1 \leq i \leq l) \ \text{and} \ \mathsf{rank}(\boldsymbol{W}_{(l+1)}) \leq a_{l+1},$$

*where $\widetilde{\boldsymbol{W}}_{(i)} = [\mathsf{vec}(\boldsymbol{W}_{(i)}^1), \mathsf{vec}(\boldsymbol{W}_{(i)}^2), \cdots, \mathsf{vec}(\boldsymbol{W}_{(i)}^{d_i})] \in \mathbb{R}^{k_i^2 d_{i-1} \times d_i}$ denotes the matrix consisting of all kernels in the $i$-th layer $(1 \leq i \leq l)$.*

The parameter low-rankness can also be defined on kernels individually by using the tensor rank (Tucker, 1966; Zhou et al., 2017; Zhou & Feng, 2017). Our proof techniques are extensible to this case and similar results can be expected.

### 3.1. Generalization Error Bound for Deep CNNs

We now proceed to establish generalization error bound for deep CNNs. Let $\boldsymbol{\mathcal{S}} = \{(\boldsymbol{D}^{(1)}, \boldsymbol{y}^{(1)}), \cdots, (\boldsymbol{D}^{(n)}, \boldsymbol{y}^{(n)})\}$ denote the set of training samples *i.i.d.* drawn from $\boldsymbol{\mathcal{D}}$. When the optimal solution $\widetilde{\boldsymbol{w}}$ to problem (1) is computed by a deterministic algorithm, the generalization error is defined as $\epsilon_g = \left|\widetilde{\boldsymbol{Q}}_n(\widetilde{\boldsymbol{w}}) - \boldsymbol{Q}(\widetilde{\boldsymbol{w}})\right|$. But in practice, a CNN model is usually optimized by randomized algorithms, *e.g.* SGD. So we adopt the following generalization error in expectation.

**Definition 1.** (Generalization error) *(Shalev-Shwartz et al., 2010) Assume a randomized algorithm $\mathcal{A}$ is employed for optimization over training samples $\boldsymbol{\mathcal{S}} = \{(\boldsymbol{D}^{(1)}, \boldsymbol{y}^{(1)}), \cdots, (\boldsymbol{D}^{(n)}, \boldsymbol{y}^{(n)})\} \sim \boldsymbol{\mathcal{D}}$ and $\widetilde{\boldsymbol{w}} = \mathrm{argmin}_{\boldsymbol{w}} \widetilde{\boldsymbol{Q}}_n(\boldsymbol{w})$ is the empirical risk minimizer (ERM). Then if we have $\mathbb{E}_{\boldsymbol{\mathcal{S}}\sim\boldsymbol{\mathcal{D}}}\left|\mathbb{E}_{\mathcal{A}}(\boldsymbol{Q}(\widetilde{\boldsymbol{w}}) - \widetilde{\boldsymbol{Q}}_n(\widetilde{\boldsymbol{w}}))\right| \leq \epsilon_k$, the ERM is said to have generalization error with rate $\epsilon_k$ under distribution $\boldsymbol{\mathcal{D}}$.*

We bound the generalization error in expectation for deep CNNs by first establishing uniform convergence of the empirical risk to its corresponding population risk, as stated in Lemma 1 with proof in Sec. D.1 in supplement.

**Lemma 1.** *Assume that in CNNs, $\sigma_1$ and $\sigma_2$ are respectively the sigmoid and softmax activation functions and the loss function $f(g(\boldsymbol{w}; \boldsymbol{D}), \boldsymbol{y})$ is squared loss. Under Assumptions 1 and 2, if $n \geq c_{f'} l^2 (b_{l+1} + \sum_{i=1}^{l} d_i b_i)^2 \max_i \sqrt{r_i c_i}/(\theta\varrho\varepsilon^2)$ where $c_{f'}$ is a universal constant, then with probability at least $1-\varepsilon$, we have*

$$\sup_{\boldsymbol{w}\in\Omega} \left|\widetilde{\boldsymbol{Q}}_n(\boldsymbol{w}) - \boldsymbol{Q}(\boldsymbol{w})\right| \leq \sqrt{\frac{\theta\varrho + \log\left(\frac{4}{\varepsilon}\right)}{2n}}, \qquad (2)$$

*where the total freedom degree $\theta$ of the network is $\theta = a_{l+1}(d_{l+1} + \tilde{r}_l \tilde{c}_l d_l + 1) + \sum_{i=1}^{l} a_i (k_i{}^2 d_{i-1} + d_i + 1)$ and $\varrho = \sum_{i=1}^{l} \log\left(\frac{\sqrt{d_i} b_i (k_i - s_i + 1)}{4p}\right) + \log(b_{l+1}) + \log\left(\frac{n}{128p^2}\right)$.*

To our best knowledge, this generalization error rate is the first one that grows linearly (in contrast to exponentially) with depth $l$ without needing any special training procedure. This can be observed from the fact that our result only depends on $\mathcal{O}(\sum_{i=1}^{l} \log(b_i))$, rather than an exponential factor $\mathcal{O}(\prod_{i=1}^{l+1} b_i)$ which appears in some existing works, *e.g.* the uniform convergence of the empirical risk in deep CNNs (Sun et al., 2016) and fully feedforward networks (Bartlett & Maass, 2003; Neyshabur et al., 2015; Zhou & Feng, 2018). This faster convergence rate is achieved by adopting similar analysis technique in (Mei et al., 2017; Zhou & Feng, 2018) but we derive tighter bounds on the related parameters featuring distributions of the empirical risk and its gradient, with milder assumptions. For instance, both (Zhou & Feng, 2018) and this work show that the empirical risk follows a sub-Gaussian distribution. But Zhou & Feng (2018) used Gaussian concentration inequality and thus need Lipschitz constant of loss which exponentially depends on the depth. In contrast, we use $\epsilon$-net to decouple the dependence between input $\boldsymbol{D}$ and parameter $\boldsymbol{w}$ and then adopt Hoeffding's inequality, only requiring the constant magnitude bound of loss and geting rid of exponential term.

Based on Lemma 1, we derive generalization error of deep CNNs in Theorem 1 with proof in Sec. D.2 in supplement.

**Theorem 1.** *Assume that in CNNs, $\sigma_1$ and $\sigma_2$ are respectively the sigmoid and softmax functions and the loss function $f(g(\boldsymbol{w}; \boldsymbol{D}), \boldsymbol{y})$ is squared loss. Suppose Assumptions 1 and 2 hold. Then with probability at least $1-\varepsilon$, the generalization error of a deep CNN model is bounded as*

$$\mathbb{E}_{\boldsymbol{\mathcal{S}}\sim\boldsymbol{\mathcal{D}}}\left|\mathbb{E}_{\mathcal{A}}\left(\boldsymbol{Q}(\widetilde{\boldsymbol{w}}) - \widetilde{\boldsymbol{Q}}_n(\widetilde{\boldsymbol{w}})\right)\right| \leq \sqrt{\frac{\theta\varrho + \log\left(\frac{4}{\varepsilon}\right)}{2n}},$$

*where $\theta$ and $\varrho$ are given in Lemma 1.*

By inspecting Theorem 1, one can find that the generalization error diminishes at the rate of $\mathcal{O}\left(1/\sqrt{n}\right)$ (up to a log factor). Besides, Theorem 1 explicitly reveals the roles of network parameters in determining model generalization performance. Such transparent results form stark contrast to the works (Sun et al., 2016) and (Kawaguchi et al., 2017) (see more comparison in Sec. 3.2). Notice, our technique also applies to other third-order differentiable activation functions, *e.g.* tanh, and other losses, *e.g.* cross entropy, with only slight difference in the results.

First, the freedom degree $\theta$ of network parameters, which depends on the network size and the redundancy in parameters, plays an important role in the generalization error bound. More specifically, to obtain smaller generalization error, more samples are needed to train a deep CNN model having larger freedom degree $\theta$. As aforementioned, although the results in Theorem 1 are obtained under the low-rankness condition defined on the parameter matrix consisting of kernels per layer, they are easily extended to the (tensor) low-rankness defined on each kernel individually. The low-rankness captures common parameter redundancy in practice. For instance, (Lebedev et al., 2014; Jaderberg et al., 2014) showed that parameter redundancy exists in a trained network model and can be squeezed by low-rank tensor decomposition. The classic residual function (He et al., 2016; Zagoruyko & Komodakis, 2016) with three-layer bottleneck architecture ($1 \times 1$, $3 \times 3$ and $1 \times 1$ convs) has rank 1 in generalized block term decomposition (Chen et al., 2017; Cohen & Shashua, 2016). Similarly, inception networks (Szegedy et al., 2017) explicitly decomposes a convolutional kernel of large size into two separate convolutional kernels of smaller size (*e.g.* a $7 \times 7$ kernel is replaced by two multiplying kernels of size $7 \times 1$ and $1 \times 7$). Employing these low-rank approximation techniques helps reduce the freedom degree and provides smaller generalization error. Notice, the low-rankness assumption only affects the freedom degree $\theta$. Without this assumption, $\theta$ will be replaced by the total parameter number of the network.

From the factor $\varrho$, one can observe that the kernel size $k_i$ and its stride $s_i$ determine the generalization error but in opposite ways. Larger kernel size $k_i$ leads to larger generalization error, while larger stride $s_i$ provides smaller one. This is because both larger kernel and smaller stride increase the model complexity, since larger kernel means more trainable parameters and smaller stride implies larger size of feature maps in the subsequent layer. Also, the pooling operation in the first $l$ convolutional layers helps reduce the generalization error, as reflected by the factor $1/p$ in $\varrho$.

Furthermore, the number of feature maps (*i.e.* channels) $d_i$ appearing in the $\theta$ and $\varrho$ also affects the generalization error. A wider network with larger $d_i$ requires more samples for training such that it can generalize well. This is because (1) a larger $d_i$ indicates more trainable parameters, which usually increases the freedom degree $\theta$, and (2) a larger $d_i$ also requires larger kernels $\boldsymbol{W}_{(i)}^j$ with more channel-wise dimensions since there are more channels to convolve, leading to a larger magnitude bound $b_i$ for the kernel $\boldsymbol{W}_{(i)}^j$. Therefore, as suggested by Theorem 1, a thin network is more preferable than a fat network. Such an observation is consistent with other analysis works on the network expression ability (Eldan & Shamir, 2016; Lu et al., 2017) and the architecture-engineering practice, such as (He et al., 2016; Szegedy et al., 2015). By comparing contributions of the architecture and parameter magnitude to the generalization performance, we find that the generalization error usually depends on the network architecture parameters linearly or more heavily, and also on parameter magnitudes but with a logarithm term $\log b_i$. This implies the architecture plays a more important role than the parameter magnitudes. Therefore, for achieving better generalization performance in practice, architecture engineering is indeed essential.

Finally, by observing the factor $\varrho$, we find that imposing certain regularization, such as $\|\boldsymbol{w}\|_2^2$, on the trainable parameters is useful. The effectiveness of such a regularization will be more significant when imposing on the weight matrix of the fully connected layer due to its large size. Such a regularization technique, in deep learning literature, is well known as "weight decay". This conclusion is consistent with other analysis works on the deep forward networks, such as (Bartlett & Maass, 2003; Neyshabur et al., 2015; Zhou & Feng, 2018).

### 3.2. Discussions

Sun et al. (2016) also analyzed generalization error bound in deep CNNs but employing different techniques. They proved that the Rademacher complexity $\mathcal{R}_m(\mathcal{F})$ of a deep CNN model with sigmoid activation functions is $\mathcal{O}(\widetilde{b}_x(2p\widetilde{b})^{l+1}\sqrt{\log(r_0c_0)}/\sqrt{n})$ where $\mathcal{F}$ denotes the function hypothesis that maps the input data $\boldsymbol{D}$ to $\boldsymbol{v} \in \mathbb{R}^{d_{l+1}}$ by the analyzed CNN model. Here $\widetilde{b}_x$ denotes the upper bound of the absolute entry values in the input datum $\boldsymbol{D}$, *i.e.* $\widetilde{b}_x \geq |\boldsymbol{D}_{i,j}|$ $(\forall i,j)$, and $\widetilde{b}$ obeys $\widetilde{b} \geq \max\{\max_i \sum_{j=1}^{d_i} \|\boldsymbol{W}_{(i)}^j\|_1, \|\boldsymbol{W}_{(l+1)}\|_1\}$. Sun et al. (2016) showed that with probability at least $1-\varepsilon$, the difference between the empirical margin error $\mathtt{err}_e^\gamma(g)$ $(g \in \mathcal{F})$ and the population margin error $\mathtt{err}_p^\gamma(g)$ can be bounded as

$$\mathtt{err}_p^\gamma(g) \leq \inf_{\gamma>0}\left[\mathtt{err}_e^\gamma(g) + \frac{8d_{l+1}(2d_{l+1}-1)}{\gamma}\mathcal{R}_m(\mathcal{F}) + \sqrt{\frac{\log\log_2(2/\gamma)}{n}} + \sqrt{\frac{\log(2/\varepsilon)}{n}}\right], \quad (3)$$

where $\gamma$ controls the error margin since it obeys $\gamma \geq \boldsymbol{v}_y - \max_{k\neq y}\boldsymbol{v}_k$ and $y$ denotes the label of $\boldsymbol{v}$. However, the bound in Eqn. (3) is practically loose, since $\mathcal{R}_m(\mathcal{F})$ involves the exponential factor $(2\widetilde{b})^{l+1}$ which is usually very large. In this case, $\mathcal{R}_m(\mathcal{F})$ is extremely large. By comparison, the bound provided in our Theorem 1 only depends on $\sum_{i=1}^{l+1}\log(b_i)$ which avoids the exponential growth along with the depth $l$, giving a much tighter and more practically meaningful bound. The generalization error bounds in (Kawaguchi et al., 2017) either depend on a specific dataset or rely on restrictive and rarely used training procedure, while our Theorem 1 is independent of any specific dataset or training procedure, rendering itself more general. More importantly, the results in Theorem 1 make the roles of network parameters transparent, which could benefit understanding and architecture design of CNNs.

# 4. Optimization Guarantees for Deep CNNs

Although deep CNNs are highly non-convex, gradient descent based algorithms usually perform quite well on optimizing the models in practice. After characterizing the roles of different network parameters for the generalization performance, here we present optimization guarantees for gradient descent based algorithms in training CNNs.

Specifically, in practice one usually adopts SGD or its variants, such as adam and RMSProp, to optimize the CNN models. Such algorithms usually terminate when the gradient magnitude decreases to a low level and the training hardly proceeds. This implies that the algorithms in fact compute an $\epsilon$-approximate stationary point $\widetilde{\boldsymbol{w}}$ for the loss function $\widetilde{\boldsymbol{Q}}_n(\boldsymbol{w})$, *i.e.* $\|\nabla_{\boldsymbol{w}}\widetilde{\boldsymbol{Q}}_n(\widetilde{\boldsymbol{w}})\|_2^2 \leq \epsilon$. Here we explore such a problem: by computing an $\epsilon$-stationary point $\widetilde{\boldsymbol{w}}$ of the empirical risk $\widetilde{\boldsymbol{Q}}_n(\boldsymbol{w})$, can we also expect $\widetilde{\boldsymbol{w}}$ to be sufficiently good for generalization, or in other words expect that it is also an approximate stationary point for the population risk $\boldsymbol{Q}(\boldsymbol{w})$? To answer this question, first we analyze the relationship between the empirical gradient $\nabla_{\boldsymbol{w}}\widetilde{\boldsymbol{Q}}_n(\boldsymbol{w})$ and its population counterpart $\nabla_{\boldsymbol{w}}\boldsymbol{Q}(\boldsymbol{w})$. Founded on this, we further establish convergence of the empirical gradient of the computed solution to its corresponding population gradient. Finally, we present the bounded distance between the computed solution $\widetilde{\boldsymbol{w}}$ and the optimum $\boldsymbol{w}^*$.

To our best knowledge, this work is the first one that analyzes the optimization behavior of gradient descent based algorithms for training multi-layer CNN models with the commonly used convolutional and pooling operations.

## 4.1. Convergence Guarantees on Gradients

Here we present guarantees on convergence of the empirical gradient to the population one in Theorem 2. As aforementioned, such results imply good generalization performance of the computed solution $\widetilde{\boldsymbol{w}}$ to the empirical risk $\widetilde{\boldsymbol{Q}}_n(\boldsymbol{w})$.

**Theorem 2.** *Assume that in CNNs, $\sigma_1$ and $\sigma_2$ respectively are the sigmoid and softmax functions and the loss function $f(g(\boldsymbol{w};\boldsymbol{D}),\boldsymbol{y})$ is squared loss. Suppose Assumptions 1 and 2 hold. Then the empirical gradient* uniformly *converges to the population gradient in Euclidean norm. More specifically, there exist universal constants $c_{g'}$ and $c_g$ such that if $n \geq c_{g'}\frac{l^2 {b_{l+1}}^2(b_{l+1}+\sum_{i=1}^l d_i b_i)^2 (r_0 c_0 d_0)^4}{d_0^4 {b_1}^8 (d\log(6)+\theta\varrho)\varepsilon^2 \max_i(r_i c_i)}$, then*

$$\sup_{\boldsymbol{w}\in\Omega}\left\|\nabla_{\boldsymbol{w}}\widetilde{\boldsymbol{Q}}_n(\boldsymbol{w})-\nabla_{\boldsymbol{w}}\boldsymbol{Q}(\boldsymbol{w})\right\|_2 \leq c_g\beta\sqrt{\frac{2d+\theta\varrho+\log\left(\frac{4}{\varepsilon}\right)}{2n}}$$

*holds with probability at least $1-\varepsilon$, where $\varrho$ is provided in Lemma 1. Here $\beta$ and $d$ are defined as $\beta = \left[\frac{r_l c_l d_l}{8p^2} + \sum_{i=1}^l \frac{{b_{l+1}}^2 d_{i-1}}{8p^2 {b_i}^2 d_i} r_{i-1}c_{i-1}\prod_{j=i}^l \frac{d_j {b_j}^2 (k_j-s_j+1)^2}{16p^2}\right]^{1/2}$ and $d = \tilde{r}_l\tilde{c}_l d_l d_{l+1} + \sum_{i=1}^l {k_i}^2 d_{i-1}d_i$, respectively.*

Its proof is given in Sec. D.3 in supplement. From Theorem 2, the empirical gradient converges to the population one at the rate of $\mathcal{O}(1/\sqrt{n})$ (up to a log factor). In Sec. 3.1, we have discussed the roles of the network architecture parameters in $\varrho$. Here we further analyze the effects of the network parameters on the optimization behavior through the factor $\beta$. The roles of the kernel size $k_i$, the stride $s_i$, the pooling size $p$ and the channel number $d_i$ in $\beta$ are consistent with those in Theorem 1. The extra factor $r_i c_i$ advocates not building such CNN networks with extremely large feature map sizes. The total number of parameters $d$ is involved here instead of the degree of freedom because the gradient $\nabla_{\boldsymbol{w}}\widetilde{\boldsymbol{Q}}_n(\boldsymbol{w})$ may not have low-rank structures.

Based on Theorem 2, we can further conclude that if the computed solution $\widetilde{\boldsymbol{w}}$ is an $\epsilon$-approximate stationary point of the empirical risk, then it is also a $4\epsilon$-approximate stationary point of the population risk. We state this result in Corollary 1 with proof in Sec. D.4 in supplement.

**Corollary 1.** *Suppose assumptions in Theorem 2 hold and we have $n \geq (d\varrho + \log(4/\varepsilon))\beta^2/\epsilon$. Then if the solution $\widetilde{\boldsymbol{w}}$ computed by minimizing the empirical risk obeys $\|\nabla\widetilde{\boldsymbol{Q}}_n(\widetilde{\boldsymbol{w}})\|_2^2 \leq \epsilon$, we have $\|\nabla\boldsymbol{Q}(\widetilde{\boldsymbol{w}})\|_2^2 \leq 4\epsilon$ with probability at least $1-\varepsilon$.*

Corollary 1 shows that by using full gradient descent algorithms to minimize the empirical risk, the computed approximate stationary point $\widetilde{\boldsymbol{w}}$ is also close to the desired stationary point $\boldsymbol{w}^*$ of the population risk. This guarantee is also applicable to other stochastic gradient descent based algorithms, like SGD, adam and RMSProp, by applying recent results on obtaining $\epsilon$-approximate stationary point for nonconvex problems (Ghadimi & Lan, 2013; Tieleman & Hinton, 2012; Kingma & Ba, 2015). Accordingly, the computed solution $\widetilde{\boldsymbol{w}}$ has guaranteed generalization performance on new data. It partially explains the success of gradient descent based optimization algorithms for CNNs.

## 4.2. Convergence of Stationary Points

Here we go further and directly characterize the distance between stationary points in the empirical risk $\widetilde{\boldsymbol{Q}}_n(\boldsymbol{w})$ and its population counterpart $\boldsymbol{Q}(\boldsymbol{w})$. Compared with the results for the risk and gradient, the results on stationary points give more direct performance guarantees for CNNs. Here we only analyze the non-degenerate stationary points including local minimum/maximum and non-degenerate saddle points, as they are geometrically isolated and thus are unique in local regions. We first introduce some necessary definitions.

**Definition 2.** (Non-degenerate stationary points and saddle points) *(Gromoll & Meyer, 1969) A stationary point $\boldsymbol{w}$ is said to be a* non-degenerate stationary point *of $\boldsymbol{Q}(\boldsymbol{w})$ if*

$$\inf_i \left|\lambda_i\left(\nabla^2\boldsymbol{Q}(\boldsymbol{w})\right)\right| \geq \zeta,$$

*where $\lambda_i\left(\nabla^2\boldsymbol{Q}(\boldsymbol{w})\right)$ is the i-th eigenvalue of the Hessian*

$\nabla^2 \boldsymbol{Q}(\boldsymbol{w})$ *and* $\zeta$ *is a positive constant. A stationary point is said to be a* saddle point *if the smallest eigenvalue of its Hessian* $\nabla^2 \boldsymbol{Q}(\boldsymbol{w})$ *has a negative value.*

Suppose $\boldsymbol{Q}(\boldsymbol{w})$ has $m$ non-degenerate stationary points which are denoted as $\{\boldsymbol{w}_{(1)}, \boldsymbol{w}_{(2)}, \cdots, \boldsymbol{w}_{(m)}\}$. We have following results on the geometry of these stationary points in Theorem 3. The proof is given in Sec. D.5 in supplement.

**Theorem 3.** *Assume in CNNs,* $\sigma_1$ *and* $\sigma_2$ *are respectively the sigmoid and softmax activation functions and the loss* $f(g(\boldsymbol{w};\boldsymbol{D}),\boldsymbol{y})$ *is squared loss. Suppose Assumptions 1 and 2 hold. Then if* $n \geq c_h \max\big(\frac{d+\theta\varrho}{\zeta^2}, \frac{l^2 b_{l+1}{}^2 (b_{l+1}+\sum_{i=1}^{l} d_i b_i)^2 (r_0 c_0 d_0)^4}{d_0^4 b_1{}^8 d\varrho\varepsilon^2 \max_i (r_i c_i)}\big)$ *where* $c_h$ *is a constant, for* $k \in \{1, \cdots, m\}$*, there exists a non-degenerate stationary point* $\boldsymbol{w}_n^{(k)}$ *of* $\widetilde{\boldsymbol{Q}}_n(\boldsymbol{w})$ *which uniquely corresponds to the non-degenerate stationary point* $\boldsymbol{w}_{(k)}$ *of* $\boldsymbol{Q}(\boldsymbol{w})$ *with probability at least* $1-\varepsilon$*. Moreover, with same probability the distance between* $\boldsymbol{w}_n^{(k)}$ *and* $\boldsymbol{w}_{(k)}$ *is bounded as*

$$\|\boldsymbol{w}_n^{(k)} - \boldsymbol{w}_{(k)}\|_2 \leq \frac{2c_g\beta}{\zeta}\sqrt{\frac{2d+\theta\varrho+\log\left(\frac{4}{\varepsilon}\right)}{2n}}, \ (1 \leq k \leq m),$$

*where* $\varrho$ *and* $\beta$ *are given in Lemma 1 and Theorem 2, respectively.*

Theorem 3 shows that there exists exact one-to-one correspondence between the non-degenerate stationary points of the empirical risk $\widetilde{\boldsymbol{Q}}_n(\boldsymbol{w})$ and the popular risk $\boldsymbol{Q}(\boldsymbol{w})$ for CNNs, if the sample size $n$ is sufficiently large. Moreover, the non-degenerate stationary point $\boldsymbol{w}_n^{(k)}$ of $\widetilde{\boldsymbol{Q}}_n(\boldsymbol{w})$ is very close to its corresponding non-degenerate stationary point $\boldsymbol{w}_{(k)}$ of $\boldsymbol{Q}(\boldsymbol{w})$. More importantly, their distance shrinks at the rate of $\mathcal{O}\left(1/\sqrt{n}\right)$ (up to a log factor). The network parameters have similar influence on the distance bounds as explained in the above subsection. Compared with gradient convergence rate in Theorem 2, the convergence rate of corresponding stationary point pairs in Theorem 3 has an extra factor $1/\zeta$ that accounts for the geometric topology of non-degenerate stationary points, similar to other works like stochastic optimization analysis (Duchi & Ruan, 2016).

For degenerate stationary points to which the corresponding Hessian matrix has zero eigenvalues, one cannot expect to establish unique correspondence for stationary points in empirical and population risks, since they are not isolated anymore and may reside in flat regions. But Theorem 2 guarantees that the gradients of $\widetilde{\boldsymbol{Q}}_n(\boldsymbol{w})$ and $\boldsymbol{Q}(\boldsymbol{w})$ at these points are close. This implies a degenerate stationary point of $\boldsymbol{Q}(\boldsymbol{w})$ will also give a near-zero gradient for $\widetilde{\boldsymbol{Q}}_n(\boldsymbol{w})$, indicating it is also a stationary point for $\widetilde{\boldsymbol{Q}}_n(\boldsymbol{w})$.

Du et al. (2017a;b) showed that for a simple and shallow CNN consisting of only one non-overlapping convolutional layer, (stochastic) gradient descent algorithms with weight normalization can converge to the global minimum. In contrast to their simplified models, we analyze complex multi-layer CNNs with the commonly used convolutional and pooling operations. Besides, we provide results on both gradient and the distance between the computed solution and desired stationary points, which are applicable to arbitrary gradient descent based algorithms.

Next, based on Theorem 3, we derive that the corresponding pair $(\boldsymbol{w}_n^{(k)}, \boldsymbol{w}_{(k)})$ in the empirical and population risks shares the same geometrical property stated in Corollary 2.

**Corollary 2.** *Suppose the assumptions in Theorem 3 hold. If any one in the pair* $(\boldsymbol{w}_n^{(k)}, \boldsymbol{w}_{(k)})$ *in Theorem 3 is a local minimum or saddle point, so is the other one.*

See the proof of Corollary 2 in Sec. D.6 in supplement. Corollary 2 tells us that the corresponding pair, $\boldsymbol{w}_n^{(k)}$ and $\boldsymbol{w}_{(k)}$, has the same geometric property. Namely, if either one in the pair is a local minimum or saddle point, so is the other one. This result is important for optimization. If the computed solution $\widetilde{\boldsymbol{w}}$ by minimizing the empirical risk $\widetilde{\boldsymbol{Q}}_n(\boldsymbol{w})$ is a local minimum, then it is also a local minimum of the population risk $\boldsymbol{Q}(\boldsymbol{w})$. Thus it partially explains why the computed solution $\widetilde{\boldsymbol{w}}$ can generalize well on new data. This property also benefits designing new optimization algorithms. For example, Saxe et al. (2014) and Kawaguchi (2016) pointed out that degenerate stationary points indeed exist for deep linear neural networks and Dauphin et al. (2014) empirically validated that saddle points are usually surrounded by high error plateaus in deep forward neural networks. So it is necessary to avoid the saddle points and find the local minimum of population risk. From Theorem 3, it is clear that one only needs to find the local minimum of empirical risk by using escaping saddle points algorithms, *e.g.* (Ge et al., 2015; Jin et al., 2017; Agarwal et al., 2017).

## 5. Comparison on DNNs And CNNs

Here we compare deep feedforward neural networks (DNNs) with deep CNNs from their generalization error and optimization guarantees to theoretically explain why CNNs are more preferable than DNNs, to some extent.

By assuming the input to be standard Gaussian $\mathcal{N}(0,\tau^2)$, Zhou & Feng (2018) proved that if $n \geq 18r^2/(d\tau^2\varepsilon^2 \log(l+1))$, with probability $1-\varepsilon$, the generalization error of an $(l+1)$-layer DNN model with sigmoid activation functions is bounded by $\epsilon_n$:

$$\epsilon_n \triangleq c_n \tau \sqrt{(1+c_r l)\max_i \boldsymbol{d}_i}\sqrt{\frac{d\log(n(l+1))+\log(4/\varepsilon)}{n}},$$

where $c_n$ is a universal constant; $\boldsymbol{d}_i$ denotes the width of the $i$-th layer; $d$ is the total parameter number of the network; $c_r = \max(\widehat{r}^2/16, \left(\widehat{r}^2/16\right)^l)$ where $\widehat{r}$ upper bounds Frobenius norm of the weight matrix in each layer. Recall that the generalization bound of CNN provided in this work is $\mathcal{O}(\sqrt{\theta\widetilde{\varrho}/(2n)})$, where $\widetilde{\varrho} = \sum_{i=1}^{l} \log\left(\sqrt{d_i} b_i(k_i - s_i + 1)/(4p)\right) + \log(b_{l+1})$.

By observing the above two generalization bounds, one can see when the layer number is fixed, CNN usually has smaller generalization error than DNN because: (1) CNN usually has much fewer parameters, *i.e.* smaller $d$, than DNN due to parameter sharing mechanism of convolutions. (2) The generalization error of CNN has a smaller factor than DNN in the network parameter magnitudes. Generalization error bound of CNN depends on a logarithm term $\mathcal{O}(\log\prod_{i=1}^{l+1} b_i)$ of the magnitude $b_i$ of each kernel/weight matrix, while the bound for DNN depends on $\mathcal{O}(\widehat{r}^{l+1})$. Since the kernel size is much smaller than that of the weight matrix in the fully connected layer by unfolding the convolutional layer, the upper magnitude bound $b_i$ $(i = 1, \cdots, l)$ of each kernel is usually much smaller than $\widehat{r}$. (3) The pooling operation and the stride in convolutional operation in CNN also benefit its generalization performance. This is because the factor $\widetilde{\varrho}$ involves $\mathcal{O}(2(l+1)\log(1/p))$ and $(k_i - s_i + 1)$ which also reduce the generalization error. Notice, by applying our analysis technique, it might be possible to remove the exponential term in DNN. But as mentioned above, the unique operations, *e.g.* convolution, pooling and striding, still benefit CNN, making it generalize better than DNN.

Because of the above factors, the empirical gradient in CNN converges to its population counterpart faster, as well as the paired non-degenerate stationary points for empirical risk and population risk. All these results guarantee that for an arbitrary gradient descent based algorithm, it is faster to compute an approximate stationary point or a local minimum in population risk of CNN compared with DNN.

# 6. Proof of Roadmap

Here we briefly introduce the proof roadmap. Due to space limitation, all the proofs of our theoretical results are deferred to the supplement. Firstly, our analysis relies on bounding the gradient magnitude and the spectral norm of Hessian of the loss $f(g(\boldsymbol{w};\boldsymbol{D}),\boldsymbol{y})$. By considering multilayer architecture of CNN, we establish recursive relation of their magnitudes in the $k$ and $k+1$ layers (Lemmas 9 $\sim$ 14 in supplement) and get their overall magnitude upper bound.

For the uniform convergence $\sup_{\boldsymbol{w}\in\Omega}|\widetilde{\boldsymbol{Q}}_n(\boldsymbol{w}) - \boldsymbol{Q}(\boldsymbol{w})|$ in Lemma 1, we resort to bound three distances: $A_1 = \sup_{\boldsymbol{w}\in\Omega}|\widetilde{\boldsymbol{Q}}_n(\boldsymbol{w}) - \widetilde{\boldsymbol{Q}}_n(\boldsymbol{w}_{k_w})|$, $A_2 = \sup_{\boldsymbol{w}_{k_w}\in\Theta} |\widetilde{\boldsymbol{Q}}_n(\boldsymbol{w}_{k_w}) - \mathbb{E}\widetilde{\boldsymbol{Q}}_n(\boldsymbol{w}_{k_w})|$ and $A_3 = \sup_{\boldsymbol{w}\in\Omega}|\mathbb{E}\widetilde{\boldsymbol{Q}}_n(\boldsymbol{w}_{k_w}) - \mathbb{E}\boldsymbol{Q}(\boldsymbol{w})|$, where $\boldsymbol{w}_{k_w}$ belongs to the $\epsilon$-net $\Theta$ of parameter domain $\Omega$. Using Markov inequality and Lipschitz property of loss, we can bound $A_1$ and $A_3$. To bound $A_2$, we prove the empirical risk is sub-Gaussian. Considering the element $\boldsymbol{w}_{k_w}$ in $\epsilon$-net $\Theta$ is independent of input $\boldsymbol{D}$, we use Hoeffding's inequality to prove empirical risk at point $\boldsymbol{w}_{k_w}$ to be sub-Gaussian for any $\boldsymbol{w}_{k_w}$ in $\Theta$. By this decoupling of $\epsilon$-net, our bound on $A_2$ depends on the constant magnitude of loss and gets rid of exponential term. Combining these bounds together, we obtain the uniform convergence of empirical risk and can derive the generalization bound.

We use a similar decomposition and decoupling strategy mentioned above to bound gradient uniform convergence $\sup_{\boldsymbol{w}\in\Omega}\|\nabla_{\boldsymbol{w}}\widetilde{\boldsymbol{Q}}_n(\boldsymbol{w}) - \nabla_{\boldsymbol{w}}\boldsymbol{Q}(\boldsymbol{w})\|_2$ in Theorem 2. But here we need to bound gradient and spectral norm of Hessian.

To prove correspondence and bounded distance of stationary points, we define a set $G = \{\boldsymbol{w}\in\Omega : \|\nabla\widetilde{\boldsymbol{Q}}_n(\boldsymbol{w})\| \leq \epsilon$ and $\inf_i |\lambda_i(\nabla^2\widetilde{\boldsymbol{Q}}_n(\boldsymbol{w}))| \geq \zeta\}$ where $\lambda_i$ is the $i$-th eigenvalue of $\nabla^2\widetilde{\boldsymbol{Q}}_n(\boldsymbol{w})$. Then $G$ is decomposed into countable components each of which has one or zero non-degenerate stationary point. Next we prove the uniform convergence between empirical and population Hessian by using a similar strategy as above. Combining uniform convergence of gradient and Hessian and the results in differential topology (Lemmas 4 & 5 in supplement), we obtain that for each component of $G$, if there is a unique non-degenerate stationary point in $\boldsymbol{Q}(\boldsymbol{w})$, then $\widetilde{\boldsymbol{Q}}_n(\boldsymbol{w})$ also has a unique one, and vice versa. This gives the one-to-one correspondence relation. Finally, the uniform convergence of gradient and Hessian can bound the distance between the corresponding points.

# 7. Conclusion

In this work, we theoretically analyzed why deep CNNs can achieve remarkable success, from its generalization performance and the optimization guarantees of (stochastic) gradient descent based algorithms. We proved that the generalization error of deep CNNs can be bounded by a factor which depends on the network parameters. Moreover, we analyzed the relationship between the computed solution by minimizing the empirical risk and the optimum solutions in population risk from their gradient and their Euclidean distance. All these results show that with sufficient training samples, the generalization performance of deep CNN models can be guaranteed. Besides, these results also reveal that the kernel size $k_i$, the stride $s_i$, the pooling size $p$, the channel number $d_i$ and the freedom degree $\theta$ of the network parameters are critical to the generalization performance of deep CNNs. We also showed that the weight parameter magnitude is also important. These suggestions on network designs accord with the widely used network architectures.

# Acknowledgements

Jiashi Feng was partially supported by NUS startup R-263-000-C08-133, MOE Tier-I R-263-000-C21-112, NUS IDS R-263-000-C67-646, ECRA R-263-000-C87-133 and MOE Tier-II R-263-000-D17-112.

# Understanding Generalization and Optimization Performance for Deep CNNs (Supplementary File)

Pan Zhou* Jiashi Feng*
* National University of Singapore, Singapore
pzhou@u.nus.edu elefjia@nus.edu.sg

## A Structure of This Document

This document gives some other necessary notations and preliminaries for our analysis in Sec. B.1 and provides auxiliary lemmas in Sec. B.2. Then in Sec. C, we present the technical lemmas for proving our final results and their proofs. Next, in Sec. D, we utilize these technical lemmas to prove our desired results. Finally, we give the proofs of other auxiliary lemmas in Sec. E.

As for the results in manuscript, the proofs of Lemma 1 and Theorem 1 in Sec. 3.1 in the manuscript are respectively provided in Sec. D.1 and Sec. D.2. As for the results in Sec. 4 in the manuscript, Sec. D.3 and D.4 present the proofs of Theorem 2 and Corollary 1, respectively. Finally, we respectively introduce the proofs of Theorem 3 and Corollary 2 in Sec. D.5 and D.6.

## B Notations and Preliminary Tools

Beyond the notations introduced in the manuscript, we need some other notations used in this document. Then we introduce several lemmas that will be used later.

### B.1 Notations

Throughout this document, we use $\langle\cdot,\cdot\rangle$ to denote the inner product and use $\widetilde{\circledast}$ to denote the convolution operation with stride 1. $\boldsymbol{A}\otimes\boldsymbol{C}$ denotes the Kronecker product between $\boldsymbol{A}$ and $\boldsymbol{C}$. Note that $\boldsymbol{A}$ and $\boldsymbol{C}$ in $\boldsymbol{A}\otimes\boldsymbol{C}$ can be matrices or vectors. For a matrix $\boldsymbol{A}\in\mathbb{R}^{n_1\times n_2}$, we use $\|\boldsymbol{A}\|_F=\sqrt{\sum_{i,j}\boldsymbol{A}_{ij}^2}$ to denote its Frobenius norm, where $\boldsymbol{A}_{ij}$ is the $(i,j)$-th entry of $\boldsymbol{A}$. We use $\|\boldsymbol{A}\|_{\text{op}}=\max_i|\lambda_i(\boldsymbol{A})|$ to denote the operation norm of a matrix $\boldsymbol{A}\in\mathbb{R}^{n_1\times n_1}$, where $\lambda_i(\boldsymbol{A})$ denotes the $i$-th eigenvalue of the matrix $\boldsymbol{A}$. For a 3-way tensor $\boldsymbol{A}\in\mathbb{R}^{s\times t\times q}$, its operation norm is computed as

$$\|\boldsymbol{A}\|_{\text{op}}=\sup_{\|\boldsymbol{\lambda}\|_2\leq 1}\left\langle\boldsymbol{\lambda}^{\otimes^3},\boldsymbol{A}\right\rangle=\sum_{i,j,k}\boldsymbol{A}_{ijk}\boldsymbol{\lambda}_i\boldsymbol{\lambda}_j\boldsymbol{\lambda}_k,$$

where $\boldsymbol{A}_{ijk}$ denotes the $(i,j,k)$-th entry of $\boldsymbol{A}$.

For brevity, in this document we use $f(\boldsymbol{w},\boldsymbol{D})$ to denote $f(g(\boldsymbol{w};\boldsymbol{D}),\boldsymbol{y})$ in the manuscript. Let $\boldsymbol{w}_{(i)}=(\boldsymbol{w}_{(i)}^1;\cdots;\boldsymbol{w}_{(i)}^{d_i})\in\mathbb{R}^{k_i{}^2d_{i-1}d_i}$ $(i=1,\cdots,l)$ be the parameter of the $i$-th layer where $\boldsymbol{w}_{(i)}^k=\mathsf{vec}\left(\boldsymbol{W}_{(i)}^k\right)\in\mathbb{R}^{k_i{}^2d_{i-1}}$ is the vectorization of $\boldsymbol{W}_{(i)}^k$. Similarly, let $\boldsymbol{w}_{(l+1)}=\mathsf{vec}\left(\boldsymbol{W}_{(l+1)}\right)\in\mathbb{R}^{r_lc_ld_ld_{l+1}}$. Then, we further define $\boldsymbol{w}=(\boldsymbol{w}_{(1)},\cdots,\boldsymbol{w}_{(l)},\boldsymbol{w}_{(l+1)})\in\mathbb{R}^{r_lc_ld_ld_{l+1}+\sum_{i=1}^l k_i{}^2d_{i-1}d_i}$ which contains all the parameter in the network. Here we use $\boldsymbol{W}_{(i)}^k$ to denote the $k$-th kernel in the $i$-th convolutional layer. For brevity, let $\boldsymbol{W}_{(i)}^{k,j}$ denotes the $j$-th slice of $\boldsymbol{W}_{(i)}^k$, *i.e.* $\boldsymbol{W}_{(i)}^{k,j}=\boldsymbol{W}_{(i)}^k(:,:,j)$.

For a matrix $\boldsymbol{M}\in\mathbb{R}^{s\times t}$, $\widehat{\boldsymbol{M}}$ denotes the matrix which is obtained by rotating the matrix $\boldsymbol{M}$ by 180 degrees. Then we use $\boldsymbol{\delta}_i$ to denote the gradient of $f(\boldsymbol{w},\boldsymbol{D})$ *w.r.t.* $\boldsymbol{X}_{(i)}$:

$$\boldsymbol{\delta}_i=\nabla_{\boldsymbol{X}_{(i)}}f(\boldsymbol{w},\boldsymbol{D})\in\mathbb{R}^{r_i\times c_i\times d_i},\quad(i=1,\cdots,l),$$

Based on $\boldsymbol{\delta}_i$, we further define $\widetilde{\boldsymbol{\delta}}_i \in \mathbb{R}^{(\tilde{r}_{i-1}-k_i+1)\times(\tilde{c}_{i-1}-k_i+1)\times d_i}$. Each slice $\widetilde{\boldsymbol{\delta}}_{i+1}^k$ can be computed as follows. Firstly, let $\widetilde{\boldsymbol{\delta}}_{i+1}^k = \boldsymbol{\delta}_{i+1}^k$. Then, we pad zeros of $s_i - 1$ rows between the neighboring rows in $\widetilde{\boldsymbol{\delta}}_{i+1}^k$ and similarly we pad zeros of $s_i - 1$ columns between the neighboring columns in $\widetilde{\boldsymbol{\delta}}_{i+1}^k$. Accordingly, the size of $\widetilde{\boldsymbol{\delta}}_{i+1}^k$ is $(s_i(r_i-1)+1)\times(s_i(c_i-1)+1)$. Finally, we pad zeros of width $k_i - 1$ around $\widetilde{\boldsymbol{\delta}}_{i+1}^k$ to obtain new $\widetilde{\boldsymbol{\delta}}_{i+1}^k \in \mathbb{R}^{(s_i(r_i-1)+2k_i-1)\times(s_i(c_i-1)+2k_i-1)}$. Note that since $r_{i+1} = (\tilde{r}_i - k_{i+1})/s_{i+1} + 1$ and $r_{i+1} = (\tilde{r}_i - k_{i+1})/s_{i+1} + 1$, we have $s_i(r_i-1)+2k_i-1 = \tilde{r}_{i-1} - k_i + 1$ and $s_i(c_i-1)+2k_i-1 = \tilde{c}_{i-1} - k_i + 1$.

Then we define four operators $\mathsf{G}(\cdot)$, $\mathsf{Q}(\cdot)$, $\mathsf{up}(\cdot)$ and $\mathsf{vec}(\cdot)$, which are useful for explaining the following analysis.

**Operation** $\mathsf{G}(\cdot)$: It maps an arbitrary vector $\boldsymbol{z} \in \mathbb{R}^d$ into a diagonal matrix $\mathsf{G}(\boldsymbol{z}) \in \mathbb{R}^{d\times d}$ with its $i$-th diagonal entry equal to $\sigma_1(\boldsymbol{z}_i)(1-\sigma_1(\boldsymbol{z}_i))$ in which $\boldsymbol{z}_i$ denotes the $i$-th entry of $\boldsymbol{z}$.

**Operation** $\mathsf{Q}(\cdot)$: it maps a vector $\boldsymbol{z} \in \mathbb{R}^d$ into a matrix of size $d^2 \times d$ whose $((i-1)d+i, i)$ $(i = 1, \cdots, d)$ entry equal to $\sigma_1(\boldsymbol{z}_i)(1-\sigma_1(\boldsymbol{z}_i))(1-2\sigma_1(\boldsymbol{z}_i))$ and rest entries are all 0.

**Operation** $\mathsf{up}(\cdot)$: $\mathsf{up}(\boldsymbol{M})$ represents conducting upsampling on $\boldsymbol{M} \in \mathbb{R}^{s\times t\times q}$. Let $\boldsymbol{N} = \mathsf{up}(\boldsymbol{M}) \in \mathbb{R}^{ps\times pt\times q}$. Specifically, for each slice $\boldsymbol{N}(:,:,i)$ $(i = 1, \cdots, q)$, we have $\boldsymbol{N}(:,:,i) = \mathsf{up}(\boldsymbol{M}(:,:,i))$. It actually upsamples each entry $\boldsymbol{M}(g,h,i)$ into a matrix of $p^2$ same entries $\frac{1}{p^2}\boldsymbol{M}(g,h,i)$.

**Operation** $\mathsf{vec}(\cdot)$: For a matrix $\boldsymbol{A} \in \mathbb{R}^{s\times t}$, $\mathsf{vec}(\boldsymbol{A})$ is defined as $\mathsf{vec}(\boldsymbol{A}) = (\boldsymbol{A}(:,1); \cdots; \boldsymbol{A}(:,t)) \in \mathbb{R}^{st}$ that vectorizes $\boldsymbol{A} \in \mathbb{R}^{s\times t}$ along its columns. If $\boldsymbol{A} \in \mathbb{R}^{t\times s\times q}$ is a 3-way tensor, $\mathsf{vec}(\boldsymbol{A}) = [\mathsf{vec}(\boldsymbol{A}(:,:,1)); \mathsf{vec}(\boldsymbol{A}(:,:,2)), \cdots, \mathsf{vec}(\boldsymbol{A}(:,:,q))] \in \mathbb{R}^{stq}$.

### B.2 Auxiliary Lemmas

We first introduce Lemmas 2 and 3 which are respectively used for bounding the $\ell_2$-norm of a vector and the operation norm of a matrix. Then we introduce Lemmas 4 and 5 which discuss the topology of functions. In Lemma 6, we introduce the Hoeffding's inequality which provides an upper bound on the probability that the sum of independent random variables deviates from its expected value. In Lemma 7, we provide the covering number of an $\epsilon$-net for a low-rank matrix. Finally, several commonly used inequalities are presented in Lemma 8 for our analysis.

**Lemma 2.** *(Vershynin, 2012) For any vector $\boldsymbol{x} \in \mathbb{R}^d$, its $\ell_2$-norm can be bounded as*

$$\|\boldsymbol{x}\|_2 \leq \frac{1}{1-\epsilon} \sup_{\boldsymbol{\lambda}\in\boldsymbol{\lambda}_\epsilon} \langle \boldsymbol{\lambda}, \boldsymbol{x} \rangle .$$

*where $\boldsymbol{\lambda}_\epsilon = \{\boldsymbol{\lambda}_1, \ldots, \boldsymbol{\lambda}_{k_w}\}$ be an $\epsilon$-covering net of $\mathsf{B}^d(1)$.*

**Lemma 3.** *(Vershynin, 2012) For any symmetric matrix $\boldsymbol{X} \in \mathbb{R}^{d\times d}$, its operator norm can be bounded as*

$$\|\boldsymbol{X}\|_{op} \leq \frac{1}{1-2\epsilon} \sup_{\boldsymbol{\lambda}\in\boldsymbol{\lambda}_\epsilon} |\langle \boldsymbol{\lambda}, \boldsymbol{X}\boldsymbol{\lambda} \rangle| .$$

*where $\boldsymbol{\lambda}_\epsilon = \{\boldsymbol{\lambda}_1, \ldots, \boldsymbol{\lambda}_{k_w}\}$ be an $\epsilon$-covering net of $\mathsf{B}^d(1)$.*

**Lemma 4.** *(Mei et al., 2017) Let $D \subseteq \mathbb{R}^d$ be a compact set with a $C^2$ boundary $\partial D$, and $f, g : A \to \mathbb{R}$ be $C^2$ functions defined on an open set $A$, with $D \subseteq A$. Assume that for all $\mathbf{w} \in \partial D$ and all $t \in [0,1]$, $t\nabla f(\mathbf{w}) + (1-t)\nabla g(\mathbf{w}) \neq \mathbf{0}$. Finally, assume that the Hessian $\nabla^2 f(\mathbf{w})$ is non-degenerate and has index equal to $r$ for all $\mathbf{w} \in D$. Then the following properties hold:*

*(1) If $g$ has no critical point in $D$, then $f$ has no critical point in $D$.*

*(2) If $g$ has a unique critical point $\mathbf{w}$ in $D$ that is non-degenerate with an index of $r$, then $f$ also has a unique critical point $\mathbf{w}'$ in $D$ with the index equal to $r$.*

**Lemma 5.** *(Mei et al., 2017) Suppose that $F(\mathbf{w}) : \Theta \to \mathbb{R}$ is a $C^2$ function where $\mathbf{w} \in \Theta$. Assume that $\{\mathbf{w}_{(1)}, \ldots, \mathbf{w}_{(m)}\}$ is its non-degenerate critical points and let $D = \{\mathbf{w} \in \Theta : \|\nabla F(\mathbf{w})\|_2 < \epsilon \text{ and } \inf_i |\lambda_i(\nabla^2 F(\mathbf{w}))| \geq \zeta\}$. Then $D$ can be decomposed into (at most) countably components, with each component containing either exactly one critical point, or no critical point. Concretely, there exist disjoint open sets $\{D_k\}_{k\in\mathbb{N}}$, with $D_k$ possibly empty for $k \geq m+1$, such that*

$$D = \cup_{k=1}^{\infty} D_k .$$

*Furthermore, $\mathbf{w}_{(k)} \in D_k$ for $1 \leq k \leq m$ and each $D_i$, $k \geq m+1$ contains no stationary points.*

**Lemma 6.** *(Hoeffding, 1963) Let $x_1, \cdots, x_n$ be independent random variables where $x_i$ is bounded by the interval $[a_i, b_i]$. Suppose $s_n = \frac{1}{n}\sum_{i=1}^n x_i$, then the following properties hold:*

$$\mathbb{P}\left(s_n - \mathbb{E}s_n \geq t\right) \leq \exp\left(-\frac{2n^2t^2}{\sum_{i=1}^n (b_i - a_i)^2}\right).$$

*where $t$ is an arbitrary positive value.*

**Lemma 7.** *(Candes & Plan, 2009) Let $\mathbb{S}_r = \{\boldsymbol{X} \in \mathbb{R}^{n_1\times n_2} : \mathsf{rank}(\boldsymbol{X}) \leq r, \|\boldsymbol{X}\|_F = b\}$. Then there exists an $\epsilon$-net $\widetilde{\mathbb{S}}_r$ for the Frobenius norm obeying*

$$|\widetilde{\mathbb{S}}_r| \leq \left(\frac{9b}{\epsilon}\right)^{r(n_1+n_2+1)}.$$

Then we give a lemma to summarize the properties of $\mathsf{G}(\cdot)$, $\mathsf{Q}(\cdot)$ and $\mathsf{up}(\cdot)$ defined in Section B.1, the convolutional operation $\circledast$ defined in manuscript.

**Lemma 8.** *For $\mathsf{G}(\cdot)$, $\mathsf{Q}(\cdot)$ and $\mathsf{up}(\cdot)$ defined in Section B.1, the convolutional operation $\circledast$ defined in manuscript, we have the following properties:*

*(1) For arbitrary vector $\boldsymbol{z}$, and arbitrary matrices $\boldsymbol{M}$ and $\boldsymbol{N}$ of proper sizes, we have*

$$\|\mathsf{G}(\boldsymbol{z})\boldsymbol{M}\|_F^2 \leq \frac{1}{16}\|\boldsymbol{M}\|_F^2 \quad \text{and} \quad \|\boldsymbol{N}\mathsf{G}(\boldsymbol{z})\|_F^2 \leq \frac{1}{16}\|\boldsymbol{N}\|_F^2.$$

*(2) For arbitrary vector $\boldsymbol{z}$, and arbitrary matrices $\boldsymbol{M}$ and $\boldsymbol{N}$ of proper sizes, we have*

$$\|\mathsf{Q}(\boldsymbol{z})\boldsymbol{M}\|_F^2 \leq \frac{2^6}{3^8}\|\boldsymbol{M}\|_F^2 \quad \text{and} \quad \|\boldsymbol{N}\mathsf{Q}(\boldsymbol{z})\|_F^2 \leq \frac{2^6}{3^8}\|\boldsymbol{N}\|_F^2.$$

*(3) For any tensor $\boldsymbol{M} \in \mathbb{R}^{s\times t\times q}$, we have*

$$\|\mathsf{up}(\boldsymbol{M})\|_F^2 \leq \frac{1}{p^2}\|\boldsymbol{M}\|_F^2.$$

*(4) For any kernel $\boldsymbol{W} \in \mathbb{R}^{k_i\times k_i\times d_i}$ and $\widetilde{\boldsymbol{\delta}}_{i+1} \in \mathbb{R}^{(\tilde{r}_{i-1}-k_i+1)\times(\tilde{c}_{i-1}-k_i+1)\times d_i}$ defined in Sec. B.1, then we have*

$$\|\widetilde{\boldsymbol{\delta}}_{i+1}\widetilde{\circledast}\boldsymbol{W}\|_F^2 \leq (k_i - s_i + 1)^2\|\boldsymbol{W}\|_F^2\|\widetilde{\boldsymbol{\delta}}_{i+1}\|_F^2.$$

*(5) For softmax activation function $\sigma_2$, we can bound the norm of difference between output $\boldsymbol{v}$ and its corresponding ont-hot label as*

$$0 \leq \|\boldsymbol{v} - \boldsymbol{y}\|_2^2 \leq 2.$$

It should be pointed out that we defer the proof of Lemma 8 to Sec. E.

# C Technical Lemmas and Their Proofs

Here we present the key lemmas and theorems for proving our desired results. For brevity, in this document we use $f(\boldsymbol{w}, \boldsymbol{D})$ to denote $f(g(\boldsymbol{w}; \boldsymbol{D}), \boldsymbol{y})$ in the manuscript.

**Lemma 9.** *Suppose that the activation function $\sigma_1$ is sigmoid and $\sigma_2$ is softmax, and the loss function $f(\boldsymbol{w}, \boldsymbol{D})$ is squared loss. Then the gradient of $f(\boldsymbol{w}, \boldsymbol{D})$ with respect to $\boldsymbol{w}_{(j)}$ can be written as*

$$\nabla_{\boldsymbol{W}_{(l+1)}} f(\boldsymbol{w}, \boldsymbol{D}) = \boldsymbol{S}(\boldsymbol{v} - \boldsymbol{y})\boldsymbol{z}_{(l)}^T \in \mathbb{R}^{d_{l+1}\times \tilde{r}_l\tilde{c}_l d_l},$$
$$\nabla_{\boldsymbol{W}_{(i)}^{k,j}} f(\boldsymbol{w}, \boldsymbol{D}) = \boldsymbol{Z}_{(i-1)}^j \widetilde{\circledast}\widetilde{\boldsymbol{\delta}}_i^k \in \mathbb{R}^{k_i\times k_i},\ (j = 1, \cdots, d_{i-1};\ k = 1, \cdots, d_i;\ i = 1, \cdots, l),$$

*where $\boldsymbol{\delta}_i^k$ is the k-slice (i.e. $\boldsymbol{\delta}_i(:,:,k)$) of $\boldsymbol{\delta}_i$ which is defined as*

$$\boldsymbol{\delta}_i = \nabla_{\boldsymbol{X}_{(i)}} f(\boldsymbol{w}, \boldsymbol{D}) \in \mathbb{R}^{r_i \times c_i \times d_i}, \quad (i = 1, \cdots, l),$$

*and further satisfies*

$$\boldsymbol{\delta}_i^j = \textit{up}\left(\sum_{k=1}^{d_{i+1}} \widetilde{\boldsymbol{\delta}}_{i+1}^k \widetilde{\circledast} \widehat{\boldsymbol{W}}_{(i+1)}^{k,j}\right) \odot \sigma_1'(\boldsymbol{X}_{(i)}^j) \in \mathbb{R}^{r_i \times c_i}, \quad (j = 1, \cdots, d_i;\ i = 1, \cdots, l-1),$$

*where the matrix $\widehat{\boldsymbol{W}}_{(i+1)}^{k,j} \in \mathbb{R}^{k_{i+1} \times k_{i+1}}$ is obtained by rotating $\boldsymbol{W}_{(i+1)}^{k,j}$ with 180 degrees. Also, $\boldsymbol{\delta}_l$ is computed as follows:*

$$\nabla_{\boldsymbol{u}} f(\boldsymbol{w}, \boldsymbol{D}) = \boldsymbol{S}(\boldsymbol{v} - \boldsymbol{y}) \in \mathbb{R}^{d_{l+1}}, \qquad \nabla_{\boldsymbol{z}_{(l)}} f(\boldsymbol{w}, \boldsymbol{D}) = (\boldsymbol{W}_{(l+1)})^T \boldsymbol{S}(\boldsymbol{v} - \boldsymbol{y}) \in \mathbb{R}^{\tilde{r}_l \tilde{c}_l d_l},$$

$$\nabla_{\boldsymbol{Z}_{(l)}} f(\boldsymbol{w}, \boldsymbol{D}) = \textit{reshape}\left(\nabla_{\boldsymbol{z}_{(l)}} f(\boldsymbol{w}, \boldsymbol{D})\right) \in \mathbb{R}^{\tilde{r}_l \times \tilde{c}_l \times d_l}, \ \boldsymbol{\delta}_l = \sigma_1'(\boldsymbol{X}_{(l)}) \odot \textit{up}\left(\frac{\partial f(\boldsymbol{w}, \boldsymbol{D})}{\partial \boldsymbol{Z}_{(l)}}\right) \in \mathbb{R}^{r_i \times c_i \times d_i}.$$

*where $\boldsymbol{S} = \boldsymbol{G}(\boldsymbol{u})$.*

*Proof.* We use chain rule to compute the gradient of $f(\boldsymbol{w}, \boldsymbol{D})$ with respect to $\boldsymbol{Z}_{(i)}$. We first compute several basis gradient. According to the relationship between $\boldsymbol{X}_{(i)}, \boldsymbol{Y}_{(i)}, \boldsymbol{Z}_{(i)}$ and $\boldsymbol{f}(\boldsymbol{w}, \boldsymbol{D})$, we have

$$\begin{aligned}
&\nabla_{\boldsymbol{u}} f(\boldsymbol{w}, \boldsymbol{D}) = \boldsymbol{S}(\boldsymbol{v} - \boldsymbol{y}) \in \mathbb{R}^{d_{l+1}},\\
&\nabla_{\boldsymbol{z}_{(l)}} f(\boldsymbol{w}, \boldsymbol{D}) = (\boldsymbol{W}_{(l+1)})^T \boldsymbol{S}(\boldsymbol{v} - \boldsymbol{y}) \in \mathbb{R}^{\tilde{r}_l \tilde{c}_l d_l},\\
&\nabla_{\boldsymbol{Z}_{(l)}} f(\boldsymbol{w}, \boldsymbol{D}) = \mathrm{reshape}\left(\nabla_{\boldsymbol{z}_{(l)}} f(\boldsymbol{w}, \boldsymbol{D})\right) \in \mathbb{R}^{\tilde{r}_l \times \tilde{c}_l \times d_l},\\
&\nabla_{\boldsymbol{X}_{(l)}} f(\boldsymbol{w}, \boldsymbol{D}) = \frac{\partial \boldsymbol{Y}_{(l)}}{\partial \boldsymbol{X}_{(l)}} \frac{\partial \boldsymbol{Z}_{(l)}}{\partial \boldsymbol{Y}_{(l)}} \frac{\partial f(\boldsymbol{w}, \boldsymbol{D})}{\partial \boldsymbol{Z}_{(l)}} = \sigma_1'(\boldsymbol{X}_{(l)}) \odot \mathrm{up}\left(\frac{\partial f(\boldsymbol{w}, \boldsymbol{D})}{\partial \boldsymbol{Z}_{(l)}}\right) \triangleq \boldsymbol{\delta}_l \in \mathbb{R}^{r_i \times c_i \times d_i}.
\end{aligned}$$

Then we can further obtain

$$\boldsymbol{\delta}_i^j = \mathrm{up}\left(\sum_{k=1}^{d_{i+1}} \widetilde{\boldsymbol{\delta}}_{i+1}^k \widetilde{\circledast} \widehat{\boldsymbol{W}}_{(i+1)}^{k,j}\right) \odot \sigma_1'(\boldsymbol{X}_{(i)}^j) \in \mathbb{R}^{r_i \times c_i}, \quad (j = 1, \cdots, d_i;\ i = 1, \cdots, l-1).$$

where $\widehat{\boldsymbol{W}}_{(i)}^{k,j}$ denotes the $j$-th slice $\widehat{\boldsymbol{W}}_{(i)}^{k}(:,:j)$ of $\widehat{\boldsymbol{W}}_{(i)}^{k}$. Note, we clockwise rotate the matrix $\boldsymbol{W}_{(i+1)}^{k,j}$ by 180 degrees to obtain $\widehat{\boldsymbol{W}}_{(i+1)}^{k,j}$. Finally, we can compute the gradient *w.r.t.* $\boldsymbol{W}_{(l+1)}$ and $\boldsymbol{W}^{(i)}$ $(i = 1, \cdots, l)$:

$$\begin{aligned}
\nabla_{\boldsymbol{W}_{(l+1)}} f(\boldsymbol{w}, \boldsymbol{D}) =& \boldsymbol{S}(\boldsymbol{v} - \boldsymbol{y}) \boldsymbol{z}_{(l)}^T \in \mathbb{R}^{d_{l+1} \times \tilde{r}_l \tilde{c}_l d_l},\\
\nabla_{\boldsymbol{W}_{(i)}^{k,j}} f(\boldsymbol{w}, \boldsymbol{D}) =& \boldsymbol{X}_{(i-1)}^j \widetilde{\circledast} \widetilde{\boldsymbol{\delta}}_i^k \in \mathbb{R}^{k_i \times k_i}, \ (j = 1, \cdots, d_{i-1};\ k = 1, \cdots, d_i;\ i = 1, \cdots, l).
\end{aligned}$$

The proof is completed. ☐

**Lemma 10.** *Suppose that the activation function $\sigma_1$ is sigmoid and $\sigma_2$ is softmax, and the loss function $f(\boldsymbol{w}, \boldsymbol{D})$ is squared loss. Then the gradient of $f(\boldsymbol{w}, \boldsymbol{D})$ with respect to $\boldsymbol{w}_{(j)}$ can be written as*

$$\|\boldsymbol{\delta}_l\|_F^2 \leq \frac{\vartheta}{16p^2} {b_{l+1}}^2, \quad \|\boldsymbol{\delta}_i\|_F^2 \leq \frac{d_{i+1} {b_{i+1}}^2 (k_{i+1} - s_{i+1} + 1)^2}{16p^2} \left\|\boldsymbol{\delta}_{i+1}\right\|_F^2, \quad \|\boldsymbol{\delta}_i\|_F^2 \leq \frac{\vartheta {b_{l+1}}^2}{16p^2} \prod_{s=i+1}^{l} \frac{d_s {b_s}^2 (k_s - s_s + 1)^2}{16p^2},$$

*where $\vartheta = 1/8$.*

*Proof.* We first bound $\boldsymbol{\delta}_l$. By Lemma 9, we have

$$\begin{aligned}
\|\boldsymbol{\delta}_l\|_F^2 =& \left\|\sigma_1'(\boldsymbol{X}_{(l)}) \odot \mathrm{up}\left(\frac{\partial f(\boldsymbol{w}, \boldsymbol{D})}{\partial \boldsymbol{Z}_{(l)}}\right)\right\|_F^2 \overset{①}{\leq} \frac{1}{16p^2} \left\|\frac{\partial f(\boldsymbol{w}, \boldsymbol{D})}{\partial \boldsymbol{Z}_{(l)}}\right\|_F^2 = \frac{1}{16p^2} \left\|\frac{\partial f(\boldsymbol{w}, \boldsymbol{D})}{\partial \boldsymbol{z}_{(l)}}\right\|_2^2\\
\leq& \frac{1}{16p^2} \left\|(\boldsymbol{W}_{(l+1)})^T \boldsymbol{S}(\boldsymbol{v} - \boldsymbol{y})\right\|_2^2 \overset{②}{\leq} \frac{\vartheta}{16p^2} {b_{l+1}}^2,
\end{aligned}$$

where ① holds since the values of the entries in the tensor $\sigma_1'(\boldsymbol{X}_{(l)}) \in \mathbb{R}^{r_l\times c_l\times d_l}$ belong to $[0, 1/4]$, and the $\mathsf{up}(\cdot)$ operation does repeat one entry into $p^2$ entries but the entry value becomes $1/p^2$ of the original entry value. ② holds since we have $\|\boldsymbol{S}(\boldsymbol{v}-\boldsymbol{y})\|_2^2 \leq 1/8$ in Lemma 8.

Also by Lemma 9, we can further bound $\|\boldsymbol{\delta}_i^j\|_F^2$ as follows:

$$\begin{aligned}
\|\boldsymbol{\delta}_i\|_F^2 =& \sum_{j=1}^{d_i}\left\|\boldsymbol{\delta}_i^j\right\|_F^2 = \sum_{j=1}^{d_i}\left\|\mathsf{up}\left(\sum_{k=1}^{d_{i+1}}\widetilde{\boldsymbol{\delta}}_{i+1}^k \widetilde{\circledast}\widehat{\boldsymbol{W}}_{(i+1)}^{k,j}\right)\odot\sigma_1'(\boldsymbol{X}_{(i)}^j)\right\|_F^2 \leq \frac{1}{16p^2}\sum_{j=1}^{d_i}\left\|\sum_{k=1}^{d_{i+1}}\widetilde{\boldsymbol{\delta}}_{i+1}^k \widetilde{\circledast}\widehat{\boldsymbol{W}}_{(i+1)}^{k,j}\right\|_F^2\\
\leq& \frac{d_{i+1}}{16p^2}\sum_{j=1}^{d_i}\sum_{k=1}^{d_{i+1}}\left\|\widetilde{\boldsymbol{\delta}}_{i+1}^k \widetilde{\circledast}\widehat{\boldsymbol{W}}_{(i+1)}^{k,j}\right\|_F^2 \leq \frac{d_{i+1}(k_{i+1}-s_{i+1}+1)^2}{16p^2}\sum_{j=1}^{d_i}\sum_{k=1}^{d_{i+1}}\left\|\widetilde{\boldsymbol{\delta}}_{i+1}^k\right\|_F^2\left\|\widehat{\boldsymbol{W}}_{(i+1)}^{k,j}\right\|_F^2\\
\overset{①}{=}& \frac{d_{i+1}(k_{i+1}-s_{i+1}+1)^2}{16p^2}\sum_{k=1}^{d_{i+1}}\left\|\boldsymbol{\delta}_{i+1}^k\right\|_F^2\left\|\boldsymbol{W}_{(i+1)}^{k}\right\|_F^2 \leq \frac{d_{i+1}{b_{i+1}}^2(k_{i+1}-s_{i+1}+1)^2}{16p^2}\left\|\boldsymbol{\delta}_{i+1}\right\|_F^2,
\end{aligned}$$

where $\widehat{\boldsymbol{W}}_{(i+1)}^{k,j}$ denotes the $j$-th slice $\widehat{\boldsymbol{W}}_{(i+1)}^{k}(:,:,j)$ of the tensor $\widehat{\boldsymbol{W}}_{(i+1)}^{k}$. ① holds since we rotate the matrix $\boldsymbol{W}_{(i+1)}^{k,j}$ by 180 degrees to obtain $\widehat{\boldsymbol{W}}_{(i+1)}^{k}$, indicating $\|\boldsymbol{W}_{(i+1)}^{k,j}\|_F^2 = \|\widehat{\boldsymbol{W}}_{(i+1)}^{k,j}\|_F^2$ and $\sum_{j=1}^{d_i}\left\|\boldsymbol{W}_{(i+1)}^{k,j}\right\|_F^2 = \left\|\boldsymbol{W}_{(i+1)}^{k}\right\|_F^2 \leq {b_{i+1}}^2$. Accordingly, the above inequality gives

$$\begin{aligned}
\|\boldsymbol{\delta}_i\|_F^2 \leq& \frac{d_{i+1}{b_{i+1}}^2(k_{i+1}-s_{i+1}+1)^2}{16p^2}\left\|\boldsymbol{\delta}_{i+1}\right\|_F^2 \leq \cdots \leq \|\boldsymbol{\delta}_l\|_F^2\prod_{s=i+1}^{l}\frac{d_s{b_s}^2(k_s-s_s+1)^2}{16p^2}\\
\leq& \frac{\vartheta {b_{l+1}}^2}{16p^2}\prod_{s=i+1}^{l}\frac{d_s{b_s}^2(k_s-s_s+1)^2}{16p^2}.
\end{aligned}$$

The proof is completed. □

**Lemma 11.** *Suppose that the activation function $\sigma_1$ is sigmoid and $\sigma_2$ is softmax, and the loss function $f(\boldsymbol{w},\boldsymbol{D})$ is squared loss. Then the gradient of $f(\boldsymbol{w},\boldsymbol{D})$ with respect to $\boldsymbol{W}^{(l+1)}$ and $\boldsymbol{w}$ can be respectively bounded as follows:*

$$\left\|\nabla_{\boldsymbol{W}_{(l+1)}} f(\boldsymbol{w},\boldsymbol{D})\right\|_F^2 \leq \vartheta\tilde{r}_l\tilde{c}_l d_l \quad \textit{and} \quad \|\nabla_{\boldsymbol{w}} f(\boldsymbol{w},\boldsymbol{D})\|_2^2 \leq \beta^2,$$

*where $\vartheta = 1/8$ and $\beta \triangleq \left[\vartheta\tilde{r}_l\tilde{c}_l d_l + \sum_{i=1}^{l}\frac{\vartheta {b_{l+1}}^2 d_{i-1}}{p^2{b_i}^2 d_i} r_{i-1}c_{i-1}\prod_{s=i}^{l}\frac{d_s{b_s}^2(k_s-s_s+1)^2}{16p^2}\right]^{1/2}$.*

*Proof.* By utilizing Lemma 10, we can bound

$$\begin{aligned}
\sum_{i=1}^{l}\left\|\nabla_{\boldsymbol{w}_{(i)}} f(\boldsymbol{w},\boldsymbol{D})\right\|_F^2 =& \sum_{i=1}^{l}\sum_{k=1}^{d_i}\sum_{j=1}^{d_{i-1}}\left\|\nabla_{\boldsymbol{W}_{(i)}^{k,j}} f(\boldsymbol{w},\boldsymbol{D})\right\|_F^2 = \sum_{i=1}^{l}\sum_{k=1}^{d_i}\sum_{j=1}^{d_{i-1}}\left\|\boldsymbol{Z}_{(i-1)}^j\widetilde{\circledast}\boldsymbol{\delta}_i^k\right\|_F^2\\
\overset{①}{\leq}& \sum_{i=1}^{l}\sum_{k=1}^{d_i}\sum_{j=1}^{d_{i-1}}(k_i-s_i+1)^2\left\|\boldsymbol{Z}_{(i-1)}^j\right\|_F^2\left\|\boldsymbol{\delta}_i^k\right\|_F^2\\
\overset{②}{\leq}& \sum_{i=1}^{l}\sum_{k=1}^{d_i}\sum_{j=1}^{d_{i-1}}\tilde{r}_{i-1}\tilde{c}_{i-1}(k_i-s_i+1)^2\left\|\boldsymbol{\delta}_i^k\right\|_F^2\\
\leq& \sum_{i=1}^{l}\tilde{r}_{i-1}\tilde{c}_{i-1}d_{i-1}(k_i-s_i+1)^2\left\|\boldsymbol{\delta}_i\right\|_F^2\\
\leq& \sum_{i=1}^{l}\tilde{r}_{i-1}\tilde{c}_{i-1}d_{i-1}(k_i-s_i+1)^2\frac{\vartheta {b_{l+1}}^2}{16p^2}\prod_{s=i+1}^{l}\frac{d_s{b_s}^2(k_s-s_s+1)^2}{16p^2}\\
=& \sum_{i=1}^{l}\frac{\vartheta {b_{l+1}}^2 d_{i-1}}{p^2{b_i}^2 d_i}r_{i-1}c_{i-1}\prod_{s=i}^{l}\frac{d_s{b_s}^2(k_s-s_s+1)^2}{16p^2},
\end{aligned}$$

where ① holds since $\left\|\boldsymbol{Z}_{(i-1)}^{j}\widetilde{\circledast}\boldsymbol{\delta}_{i}^{k}\right\|_F^2 \leq (k_i-s_i+1)^2\left\|\boldsymbol{Z}_{(i-1)}^{j}\right\|_F^2\left\|\boldsymbol{\delta}_{i}^{k}\right\|_F^2$; and ② holds since the values of entries in $\boldsymbol{Z}_{(i-1)}^{j}$ belong to $[0,1]$.

On the other hand, we can bound

$$\left\|\nabla_{\boldsymbol{W}_{(l+1)}} f(\boldsymbol{w},\boldsymbol{D})\right\|_F^2 = \left\|\boldsymbol{S}(\boldsymbol{v}-\boldsymbol{y})\boldsymbol{z}_{(l)}^T\right\|_F^2 \leq \vartheta\tilde{r}_l\tilde{c}_l d_l.$$

So we can bound the $\ell_2$ norm of the gradient as follows:

$$\begin{aligned}\left\|\nabla_{\boldsymbol{w}} f(\boldsymbol{w},\boldsymbol{D})\right\|_F^2 =& \left\|\nabla_{\boldsymbol{W}_{(l+1)}} f(\boldsymbol{w},\boldsymbol{x})\right\|_F^2 + \sum_{i=1}^{l}\left\|\nabla_{\boldsymbol{w}_{(i)}} f(\boldsymbol{w},\boldsymbol{D})\right\|_F^2\\ \leq&\vartheta\tilde{r}_l\tilde{c}_l d_l + \sum_{i=1}^{l}\frac{\vartheta {b_{l+1}}^2 d_{i-1}}{p^2 {b_i}^2 d_i} r_{i-1}c_{i-1}\prod_{s=i}^{l}\frac{d_s {b_s}^2(k_s-s_s+1)^2}{16p^2}.\end{aligned}$$

The proof is completed. ∎

**Lemma 12.** *Suppose that the activation function $\sigma_1$ is sigmoid and $\sigma_2$ is softmax, and the loss function $f(\boldsymbol{w},\boldsymbol{D})$ is squared loss. Then for both cases, the gradient of $\boldsymbol{x}_{(i)}$ with respect to $\boldsymbol{w}_{(j)}$ can be bounded as follows:*

$$\left\|\frac{\partial \mathit{vec}\left(\boldsymbol{X}_{(i)}\right)}{\partial \boldsymbol{w}_{(j)}}\right\|_F^2 = \left\|\frac{\partial \boldsymbol{x}_{(i)}}{\partial \boldsymbol{w}_{(j)}}\right\|_F^2 \leq d_i r_i c_i \tilde{r}_{j-1}\tilde{c}_{j-1}d_{j-1}(k_j-s_j+1)^2\prod_{s=j+1}^{i}\frac{d_s {b_s}^2(k_s-s_s+1)^2}{16p^2}$$

*and*

$$\max_s\left\|\frac{\partial \mathit{vec}\left(\boldsymbol{X}_{(i)}^s\right)}{\partial \boldsymbol{w}_{(j)}}\right\|_F^2 = \max_s\left\|\frac{\partial \boldsymbol{x}_{(i)}^s}{\partial \boldsymbol{w}_{(j)}}\right\|_F^2 \leq r_i c_i \tilde{r}_{j-1}\tilde{c}_{j-1}d_{j-1}(k_j-s_j+1)^2\prod_{s=j+1}^{i}\frac{d_s {b_s}^2(k_s-s_s+1)^2}{16p^2}.$$

*Proof.* For brevity, let $\boldsymbol{X}_{(i)}^k(s,t)$ denotes the $(s,t)$-th entry in the matrix $\boldsymbol{X}_{(i)}^k\in\mathbb{R}^{r_i\times c_i}$. We also let $\boldsymbol{\phi}_{(i,m)} = \frac{\partial \boldsymbol{X}_{(i)}^k(s,t)}{\partial \boldsymbol{X}_{(m)}} \in \mathbb{R}^{r_m\times c_m\times d_m}$. So similar to Lemma 9, we have

$$\boldsymbol{\phi}_{(i,m)}^q = \mathrm{up}\left(\sum_{k=1}^{d_{m+1}}\widetilde{\boldsymbol{\phi}}_{(i,m+1)}^k\widetilde{\circledast}\widehat{\boldsymbol{W}}_{(m+1)}^{k,q}\right)\odot\sigma_1'(\boldsymbol{X}_{(m)}^q)\in\mathbb{R}^{r_m\times c_m},\quad (q=1,\cdots,d_m),$$

where the matrix $\widehat{\boldsymbol{W}}_{(m+1)}^{k,q}\in\mathbb{R}^{k_{m+1}\times k_{m+1}}$ is obtained by rotating $\boldsymbol{W}_{(m+1)}^{k,q}$ with 180 degrees. Then according to the relationship between $\boldsymbol{X}_{(j)}$ and $\boldsymbol{W}_{(j)}$, we can compute

$$\frac{\partial \boldsymbol{X}_{(i)}^k(s,t)}{\partial \boldsymbol{W}_{(j)}^{g,h}} = \boldsymbol{Z}_{(j-1)}^h\widetilde{\circledast}\boldsymbol{\phi}_{(i,j)}^g\in\mathbb{R}^{k_j\times k_j},\ (h=1,\cdots,d_{j-1};\ g=1,\cdots,d_j).$$

Therefore, we can further obtain

$$\begin{aligned}\left\|\frac{\partial \boldsymbol{X}_{(i)}^k(s,t)}{\partial \boldsymbol{w}_{(j)}}\right\|_F^2 =& \sum_{g=1}^{d_j}\sum_{h=1}^{d_{j-1}}\left\|\frac{\partial \boldsymbol{X}_{(i)}^k(s,t)}{\partial \boldsymbol{W}_{(j)}^{g,h}}\right\|_F^2 = \sum_{g=1}^{d_j}\sum_{h=1}^{d_{j-1}}\left\|\frac{\partial \boldsymbol{X}_{(i)}^k(s,t)}{\partial \boldsymbol{W}_{(j)}^{g,h}}\right\|_F^2 = \sum_{g=1}^{d_j}\sum_{h=1}^{d_{j-1}}\left\|\boldsymbol{Z}_{(j-1)}^h\widetilde{\circledast}\boldsymbol{\phi}_{(i,j)}^g\right\|_F^2\\ \leq&\sum_{g=1}^{d_j}\sum_{h=1}^{d_{j-1}}(k_j-s_j+1)^2\left\|\boldsymbol{Z}_{(j-1)}^h\right\|_F^2\left\|\boldsymbol{\phi}_{(i,j)}^g\right\|_F^2 \leq (k_j-s_j+1)^2\left\|\boldsymbol{Z}_{(j-1)}\right\|_F^2\left\|\boldsymbol{\phi}_{(i,j)}\right\|_F^2\\ \leq&\tilde{r}_{j-1}\tilde{c}_{j-1}d_{j-1}(k_j-s_j+1)^2\left\|\boldsymbol{\phi}_{(i,j)}\right\|_F^2.\end{aligned}$$

On the other hand, by Lemma 9, we can further bound $\|\boldsymbol{\phi}_{(i,j)}\|_F^2$ as follows:

$$\begin{aligned}
\left\|\boldsymbol{\phi}_{(i,m)}\right\|_F^2 =& \sum_{q=1}^{d_m}\left\|\boldsymbol{\phi}_{(i,m)}^q\right\|_F^2 = \sum_{q=1}^{d_m}\left\|\mathsf{up}\left(\sum_{k=1}^{d_{m+1}}\widetilde{\boldsymbol{\phi}}_{(i,m+1)}^k\widetilde{\circledast}\widehat{\boldsymbol{W}}_{(m+1)}^{k,q}\right)\odot\sigma_1'(\boldsymbol{X}_{(m)}^q)\right\|_F^2\\
\leq&\frac{1}{16p^2}\sum_{q=1}^{d_m}\left\|\sum_{k=1}^{d_{m+1}}\widetilde{\boldsymbol{\phi}}_{(i,m+1)}^k\widetilde{\circledast}\widehat{\boldsymbol{W}}_{(m+1)}^{k,q}\right\|_F^2 \leq \frac{d_{m+1}}{16p^2}\sum_{q=1}^{d_m}\sum_{k=1}^{d_{m+1}}\left\|\widetilde{\boldsymbol{\phi}}_{(i,m+1)}^k\widetilde{\circledast}\widehat{\boldsymbol{W}}_{(m+1)}^{k,q}\right\|_F^2\\
\leq&\frac{d_{m+1}(k_{m+1}-s_{m+1}+1)^2}{16p^2}\sum_{q=1}^{d_m}\sum_{k=1}^{d_{m+1}}\left\|\widetilde{\boldsymbol{\phi}}_{(i,m+1)}^k\right\|_F^2\left\|\widehat{\boldsymbol{W}}_{(m+1)}^{k,q}\right\|_F^2\\
\overset{①}{=}&\frac{d_{m+1}(k_{m+1}-s_{m+1}+1)^2}{16p^2}\sum_{k=1}^{d_{m+1}}\left\|\widetilde{\boldsymbol{\phi}}_{(i,m+1)}^k\right\|_F^2\left\|\widehat{\boldsymbol{W}}_{(m+1)}^{k}\right\|_F^2\\
\leq&\frac{d_{m+1}{b_{m+1}}^2(k_{m+1}-s_{m+1}+1)^2}{16p^2}\left\|\boldsymbol{\phi}_{(i,m+1)}\right\|_F^2,
\end{aligned}$$

where ① holds since $\|\boldsymbol{W}_{(m+1)}^{k,q}\|_F^2 = \|\widehat{\boldsymbol{W}}_{(m+1)}^{k,q}\|_F^2$. It further yields

$$\left\|\boldsymbol{\phi}_{(i,m)}\right\|_F^2 \leq \left\|\boldsymbol{\phi}_{(i,i)}\right\|_F^2\prod_{s=m+1}^{i}\frac{d_s{b_s}^2(k_s-s_s+1)^2}{16p^2} \overset{①}{=} \prod_{s=m+1}^{i}\frac{d_s{b_s}^2(k_s-s_s+1)^2}{16p^2}.$$

where ① holds since we have $\left\|\boldsymbol{\phi}_{(i,i)}\right\|_F^2 = \left\|\frac{\partial\boldsymbol{X}_{(i)}^k(s,t)}{\partial\boldsymbol{X}_{(i)}}\right\|_F^2 = 1$.

Therefore, we have

$$\begin{aligned}
\left\|\frac{\partial\boldsymbol{x}_{(i)}^k}{\partial\boldsymbol{w}_{(j)}}\right\|_F^2 =& \sum_{s=1}^{r_i}\sum_{t=1}^{c_i}\left\|\frac{\partial\boldsymbol{X}_{(i)}^k(s,t)}{\partial\boldsymbol{w}_{(j)}}\right\|_F^2 \leq \sum_{s=1}^{r_i}\sum_{t=1}^{c_i}\tilde{r}_{j-1}\tilde{c}_{j-1}d_{j-1}(k_j-s_j+1)^2\left\|\boldsymbol{\phi}_{(i,j)}\right\|_F^2\\
=& r_ic_i\tilde{r}_{j-1}\tilde{c}_{j-1}d_{j-1}(k_j-s_j+1)^2\left\|\boldsymbol{\phi}_{(i,j)}\right\|_F^2\\
\leq& r_ic_i\tilde{r}_{j-1}\tilde{c}_{j-1}d_{j-1}(k_j-s_j+1)^2\prod_{s=j+1}^{i}\frac{d_s{b_s}^2(k_s-s_s+1)^2}{16p^2}.
\end{aligned}$$

It further gives

$$\left\|\frac{\partial\boldsymbol{x}_{(i)}}{\partial\boldsymbol{w}_{(j)}}\right\|_F^2 = \sum_{s=1}^{d_i}\left\|\frac{\partial\boldsymbol{x}_{(i)}^s}{\partial\boldsymbol{w}_{(j)}}\right\|_F^2 \leq d_ir_ic_i\tilde{r}_{j-1}\tilde{c}_{j-1}d_{j-1}(k_j-s_j+1)^2\prod_{s=j+1}^{i}\frac{d_s{b_s}^2(k_s-s_s+1)^2}{16p^2}.$$

The proof is completed. □

**Lemma 13.** *Suppose that the activation function $\sigma_1$ is sigmoid and $\sigma_2$ is softmax, and the loss function $f(\boldsymbol{w},\boldsymbol{D})$ is squared loss. Then the gradient of $\boldsymbol{\delta}_l^s$ with respect to $\boldsymbol{w}_{(j)}$ can be bounded as follows:*

$$\left\|\frac{\partial\boldsymbol{\delta}_l^s}{\partial\boldsymbol{w}_{(j)}}\right\|_F^2 \leq \frac{\tilde{\vartheta}{b_{l+1}}^2}{16p^2}\left\|\boldsymbol{W}_{(l+1)}^s\right\|_F^2 d_lr_lc_l\tilde{r}_{j-1}\tilde{c}_{j-1}d_{j-1}(k_j-s_j+1)^2\prod_{s=j+1}^{l}\frac{d_s{b_s}^2(k_s-s_s+1)^2}{16p^2}$$

*and*

$$\left\|\frac{\partial\boldsymbol{\delta}_l}{\partial\boldsymbol{w}_{(j)}}\right\|_F^2 \leq \frac{\tilde{\vartheta}{b_{l+1}}^4}{16p^2}d_lr_lc_l\tilde{r}_{j-1}\tilde{c}_{j-1}d_{j-1}(k_j-s_j+1)^2\prod_{s=j+1}^{l}\frac{d_s{b_s}^2(k_s-s_s+1)^2}{16p^2},$$

*where $\tilde{\vartheta} = \frac{3}{64}$.*

*Proof.* Assume that $\boldsymbol{W}_{(l+1)} = [\boldsymbol{W}_{(l+1)}^1, \boldsymbol{W}_{(l+1)}^2, \cdots, \boldsymbol{W}_{(l+1)}^{d_l}]$ where $\boldsymbol{W}_{(l+1)}^i \in \mathbb{R}^{d_{l+1}\times \tilde{r}_l\tilde{c}_l}$ is a submatrix in $\boldsymbol{W}_{(l+1)}$. Then we have $\boldsymbol{v} = \sigma_2(\sum_{k=1}^{d_l} \boldsymbol{W}_{(l+1)}^k \boldsymbol{z}_{(l)}^k)$. For brevity, we further define a matrix $\boldsymbol{G}_{(k)}$ as follows:

$$\boldsymbol{G}_{(k)}=\Big[\underbrace{\sigma_1'\left(\boldsymbol{x}_{(k)}\right),\sigma_1'\left(\boldsymbol{x}_{(k)}\right),\cdots,\sigma_1'\left(\boldsymbol{x}_{(k)}\right)}_{r_k c_k \text{ columns}}\Big] \in \mathbb{R}^{r_k c_k d_k \times r_k c_k},$$

Then we have

$$\frac{\partial}{\partial \boldsymbol{z}_{(l)}}\left(\frac{\partial f(\boldsymbol{w},\boldsymbol{D})}{\partial \boldsymbol{z}_{(l)}^s}\right) = \left[(\boldsymbol{v}-\boldsymbol{y})^T \otimes (\boldsymbol{W}_{(l+1)}^s)^T\right] \mathsf{Q}\left(\boldsymbol{u}\right)\boldsymbol{W}_{(l+1)} + (\boldsymbol{W}_{(l+1)}^s)^T \mathsf{G}\left(\boldsymbol{u}\right)\mathsf{G}\left(\boldsymbol{u}\right)\boldsymbol{W}_{(l+1)},$$

where $\mathsf{Q}\left(\boldsymbol{u}\right)$ is a matrix of size $d_{l+1}^2 \times d_{l+1}$ whose $(s,(s-1)d_{l+1}+s)$ entry equal to $\sigma_1(\boldsymbol{u}_s)(1-\sigma_1(\boldsymbol{u}_s))(1-2\sigma_1(\boldsymbol{u}_s))$ and rest entries are all 0. Accordingly, we have

$$\begin{aligned}
\left\|\frac{\partial}{\partial \boldsymbol{z}_{(l)}}\left(\frac{\partial f(\boldsymbol{w},\boldsymbol{D})}{\partial \boldsymbol{z}_{(l)}^s}\right)\right\|_F^2 \leq & 2\left(\left\|\left[(\boldsymbol{v}-\boldsymbol{y})^T \otimes (\boldsymbol{W}_{(l+1)}^s)^T\right] \mathsf{Q}\left(\boldsymbol{u}\right)\boldsymbol{W}_{(l+1)}\right\|_F^2 + \left\|(\boldsymbol{W}_{(l+1)}^s)^T \mathsf{G}\left(\boldsymbol{u}\right)\mathsf{G}\left(\boldsymbol{u}\right)\boldsymbol{W}_{(l+1)}\right\|_F^2\right)\\
\leq & 2\left(\frac{2^6}{3^8}\left\|\boldsymbol{v}-\boldsymbol{y}\right\|_F^2 \left\|\boldsymbol{W}_{(l+1)}^s\right\|_F^2 \left\|\boldsymbol{W}_{(l+1)}\right\|_F^2 + \frac{1}{16^2}\left\|\boldsymbol{W}_{(l+1)}^s\right\|_F^2 \left\|\boldsymbol{W}_{(l+1)}\right\|_F^2\right)\\
\overset{①}{\leq} & 2\left(\frac{2^7}{3^8}+\frac{1}{16^2}\right) {b_{l+1}}^2 \left\|\boldsymbol{W}_{(l+1)}^s\right\|_F^2\\
\leq & \frac{3}{64} {b_{l+1}}^2 \left\|\boldsymbol{W}_{(l+1)}^s\right\|_F^2,
\end{aligned}$$

where we have $\left\|\boldsymbol{v}-\boldsymbol{y}\right\|_F^2 \leq 2$ by Lemma 8. Then by similar way, we can have

$$\begin{aligned}
\left\|\frac{\partial \boldsymbol{\delta}_l^s}{\partial \boldsymbol{w}_{(j)}}\right\|_F^2 = & \left\|\frac{\partial}{\partial \boldsymbol{z}_{(l)}}\left(\frac{\partial f(\boldsymbol{w},\boldsymbol{D})}{\partial \boldsymbol{z}_{(l)}^s}\right)\frac{\partial \boldsymbol{z}_{(l)}}{\partial \boldsymbol{y}_l}\frac{\partial \boldsymbol{y}_l}{\partial \boldsymbol{x}_{(l)}}\frac{\partial \boldsymbol{x}_{(l)}}{\partial \boldsymbol{w}_{(j)}}\right\|_F^2 \leq \frac{3}{64*16p^2}{b_{l+1}}^2\left\|\boldsymbol{W}_{(l+1)}^s\right\|_F^2\left\|\frac{\partial \boldsymbol{x}_{(l)}}{\partial \boldsymbol{w}_{(j)}}\right\|_F^2\\
\leq & \frac{3}{64*16p^2}{b_{l+1}}^2\left\|\boldsymbol{W}_{(l+1)}^s\right\|_F^2 d_l r_l c_l \tilde{r}_{j-1}\tilde{c}_{j-1}d_{j-1}(k_j-s_j+1)^2\prod_{s=j+1}^{l}\frac{d_s{b_s}^2(k_s-s_s+1)^2}{16p^2}.
\end{aligned}$$

Therefore, we can further obtain:

$$\left\|\frac{\partial \boldsymbol{\delta}_l}{\partial \boldsymbol{w}_{(j)}}\right\|_F^2 = \sum_{s=1}^{d_{l+1}}\left\|\frac{\partial \boldsymbol{\delta}_l^s}{\partial \boldsymbol{w}_{(j)}}\right\|_F^2 \leq \frac{3}{64*16p^2}{b_{l+1}}^4 d_l r_l c_l \tilde{r}_{j-1}\tilde{c}_{j-1}d_{j-1}(k_j-s_j+1)^2\prod_{s=j+1}^{l}\frac{d_s{b_s}^2(k_s-s_s+1)^2}{16p^2}.$$

The proof is completed. □

**Lemma 14.** *Suppose that the activation function $\sigma_1$ is sigmoid and $\sigma_2$ is softmax, and the loss function $f(\boldsymbol{w},\boldsymbol{D})$ is squared loss. Then the Hessian of $f(\boldsymbol{w},\boldsymbol{x})$ with respect to $\boldsymbol{w}$ can be bounded as follows:*

$$\left\|\nabla_{\boldsymbol{w}}^2 f(\boldsymbol{w},\boldsymbol{D})\right\|_F^2 \leq \mathcal{O}\left(\gamma^2\right),$$

*where $\gamma = \left(\frac{\vartheta {b_{l+1}}^2 d_0^2}{{b_1}^4 d_1^2} l^2 r_0^2 c_0^2 \left[\prod_{s=1}^{l}\frac{d_s{b_s}^2(k_s-s_s+1)^2}{8\sqrt{2}p^2}\right]^2\right)^{1/2}$. With the same condition, we can bound the operation norm of $\nabla_{\boldsymbol{w}}^3 f(\boldsymbol{w},\boldsymbol{D})$. That is, there exists a universal constant $\nu$ such that $\left\|\nabla_{\boldsymbol{w}}^3 f(\boldsymbol{w},\boldsymbol{D})\right\|_{op} \leq \nu$.*

*Proof.* From Lemma 9, we can further compute the Hessian matrix $\nabla_{\boldsymbol{w}}^2 f(\boldsymbol{w},\boldsymbol{D})$. Recall that $\boldsymbol{w}_{(i)}^k \in \mathbb{R}^{{k_i}^2 d_{i-1}}$ $(k = 1,\cdots,d_l)$ is the vectorization of $\boldsymbol{W}_{(i)}^k \in \mathbb{R}^{k_i\times k_i\times d_{i-1}}$, *i.e.* $\boldsymbol{w}_{(i)}^k = \left[\mathsf{vec}\left(\boldsymbol{W}_{(i)}^1(:,:1)\right);\cdots;\ \mathsf{vec}\left(\boldsymbol{W}_{(i)}^{d_i}(:,:,d_{i-1})\right)\right]$.

Let $\boldsymbol{w}_{(i)}=\left[\boldsymbol{w}_{(i)}^{1};\cdots;\boldsymbol{w}_{(i)}^{d_i}\right]\in\mathbb{R}^{{k_i}^2 d_{i-1}\times d_i}$ $(i=1,\cdots,l)$. Also, $\boldsymbol{w}_{(l+1)}\in\mathbb{R}^{\tilde{r}_l\tilde{c}_l d_l d_{l+1}}$ is the vectorization of the weight matrix $\boldsymbol{W}_{(l+1)}$. Then if $1\leq i,j\leq l$, we can have

$$\begin{aligned}\frac{\partial^2 f(\boldsymbol{w},\boldsymbol{D})}{\partial\boldsymbol{w}_{(j)}\partial\boldsymbol{w}_{(i)}^{k}}=&\begin{bmatrix}\frac{\partial^2 f(\boldsymbol{w},\boldsymbol{D})}{\partial\boldsymbol{w}_{(j)}\partial\boldsymbol{w}_{(i)}^{k,1}}\\ \frac{\partial^2 f(\boldsymbol{w},\boldsymbol{D})}{\partial\boldsymbol{w}_{(j)}\partial\boldsymbol{w}_{(i)}^{k,2}}\\ \vdots\\ \frac{\partial^2 f(\boldsymbol{w},\boldsymbol{D})}{\partial\boldsymbol{w}_{(j)}\partial\boldsymbol{w}_{(i)}^{k,d_{i-1}}}\end{bmatrix}=\begin{bmatrix}\frac{\partial\left(\mathsf{vec}\left(\boldsymbol{Z}_{(i-1)}^{1}\widetilde{\circledast}\widetilde{\boldsymbol{\delta}}_i^k\right)\right)}{\partial\boldsymbol{w}_{(j)}}\\ \frac{\partial\left(\mathsf{vec}\left(\boldsymbol{Z}_{(i-1)}^{2}\widetilde{\circledast}\widetilde{\boldsymbol{\delta}}_i^k\right)\right)}{\partial\boldsymbol{w}_{(j)}}\\ \vdots\\ \frac{\partial\left(\mathsf{vec}\left(\boldsymbol{Z}_{(i-1)}^{d_{i-1}}\widetilde{\circledast}\widetilde{\boldsymbol{\delta}}_i^k\right)\right)}{\partial\boldsymbol{w}_{(j)}}\end{bmatrix}\\ =&\begin{bmatrix}\mathsf{P}_1\left(\widetilde{\boldsymbol{\delta}}_i^k\right)\frac{\partial\left(\mathsf{vec}\left(\boldsymbol{Z}_{(i-1)}^{1}\right)\right)}{\partial\boldsymbol{w}_{(j)}}+\mathsf{P}_2\left(\boldsymbol{Z}_{(i-1)}^{1}\right)\frac{\partial\left(\mathsf{vec}\left(\widetilde{\boldsymbol{\delta}}_i^k\right)\right)}{\partial\boldsymbol{w}_{(j)}}\\ \mathsf{P}_1\left(\widetilde{\boldsymbol{\delta}}_i^k\right)\frac{\partial\left(\mathsf{vec}\left(\boldsymbol{Z}_{(i-1)}^{2}\right)\right)}{\partial\boldsymbol{w}_{(j)}}+\mathsf{P}_2\left(\boldsymbol{Z}_{(i-1)}^{2}\right)\frac{\partial\left(\mathsf{vec}\left(\widetilde{\boldsymbol{\delta}}_i^k\right)\right)}{\partial\boldsymbol{w}_{(j)}}\\ \vdots\\ \mathsf{P}_1\left(\widetilde{\boldsymbol{\delta}}_i^k\right)\frac{\partial\left(\mathsf{vec}\left(\boldsymbol{Z}_{(i-1)}^{d_{i-1}}\right)\right)}{\partial\boldsymbol{w}_{(j)}}+\mathsf{P}_2\left(\boldsymbol{Z}_{(i-1)}^{d_{i-1}}\right)\frac{\partial\left(\mathsf{vec}\left(\widetilde{\boldsymbol{\delta}}_i^k\right)\right)}{\partial\boldsymbol{w}_{(j)}}\end{bmatrix}\in\mathbb{R}^{{k_i}^2 d_{i-1}\times {k_j}^2 d_j d_{j-1}},\end{aligned}\tag{4}$$

where $\mathsf{P}_1\left(\widetilde{\boldsymbol{\delta}}_i^k\right)\in\mathbb{R}^{{k_i}^2\times\tilde{r}_{i-1}\tilde{c}_{i-1}d_{i-1}}$ and $\mathsf{P}_2\left(\boldsymbol{Z}_{(i-1)}^{d_{i-1}}\right)\in\mathbb{R}^{{k_i}^2\times(\tilde{r}_i-k_i+1)(\tilde{c}_i-k_i+1)}$ satisfy: each row in $\mathsf{P}_1\left(\widetilde{\boldsymbol{\delta}}_i^k\right)$ contains the vectorization of $(\widetilde{\boldsymbol{\delta}}_i^k)^T$ at the right position and the remaining entries are 0s, and each row in $\mathsf{P}_2\left(\boldsymbol{Z}_{(i-1)}^{d_{i-1}}\right)$ is the submatrix in $\boldsymbol{Z}_{(i-1)}^{d_{i-1}}$ that need to conduct inner product with $\widetilde{\boldsymbol{\delta}}_i^k$ in turn. Note that there are $s_i-1$ rows and columns between each neighboring nonzero entries in $\boldsymbol{N}$ which is decided by the definition of $\widetilde{\boldsymbol{\delta}}_{i+1}$ in Sec. B.1. Accordingly, we have

$$\left\|\mathsf{P}_1\left(\widetilde{\boldsymbol{\delta}}_i^k\right)\frac{\partial\left(\mathsf{vec}\left(\boldsymbol{Z}_{(i-1)}^{d_{i-1}}\right)\right)}{\partial\boldsymbol{w}_{(j)}}\right\|_F^2\leq(k_i-s_i+1)^2\|\widetilde{\boldsymbol{\delta}}_i^k\|_F^2\left\|\frac{\partial\left(\mathsf{vec}\left(\boldsymbol{Z}_{(i-1)}^{d_{i-1}}\right)\right)}{\partial\boldsymbol{w}_{(j)}}\right\|_F^2=(k_i-s_i+1)^2\|\boldsymbol{\delta}_i^k\|_F^2\left\|\frac{\partial\left(\mathsf{vec}\left(\boldsymbol{Z}_{(i-1)}^{d_{i-1}}\right)\right)}{\partial\boldsymbol{w}_{(j)}}\right\|_F^2$$

and

$$\left\|\mathsf{P}_2\left(\boldsymbol{Z}_{(i-1)}^{d_{i-1}}\right)\frac{\partial\left(\mathsf{vec}\left(\widetilde{\boldsymbol{\delta}}_i^k\right)\right)}{\partial\boldsymbol{w}_{(j)}}\right\|_F^2\leq(k_i-s_i+1)^2\|\boldsymbol{Z}_{(i-1)}^{d_{i-1}}\|_F^2\left\|\frac{\partial\left(\mathsf{vec}\left(\widetilde{\boldsymbol{\delta}}_i^k\right)\right)}{\partial\boldsymbol{w}_{(j)}}\right\|_F^2=(k_i-s_i+1)^2\|\boldsymbol{Z}_{(i-1)}^{d_{i-1}}\|_F^2\left\|\frac{\partial\left(\mathsf{vec}\left(\boldsymbol{\delta}_i^k\right)\right)}{\partial\boldsymbol{w}_{(j)}}\right\|_F^2.$$

Then in order to bound

$$\|\nabla_{\boldsymbol{w}}^2 f(\boldsymbol{w},\boldsymbol{D})\|_F^2=\sum_{i=1}^{l+1}\sum_{j=1}^{l+1}\left\|\frac{\partial^2 f(\boldsymbol{w},\boldsymbol{D})}{\partial\boldsymbol{w}_{(j)}\partial\boldsymbol{w}_{(i)}}\right\|_F^2,$$

we try to bound each term separately. So we consider the following five cases: $l\geq i\geq j$, $i\leq j\leq l$, $l+1=i>j$, $l+1=j>i$ and $l+1=i=j$.

**Case 1:** $l\geq i\geq j$

In the following, we first consider the first case, *i.e.* $i \geq j$, and bound

$$
\begin{aligned}
&\left\|\frac{\partial\left(\mathsf{vec}\left(\widetilde{\boldsymbol{\delta}}_i^k\right)\right)}{\partial \boldsymbol{w}_{(j)}}\right\|_F^2 = \left\|\frac{\partial\left(\mathsf{vec}\left(\boldsymbol{\delta}_i^k\right)\right)}{\partial \boldsymbol{w}_{(j)}}\right\|_F^2 = \left\|\frac{\partial}{\partial \boldsymbol{w}_{(j)}}\mathsf{vec}\left[\mathsf{up}\left(\sum_{s=1}^{d_{i+1}}\widetilde{\boldsymbol{\delta}}_{i+1}^s \widetilde{\circledast} \widehat{\boldsymbol{W}}_{(i+1)}^{s,k}\right)\odot \sigma_1'(\boldsymbol{X}_{(i)}^k)\right]\right\|_F^2\\
\overset{①}{\leq}&\frac{2}{16}\left\|\frac{\partial}{\partial \boldsymbol{w}_{(j)}}\mathsf{vec}\left[\mathsf{up}\left(\sum_{s=1}^{d_{i+1}}\widetilde{\boldsymbol{\delta}}_{i+1}^s \widetilde{\circledast} \widehat{\boldsymbol{W}}_{(i+1)}^{s,k}\right)\right]\right\|_F^2 + 2\left\|\mathsf{up}\left(\sum_{s=1}^{d_{i+1}}\widetilde{\boldsymbol{\delta}}_{i+1}^s \widetilde{\circledast} \widehat{\boldsymbol{W}}_{(i+1)}^{s,k}\right)\right\|_F^2\left\|\frac{\partial \sigma_1'(\boldsymbol{X}_{(i)}^k)}{\partial \boldsymbol{w}_{(j)}}\right\|_F^2\\
=&\frac{2}{16p^2}\left\|\frac{\partial}{\partial \boldsymbol{w}_{(j)}}\mathsf{vec}\left[\sum_{s=1}^{d_{i+1}}\widetilde{\boldsymbol{\delta}}_{i+1}^s \widetilde{\circledast} \widehat{\boldsymbol{W}}_{(i+1)}^{s,k}\right]\right\|_F^2 + \frac{2}{p^2}\left\|\sum_{s=1}^{d_{i+1}}\widetilde{\boldsymbol{\delta}}_{i+1}^s \widetilde{\circledast} \widehat{\boldsymbol{W}}_{(i+1)}^{s,k}\right\|_F^2\left\|\frac{\partial \mathsf{vec}\left(\sigma'(\boldsymbol{x}_{(i)}^k)\right)}{\partial \boldsymbol{x}_{(i)}^k}\frac{\partial \boldsymbol{x}_{(i)}^k}{\partial \boldsymbol{w}_{(j)}}\right\|_F^2\\
\overset{②}{\leq}&\frac{2}{16p^2}\left\|\frac{\partial}{\partial \boldsymbol{w}_{(j)}}\mathsf{vec}\left[\sum_{s=1}^{d_{i+1}}\widetilde{\boldsymbol{\delta}}_{i+1}^s \widetilde{\circledast} \widehat{\boldsymbol{W}}_{(i+1)}^{s,k}\right]\right\|_F^2 + \frac{2\cdot 2^6}{3^8p^2}\left\|\sum_{s=1}^{d_{i+1}}\widetilde{\boldsymbol{\delta}}_{i+1}^s \widetilde{\circledast} \widehat{\boldsymbol{W}}_{(i+1)}^{s,k}\right\|_F^2\left\|\frac{\partial \boldsymbol{X}_{(i)}^k}{\partial \boldsymbol{w}_{(j)}}\right\|_F^2\\
\overset{③}{\leq}&\frac{2d_{i+1}}{16p^2}(k_{i+1}-s_{i+1}+1)^2\sum_{s=1}^{d_{i+1}}\|\widehat{\boldsymbol{W}}_{(i+1)}^{s,k}\|_F^2\left\|\frac{\partial \widetilde{\boldsymbol{\delta}}_{i+1}^s}{\partial \boldsymbol{w}_{(j)}}\right\|_F^2 + \frac{2\cdot 2^6}{3^8p^2}d_{i+1}(k_{i+1}-s_{i+1}+1)^2\sum_{s=1}^{d_{i+1}}\left\|\widehat{\boldsymbol{W}}_{(i+1)}^{s,k}\right\|_F^2\left\|\widetilde{\boldsymbol{\delta}}_{i+1}^s\right\|_F^2\left\|\frac{\partial \boldsymbol{X}_{(i)}^k}{\partial \boldsymbol{w}_{(j)}}\right\|_F^2\\
\overset{④}{=}&\frac{2d_{i+1}}{16p^2}(k_{i+1}-s_{i+1}+1)^2\sum_{s=1}^{d_{i+1}}\|\boldsymbol{W}_{(i+1)}^{s,k}\|_F^2\left\|\frac{\partial \boldsymbol{\delta}_{i+1}^s}{\partial \boldsymbol{w}_{(j)}}\right\|_F^2 + \frac{2\cdot 2^6}{3^8p^2}d_{i+1}(k_{i+1}-s_{i+1}+1)^2\sum_{s=1}^{d_{i+1}}\left\|\boldsymbol{W}_{(i+1)}^{s,k}\right\|_F^2\left\|\boldsymbol{\delta}_{i+1}^s\right\|_F^2\left\|\frac{\partial \boldsymbol{X}_{(i)}^k}{\partial \boldsymbol{w}_{(j)}}\right\|_F^2.
\end{aligned}
\tag{5}
$$

① holds since $\boldsymbol{X}_{(i)}^k$ is independent on $\boldsymbol{w}_{(j)}$ and the values of entries in $\sigma_1'(\boldsymbol{X}_{(i)}^k)$ is not larger than $1/4$ since for any constant $a$, $\sigma'(a)=\sigma(a)(1-\sigma(a))\leq 1/4$. ② holds since for arbitrary tensor $\boldsymbol{M}$, we have $\|\mathsf{up}(\boldsymbol{M})\|_F^2 \leq \|\boldsymbol{M}\|_F^2/p^2$ in Lemma 8, and we also have

$$
\left\|\frac{\partial \mathsf{vec}\left(\sigma'(\boldsymbol{x}_{(i)}^k)\right)}{\partial \boldsymbol{x}_{(i)}^k}\frac{\partial \boldsymbol{x}_{(i)}^k}{\partial \boldsymbol{w}_{(j)}}\right\|_F^2 = \left\|\mathsf{Q}\left(\boldsymbol{x}_{(i)}^k\right)\frac{\partial \boldsymbol{x}_{(i)}^k}{\partial \boldsymbol{w}_{(j)}}\right\|_F^2 \leq \frac{2^6}{3^8}\left\|\frac{\partial \boldsymbol{x}_{(i)}^k}{\partial \boldsymbol{w}_{(j)}}\right\|_F^2 = \frac{2^6}{3^8}\left\|\frac{\partial \boldsymbol{X}_{(i)}^k}{\partial \boldsymbol{w}_{(j)}}\right\|_F^2.
$$

③ holds since we can just adopt similar strategy in Eqn. (4) to separate $\widehat{\boldsymbol{W}}_{(i+1)}^{s,k}$ and the conclusion in Lemma 8; ④ holds since the difference between $\widetilde{\boldsymbol{\delta}}_{i+1}^s$ and $\boldsymbol{\delta}_{i+1}^s$ is that we pad 0 around $\boldsymbol{\delta}_{i+1}^s$ to obtain $\widetilde{\boldsymbol{\delta}}_{i+1}^s$, indicating $\|\boldsymbol{\delta}_{i+1}^s\|_F^2 = \|\widetilde{\boldsymbol{\delta}}_{i+1}^s\|_F^2$. Accordingly, we can further bound

$$
\begin{aligned}
&\left\|\frac{\partial \widetilde{\boldsymbol{\delta}}_i}{\partial \boldsymbol{w}_{(j)}}\right\|_F^2 = \sum_{k=1}^{d_{i-1}}\left\|\frac{\partial\left(\mathsf{vec}\left(\boldsymbol{\delta}_i^k\right)\right)}{\partial \boldsymbol{w}_{(j)}}\right\|_F^2\\
\leq&\frac{2d_{i+1}}{16p^2}(k_{i+1}-s_{i+1}+1)^2\sum_{k=1}^{d_{i-1}}\sum_{s=1}^{d_{i+1}}\|\boldsymbol{W}_{(i+1)}^{s,k}\|_F^2\left\|\frac{\partial \boldsymbol{\delta}_{i+1}^s}{\partial \boldsymbol{w}_{(j)}}\right\|_F^2 + \frac{2\cdot 2^6}{3^8p^2}d_{i+1}(k_{i+1}-s_{i+1}+1)^2\sum_{k=1}^{d_{i-1}}\sum_{s=1}^{d_{i+1}}\left\|\boldsymbol{W}_{(i+1)}^{s,k}\right\|_F^2\left\|\boldsymbol{\delta}_{i+1}^s\right\|_F^2\left\|\frac{\partial \boldsymbol{X}_{(i)}^k}{\partial \boldsymbol{w}_{(j)}}\right\|_F^2\\
\leq&\frac{2d_{i+1}}{16p^2}(k_{i+1}-s_{i+1}+1)^2\sum_{s=1}^{d_{i+1}}\|\boldsymbol{W}_{(i+1)}^{s}\|_F^2\left\|\frac{\partial \boldsymbol{\delta}_{i+1}^s}{\partial \boldsymbol{w}_{(j)}}\right\|_F^2 + \frac{2\cdot 2^6}{3^8p^2}d_{i+1}(k_{i+1}-s_{i+1}+1)^2\left\|\boldsymbol{\delta}_{i+1}\right\|_F^2\max_s\left\|\boldsymbol{W}_{(i+1)}^{s}\right\|_F^2\max_k\left\|\frac{\partial \boldsymbol{X}_{(i)}^k}{\partial \boldsymbol{w}_{(j)}}\right\|_F^2\\
\overset{①}{\leq}&\frac{2d_{i+1}}{16p^2}(k_{i+1}-s_{i+1}+1)^2{b_{i+1}}^2\left\|\frac{\partial \boldsymbol{\delta}_{i+1}}{\partial \boldsymbol{w}_{(j)}}\right\|_F^2 + \frac{2\cdot 2^6}{3^8p^2}d_{i+1}(k_{i+1}-s_{i+1}+1)^2{b_{i+1}}^2\left\|\boldsymbol{\delta}_{i+1}\right\|_F^2\max_k\left\|\frac{\partial \boldsymbol{X}_{(i)}^k}{\partial \boldsymbol{w}_{(j)}}\right\|_F^2\\
\overset{②}{\leq}&\frac{2d_{i+1}}{16p^2}(k_{i+1}-s_{i+1}+1)^2{b_{i+1}}^2\left\|\frac{\partial \boldsymbol{\delta}_{i+1}}{\partial \boldsymbol{w}_{(j)}}\right\|_F^2 + \frac{\vartheta {b_{l+1}}^2 d_{j-1}}{3p^2{b_j}^2 d_j}r_ic_ir_{j-1}c_{j-1}\prod_{s=j}^{l}\frac{d_s{b_s}^2(k_s-s_s+1)^2}{16p^2},
\end{aligned}
$$

where ① holds since we have $\|\boldsymbol{W}_{(i+1)}^s\|_F \leq r_w$; ② holds due to the bounds of $\left\|\boldsymbol{\delta}_{i+1}\right\|_F^2$ and $\left\|\frac{\partial \boldsymbol{X}_{(i)}^k}{\partial \boldsymbol{w}_{(j)}}\right\|_F^2$ in Lemma 10 and 12.

Then, we can use the above recursion inequality to further obtain

$$\begin{aligned}
&\left\|\frac{\partial \boldsymbol{\delta}_i}{\partial \boldsymbol{w}_{(j)}}\right\|_F^2\\
\leq&\left[\prod_{s=i+1}^{i+1}\frac{2d_s}{16p^2}(k_s-s_s+1)^2{b_s}^2\right]\left(\frac{2d_{i+2}}{16p^2}(k_{i+2}-s_{i+2}+1)^2{b_{i+2}}^2\left\|\frac{\partial \boldsymbol{\delta}_{i+2}}{\partial \boldsymbol{w}_{(j)}}\right\|_F^2+\frac{\vartheta {b_{l+1}}^2 d_{j-1}}{3p^2{b_j}^2 d_j}r_{j-1}c_{j-1}r_{i+1}c_{i+1}\prod_{s=j}^{l}\frac{d_s{b_s}^2(k_s-s_s+1)^2}{16p^2}\right)\\
&+\frac{\vartheta {b_{l+1}}^2 d_{j-1}}{3p^2{b_j}^2 d_j}r_ic_ir_{j-1}c_{j-1}\prod_{s=j}^{l}\frac{d_s{b_s}^2(k_s-s_s+1)^2}{16p^2}\\
\leq&\left[\prod_{s=i+1}^{l}\frac{2d_s}{16p^2}(k_s-s_s+1)^2{b_s}^2\right]\left\|\frac{\partial \boldsymbol{\delta}_l}{\partial \boldsymbol{w}_{(j)}}\right\|_F^2+\frac{\vartheta {b_{l+1}}^2 d_{j-1}}{3p^2{b_j}^2 d_j}r_{j-1}c_{j-1}\prod_{s=j}^{l}\frac{d_s{b_s}^2(k_s-s_s+1)^2}{16p^2}\left[r_ic_i+r_{i+1}c_{i+1}\left[\prod_{s=i+1}^{i+1}\frac{2d_s}{16p^2}(k_s-s_s+1)^2{b_s}^2\right]\right.\\
&\left.+r_{i+2}c_{i+2}\left[\prod_{s=i+1}^{i+2}\frac{2d_s}{16p^2}(k_s-s_s+1)^2{b_s}^2\right]+\cdots+r_lc_l\left[\prod_{s=i+1}^{l}\frac{2d_s}{16p^2}(k_s-s_s+1)^2{b_s}^2\right]\right].
\end{aligned}$$

By Lemma 13, we have

$$\left\|\frac{\partial \boldsymbol{\delta}_l}{\partial \boldsymbol{w}_{(j)}}\right\|_F^2\leq\frac{\tilde{\vartheta}{b_{l+1}}^4 d_{j-1}}{p^2{b_j}^2 d_j}d_lr_lc_lr_{j-1}c_{j-1}\prod_{s=j}^{l}\frac{d_s{b_s}^2(k_s-s_s+1)^2}{16p^2},$$

where $\tilde{\vartheta}=\frac{3}{64}$. Thus, we can establish

$$\left\|\frac{\partial \boldsymbol{\delta}_i}{\partial \boldsymbol{w}_{(j)}}\right\|_F^2\leq\frac{\tilde{\vartheta}{b_{l+1}}^2 d_{j-1}}{p^2{b_j}^2 d_j}r_{j-1}c_{j-1}\left[\frac{\tau}{3}+{b_{l+1}}^2d_lr_lc_l\prod_{s=i+1}^{l}\frac{d_s{b_s}^2(k_s-s_s+1)^2}{16p^2}\right]\prod_{s=j}^{l}\frac{d_s{b_s}^2(k_s-s_s+1)^2}{16p^2}.$$

where $\tau=r_ic_i+r_{i+1}c_{i+1}\left[\prod_{s=i+1}^{i+1}\frac{2d_s}{16p^2}(k_s-s_s+1)^2{b_s}^2\right]+\cdots+r_lc_l\left[\prod_{s=i+1}^{l}\frac{2d_s}{16p^2}(k_s-s_s+1)^2{b_s}^2\right]$. It further gives the bound of $\left\|\frac{\partial^2 f(\boldsymbol{w},\boldsymbol{x})}{\partial \boldsymbol{w}_{(j)}\partial \boldsymbol{w}_{(i)}}\right\|_F^2$ as follows:

$$\begin{aligned}
\left\|\frac{\partial^2 f(\boldsymbol{w},\boldsymbol{D})}{\partial \boldsymbol{w}_{(j)}\partial \boldsymbol{w}_{(i)}}\right\|_F^2=&\sum_{k=1}^{d_{i-1}}\left\|\frac{\partial^2 f(\boldsymbol{w},\boldsymbol{D})}{\partial \boldsymbol{w}_{(j)}\partial \boldsymbol{w}_{(i)}^k}\right\|_F^2=\sum_{k=1}^{d_{i-1}}\sum_{s=1}^{d_i}\left\|\mathsf{P}_1\left(\boldsymbol{\delta}_i^k\right)\frac{\partial\left(\mathsf{vec}\left(\boldsymbol{X}_{(i-1)}^s\right)\right)}{\partial \boldsymbol{w}_{(j)}}+\mathsf{P}_2\left(\boldsymbol{X}_{(i-1)}^s\right)\frac{\partial\left(\mathsf{vec}\left(\boldsymbol{\delta}_i^k\right)\right)}{\partial \boldsymbol{w}_{(j)}}\right\|_F^2\\
\leq&2\sum_{k=1}^{d_{i-1}}\sum_{s=1}^{d_i}\left(\left\|\mathsf{P}_1\left(\boldsymbol{\delta}_i^k\right)\frac{\partial\left(\mathsf{vec}\left(\boldsymbol{X}_{(i-1)}^s\right)\right)}{\partial \boldsymbol{w}_{(j)}}\right\|_F^2+\left\|\mathsf{P}_2\left(\boldsymbol{X}_{(i-1)}^s\right)\frac{\partial\left(\mathsf{vec}\left(\boldsymbol{\delta}_i^k\right)\right)}{\partial \boldsymbol{w}_{(j)}}\right\|_F^2\right)\\
\leq&2(k_i-s_i+1)^2\sum_{k=1}^{d_{i-1}}\sum_{s=1}^{d_i}\left(\left\|\boldsymbol{\delta}_i^k\right\|_F^2\left\|\frac{\partial\left(\mathsf{vec}\left(\boldsymbol{X}_{(i-1)}^s\right)\right)}{\partial \boldsymbol{w}_{(j)}}\right\|_F^2+\left\|\boldsymbol{X}_{(i-1)}^s\right\|_F^2\left\|\frac{\partial\left(\mathsf{vec}\left(\boldsymbol{\delta}_i^k\right)\right)}{\partial \boldsymbol{w}_{(j)}}\right\|_F^2\right)\\
\leq&2(k_i-s_i+1)^2\left(\left\|\boldsymbol{\delta}_i\right\|_F^2\left\|\frac{\partial\left(\mathsf{vec}\left(\boldsymbol{X}_{(i-1)}\right)\right)}{\partial \boldsymbol{w}_{(j)}}\right\|_F^2+\left\|\boldsymbol{X}_{(i-1)}\right\|_F^2\left\|\frac{\partial\left(\mathsf{vec}\left(\boldsymbol{\delta}_i\right)\right)}{\partial \boldsymbol{w}_{(j)}}\right\|_F^2\right)\\
\overset{①}{\leq}&\frac{32\vartheta {b_{l+1}}^2 d_{i-1}}{{b_i}^2{b_j}^2 d_id_j}r_{i-1}c_{i-1}r_{j-1}c_{j-1}\prod_{s=i+1}^{l}\frac{d_s{b_s}^2(k_s-s_s+1)^2}{16p^2}+r_{i-1}c_{i-1}d_{i-1}\left\|\frac{\partial\left(\mathsf{vec}\left(\boldsymbol{\delta}_i\right)\right)}{\partial \boldsymbol{w}_{(j)}}\right\|_F^2\\
\overset{②}{\leq}&\mathcal{O}\left(\frac{\vartheta {b_{l+1}}^2 d_{i-1}d_{j-1}}{{b_i}^2{b_j}^2 d_id_j}r_{i-1}c_{i-1}r_{j-1}c_{j-1}\left[\prod_{s=j}^{l}\frac{d_s{b_s}^2(k_s-s_s+1)^2}{16p^2}\right]\left[\prod_{s=i}^{l}\frac{2d_s{b_s}^2(k_s-s_s+1)^2}{16p^2}\right]\right).
\end{aligned}$$

where ① holds because of Lemma 12, while ② holds due to 13.

**Case 2:** $i \le j \le l$

Since $\frac{\partial^2 f(\boldsymbol{w},\boldsymbol{D})}{\partial \boldsymbol{w}\partial \boldsymbol{w}^T}$ is symmetrical, we have $\frac{\partial^2 f(\boldsymbol{w},\boldsymbol{D})}{\partial \boldsymbol{w}_{(i)}^T\partial \boldsymbol{w}_{(j)}} = \left(\frac{\partial^2 f(\boldsymbol{w},\boldsymbol{D})}{\partial \boldsymbol{w}_{(j)}^T\partial \boldsymbol{w}_{(i)}}\right)^T$ $(1 \le i, j \le l)$. Thus, it yields

$$\left\|\frac{\partial^2 f(\boldsymbol{w},\boldsymbol{D})}{\partial \boldsymbol{w}_{(i)}^T\partial \boldsymbol{w}_{(j)}}\right\|_F^2 = \left\|\frac{\partial^2 f(\boldsymbol{w},\boldsymbol{D})}{\partial \boldsymbol{w}_{(j)}^T\partial \boldsymbol{w}_{(i)}}\right\|_F^2.$$

**Case 3:** $l+1 = i > j$

In the following, we first consider the first case, *i.e.* cross entropy and softmax activation, and bound

$$\begin{aligned}\left\|\frac{\partial^2 f(\boldsymbol{w},\boldsymbol{D})}{\partial \boldsymbol{w}_{(j)}\partial \boldsymbol{w}_{(l+1)}}\right\|_F^2 =& \left\|\frac{\partial (\boldsymbol{v}-\boldsymbol{y})\boldsymbol{z}_{(l)}^T}{\partial \boldsymbol{w}_{(j)}}\right\|_F^2 = \left\|\left[\boldsymbol{I}_{\tilde{r}_l\tilde{c}_l d_l}\otimes(\boldsymbol{v}-\boldsymbol{y})\right]\frac{\partial \boldsymbol{z}_{(l)}^T}{\partial \boldsymbol{x}_{(l)}}\frac{\partial \boldsymbol{x}_{(l)}}{\partial \boldsymbol{w}_{(j)}} + \left[\boldsymbol{z}_{(l)}\otimes \boldsymbol{I}_{d_{l+1}}\right]\frac{\partial \boldsymbol{v}}{\partial \boldsymbol{z}_{(l)}}\frac{\partial \boldsymbol{z}_{(l)}}{\partial \boldsymbol{x}_{(l)}}\frac{\partial \boldsymbol{x}_{(l)}}{\partial \boldsymbol{w}_{(j)}}\right\|_F^2\\ =&\left\|\widetilde{\mathsf{up}}\left(\left[\boldsymbol{I}_{\tilde{r}_l\tilde{c}_l d_l}\otimes(\boldsymbol{v}-\boldsymbol{y})\right] + \left[\boldsymbol{z}_{(l)}\otimes \boldsymbol{I}_{d_{l+1}}\right]\mathsf{diag}\left(\sigma_2'(\boldsymbol{u})\right)\boldsymbol{W}_{(l+1)}\right)\odot \boldsymbol{G}_{(l)}\frac{\partial \boldsymbol{x}_{(l)}}{\partial \boldsymbol{w}_{(j)}}\right\|_F^2,\end{aligned}$$

where $\boldsymbol{G}_{(l)}$ is defined as

$$\boldsymbol{G}_{(l)}=\Big[\underbrace{\sigma_1'\left(\boldsymbol{x}_{(l)}\right),\sigma_1'\left(\boldsymbol{x}_{(l)}\right),\cdots,\sigma_1'\left(\boldsymbol{x}_{(l)}\right)}_{r_lc_l\ \text{columns}}\Big]\in\mathbb{R}^{r_lc_ld_l\times r_lc_l}.$$

Thus, we can further obtain

$$\begin{aligned}\left\|\frac{\partial^2 f(\boldsymbol{w},\boldsymbol{D})}{\partial \boldsymbol{w}_{(j)}\partial \boldsymbol{w}_{(l+1)}}\right\|_F^2 \le& \frac{1}{16p^2}\left\|\left[\boldsymbol{I}_{\tilde{r}_l\tilde{c}_l d_l}\otimes(\boldsymbol{v}-\boldsymbol{y})\right] + \left[\boldsymbol{z}_{(l)}\otimes \boldsymbol{I}_{d_{l+1}}\right]\mathsf{diag}\left(\sigma_2'(\boldsymbol{u})\right)\boldsymbol{W}_{(l+1)}\right\|_F^2\left\|\frac{\partial \boldsymbol{x}_{(l)}}{\partial \boldsymbol{w}_{(j)}}\right\|_F^2\\ \le& \frac{2}{16p^2}\left(\left\|\boldsymbol{I}_{\tilde{r}_l\tilde{c}_l d_l}\otimes(\boldsymbol{v}-\boldsymbol{y})\right\|_F^2 + \left\|\left[\boldsymbol{z}_{(l)}\otimes \boldsymbol{I}_{d_{l+1}}\right]\mathsf{diag}\left(\sigma_2'(\boldsymbol{u})\right)\boldsymbol{W}_{(l+1)}\right\|_F^2\right)\left\|\frac{\partial \boldsymbol{x}_{(l)}}{\partial \boldsymbol{w}_{(j)}}\right\|_F^2\\ \overset{①}{\le}& \frac{2}{16p^2}\left(\left\|\boldsymbol{I}_{\tilde{r}_l\tilde{c}_l d_l}\otimes(\boldsymbol{v}-\boldsymbol{y})\right\|_F^2 + \left\|\boldsymbol{z}_{(l)}\otimes \left[\mathsf{diag}\left(\sigma_2'(\boldsymbol{u})\right)\boldsymbol{W}_{(l+1)}\right]\right\|_F^2\right)\left\|\frac{\partial \boldsymbol{x}_{(l)}}{\partial \boldsymbol{w}_{(j)}}\right\|_F^2\\ \overset{②}{\le}& \frac{1}{8p^2}\tilde{r}_l\tilde{c}_ld_l\left(2+\frac{1}{16}{b_{l+1}}^2\right)d_lr_lc_l\tilde{r}_{j-1}\tilde{c}_{j-1}d_{j-1}(k_j-s_j+1)^2\prod_{s=j+1}^{l}\frac{d_s{b_s}^2(k_s-s_s+1)^2}{16p^2}\\ =&\frac{2d_{j-1}}{p^4{b_j}^2d_j}r_l^2c_l^2d_l^2r_{j-1}c_{j-1}\left(2+\frac{1}{16}{b_{l+1}}^2\right)\prod_{s=j}^{l}\frac{d_s{b_s}^2(k_s-s_s+1)^2}{16p^2}\end{aligned}$$

where ① holds since for an arbitrary vector $\boldsymbol{u}\in\mathbb{R}^k$ and an arbitrary matrix $\boldsymbol{M}\in\mathbb{R}^{k\times k}$, we have $(\boldsymbol{u}\otimes \boldsymbol{I}_k)\boldsymbol{M} = \boldsymbol{u}\otimes\boldsymbol{M}$; ② holds since we use Lemma 12 and the assumption that $\|\boldsymbol{W}_{(l+1)}\|_F^2 \le {b_{l+1}}^2$.

Now we consider the least square loss and softmax activation function. In such a case, we can further obtain:

$$\begin{aligned}&\left\|\frac{\partial^2 f(\boldsymbol{w},\boldsymbol{D})}{\partial \boldsymbol{w}_{(j)}\partial \boldsymbol{w}_{(l+1)}}\right\|_F^2 = \left\|\frac{\partial (\boldsymbol{v}-\boldsymbol{y})\mathsf{G}(\boldsymbol{u})\boldsymbol{z}_{(l)}^T}{\partial \boldsymbol{w}_{(j)}}\right\|_F^2\\ =&\left\|\left[\boldsymbol{I}_{\tilde{r}_l\tilde{c}_l d_l}\otimes(\boldsymbol{v}-\boldsymbol{y})\right]\frac{\partial \boldsymbol{z}_{(l)}^T}{\partial \boldsymbol{x}_{(l)}}\frac{\partial \boldsymbol{x}_{(l)}}{\partial \boldsymbol{w}_{(j)}} + \left[\boldsymbol{z}_{(l)}\otimes(\boldsymbol{v}-\boldsymbol{y})\right]\frac{\partial \mathsf{vec}\left(\mathsf{G}(\boldsymbol{u})\right)}{\partial \boldsymbol{u}}\frac{\partial \boldsymbol{u}}{\partial \boldsymbol{z}_{(l)}}\frac{\partial \boldsymbol{z}_{(l)}}{\partial \boldsymbol{x}_{(l)}}\frac{\partial \boldsymbol{x}_{(l)}}{\partial \boldsymbol{w}_{(j)}} + \left[\boldsymbol{z}_{(l)}\otimes \boldsymbol{I}_{d_{l+1}}\right]\frac{\partial \boldsymbol{v}}{\partial \boldsymbol{z}_{(l)}}\frac{\partial \boldsymbol{z}_{(l)}}{\partial \boldsymbol{x}_{(l)}}\frac{\partial \boldsymbol{x}_{(l)}}{\partial \boldsymbol{w}_{(j)}}\right\|_F^2\\ =&\left\|\widetilde{\mathsf{up}}\left(\left[\boldsymbol{I}_{\tilde{r}_l\tilde{c}_l d_l}\otimes(\boldsymbol{v}-\boldsymbol{y})\right] + \left[\boldsymbol{z}_{(l)}\otimes(\boldsymbol{v}-\boldsymbol{y})\right]\mathsf{Q}(\boldsymbol{u})\boldsymbol{W}_{(l+1)} + \left[\boldsymbol{z}_{(l)}\otimes \boldsymbol{I}_{d_{l+1}}\right]\mathsf{Q}(\boldsymbol{u})\mathsf{G}(\boldsymbol{u})\boldsymbol{W}_{(l+1)}\right)\odot \boldsymbol{G}_{(l)}\frac{\partial \boldsymbol{x}_{(l)}}{\partial \boldsymbol{w}_{(j)}}\right\|_F^2.\end{aligned}$$

Thus, we can further obtain

$$\begin{aligned}
&\left\|\frac{\partial^2 f(\boldsymbol{w},\boldsymbol{D})}{\partial \boldsymbol{w}_{(j)}\partial \boldsymbol{w}_{(l+1)}}\right\|_F^2\\
\leq&\frac{1}{16p^2}\left\|\left[\boldsymbol{I}_{\tilde{r}_l\tilde{c}_l d_l}\otimes(\boldsymbol{v}-\boldsymbol{y})\right]+\left[\boldsymbol{z}_{(l)}\otimes(\boldsymbol{v}-\boldsymbol{y})\right]\mathsf{Q}\left(\boldsymbol{u}\right)\boldsymbol{W}_{(l+1)}+\left[\boldsymbol{z}_{(l)}\otimes\boldsymbol{I}_{d_{l+1}}\right]\mathsf{Q}\left(\boldsymbol{u}\right)\mathsf{G}\left(\boldsymbol{u}\right)\boldsymbol{W}_{(l+1)}\right\|_F^2\left\|\frac{\partial \boldsymbol{x}_{(l)}}{\partial \boldsymbol{w}_{(j)}}\right\|_F^2\\
\leq&\frac{3}{16p^2}\left(\left\|\boldsymbol{I}_{\tilde{r}_l\tilde{c}_l d_l}\otimes(\boldsymbol{v}-\boldsymbol{y})\right\|_F^2+\left\|\left[\boldsymbol{z}_{(l)}\otimes(\boldsymbol{v}-\boldsymbol{y})\right]\mathsf{Q}\left(\boldsymbol{u}\right)\boldsymbol{W}_{(l+1)}\right\|_F^2+\left\|\left[\boldsymbol{z}_{(l)}\otimes\boldsymbol{I}_{d_{l+1}}\right]\mathsf{Q}\left(\boldsymbol{u}\right)\mathsf{G}\left(\boldsymbol{u}\right)\boldsymbol{W}_{(l+1)}\right\|_F^2\right)\left\|\frac{\partial \boldsymbol{x}_{(l)}}{\partial \boldsymbol{w}_{(j)}}\right\|_F^2\\
\overset{①}{\leq}&\frac{3}{16p^2}\tilde{r}_l\tilde{c}_l d_l\left(2+\frac{3}{100}{b_{l+1}}^2\right)d_l r_l c_l\tilde{r}_{j-1}\tilde{c}_{j-1}d_{j-1}(k_j-s_j+1)^2\prod_{s=j+1}^{l}\frac{d_s {b_s}^2(k_s-s_s+1)^2}{16p^2}\\
=&\frac{3d_{j-1}}{p^4{b_j}^2 d_j}r_{j-1}c_{j-1}d_l^2 r_l^2 c_l^2\left(2+\frac{2^7}{3^8}{b_{l+1}}^2+\frac{2^6}{16\cdot 3^8}{b_{l+1}}^2\right)\prod_{s=j}^{l}\frac{d_s {b_s}^2(k_s-s_s+1)^2}{16p^2},
\end{aligned}$$

where ① holds since we use Lemma 12 and the fact that $\|\boldsymbol{W}_{(l+1)}\|_F^2\leq {b_{l+1}}^2$.

**Case 4:** $i<j=l+1$

Similar to the Case 2, we also can have

$$\left\|\frac{\partial^2 f(\boldsymbol{w},\boldsymbol{D})}{\partial \boldsymbol{w}_{(i)}^T\partial \boldsymbol{w}_{(j)}}\right\|_F^2=\left\|\frac{\partial^2 f(\boldsymbol{w},\boldsymbol{D})}{\partial \boldsymbol{w}_{(j)}^T\partial \boldsymbol{w}_{(i)}}\right\|_F^2.$$

So in this case, we can just directly use the bound in case 3 to bound $\left\|\frac{\partial^2 f(\boldsymbol{w},\boldsymbol{D})}{\partial \boldsymbol{w}_{(i)}^T\partial \boldsymbol{w}_{(j)}}\right\|_F^2$.

**Case 5:** $i=j=l+1$

In the following, we first consider the first case, *i.e.* $i=l+1$, and bound

$$\begin{aligned}
\left\|\frac{\partial^2 f(\boldsymbol{w},\boldsymbol{D})}{\partial \boldsymbol{w}_{(l+1)}\partial \boldsymbol{w}_{(l+1)}}\right\|_F^2=&\left\|\frac{\partial(\boldsymbol{v}-\boldsymbol{y})\boldsymbol{z}_{(l)}^T}{\partial \boldsymbol{w}_{(l+1)}}\right\|_F^2=\left\|\left[\boldsymbol{z}_{(l)}\otimes\boldsymbol{I}_{d_{l+1}}\right]\frac{\partial \boldsymbol{v}}{\partial \boldsymbol{w}_{(l+1)}}\right\|_F^2\\
=&\left\|\left[\boldsymbol{z}_{(l)}\otimes\boldsymbol{I}_{d_{l+1}}\right]\mathsf{G}\left(\boldsymbol{u}\right)\left[\boldsymbol{z}_{(l)}\otimes\boldsymbol{I}_{d_{l+1}}\right]^T\right\|_F^2\\
\overset{①}{=}&\left\|\left[\boldsymbol{z}_{(l)}\left(\boldsymbol{z}_{(l)}\otimes\mathsf{G}\left(\boldsymbol{u}\right)\right)^T\right]^T\right\|_F^2\\
\leq&\left\|\boldsymbol{z}_{(l)}\right\|_F^4\left\|\mathsf{G}\left(\boldsymbol{u}\right)\right\|_F^2\leq\frac{1}{16}\tilde{r}_l^2\tilde{c}_l^2 d_l^2 d_{l+1},
\end{aligned}$$

where ① holds since for an arbitrary vector $\boldsymbol{u}\in\mathbb{R}^k$ and an arbitrary matrix $\boldsymbol{M}\in\mathbb{R}^{k\times k}$, we have $\left(\boldsymbol{u}\otimes\boldsymbol{I}_k\right)\boldsymbol{M}=\boldsymbol{u}\otimes\boldsymbol{M}$.

Now we can bound $\left\|\frac{\partial^2 f(\boldsymbol{w},\boldsymbol{D})}{\partial \boldsymbol{w}\partial \boldsymbol{w}}\right\|_F^2$ as follows:

$$\begin{aligned}
\left\|\frac{\partial^2 f(\boldsymbol{w},\boldsymbol{D})}{\partial \boldsymbol{w}\partial \boldsymbol{w}}\right\|_F^2=&\left\|\frac{\partial^2 f(\boldsymbol{w},\boldsymbol{D})}{\partial \boldsymbol{w}_{(l+1)}\partial \boldsymbol{w}_{(l+1)}}\right\|_F^2+2\sum_{j=1}^{l}\left\|\frac{\partial^2 f(\boldsymbol{w},\boldsymbol{D})}{\partial \boldsymbol{w}_{(j)}\partial \boldsymbol{w}_{(l+1)}}\right\|_F^2+2\sum_{j=1}^{l}\sum_{i=j}^{l}\left\|\frac{\partial^2 f(\boldsymbol{w},\boldsymbol{D})}{\partial \boldsymbol{w}_{(j)}\partial \boldsymbol{w}_{(i)}}\right\|_F^2\\
\leq&\mathcal{O}\left(\frac{l^2}{{k_1}^4}\max_{1\leq i,j\leq l}\tilde{r}_i\tilde{c}_i r_j c_j\frac{{b_{l+1}}^4}{16p^2}\left(\frac{{r_w}^2}{8\sqrt{2}p^2}\right)^{2l-2}\prod_{s=1}^{l}(d_s{k_s}^2)^2\right)\\
\leq&\mathcal{O}\left(\frac{\vartheta {b_{l+1}}^2 d_0^2}{{b_1}^4 d_1^2}l^2 r_0^2 c_0^2\left[\prod_{s=1}^{l}\frac{d_s{b_s}^2(k_s-s_s+1)^2}{8\sqrt{2}p^2}\right]^2\right).
\end{aligned}$$

On the other hand, if the activation functions $\sigma_1$ and $\sigma_2$ are respectively sigmoid function and softmax function, $f(\boldsymbol{w},\boldsymbol{D})$ is infinitely differentiable. Also $\sigma(a)$, $\sigma'(a)$, $\sigma''(a)$ and $\sigma'''(a)$ are all bounded. This means that $\nabla^3_{\boldsymbol{w}} f(\boldsymbol{w},\boldsymbol{D})$ exists. Also since input $\boldsymbol{D}$ and the parameter $\boldsymbol{w}$ are bounded, we can always find a universal constant $\nu$ such that

$$\|\nabla^3_{\boldsymbol{w}} f(\boldsymbol{w},\boldsymbol{D})\|_{\text{op}} = \sup_{\|\boldsymbol{\lambda}\|_2\leq 1} \left\langle \boldsymbol{\lambda}^{\otimes^3}, \nabla^3_{\boldsymbol{w}} f(\boldsymbol{w},\boldsymbol{D})\right\rangle = \sum_{i,j,k} [\nabla^3_{\boldsymbol{w}} f(\boldsymbol{w},\boldsymbol{D})]_{ijk}\boldsymbol{\lambda}_i\boldsymbol{\lambda}_j\boldsymbol{\lambda}_k \leq \nu < +\infty.$$

The proof is completed. □

**Lemma 15.** *Suppose that the activation function $\sigma_1$ is sigmoid and $\sigma_2$ is softmax, and the loss function $f(\boldsymbol{w},\boldsymbol{D})$ is squared loss. Suppose Assumption 1 on the input data $\boldsymbol{D}$ holds. Then for any $t>0$, the objective $f(\boldsymbol{w},\boldsymbol{x})$ obeys*

$$\mathbb{P}\left(\frac{1}{n}\sum_{i=1}^{n}\Big(f(\boldsymbol{w},\boldsymbol{D}^{(i)})-\mathbb{E}(f(\boldsymbol{w},\boldsymbol{D}^{(i)}))\Big)>t\right) \leq 2\exp\left(-\frac{2nt^2}{\alpha^2}\right),$$

*where $\alpha=1$.*

*Proof.* Since the input $\boldsymbol{D}^{(i)}$ $(i=1,\cdots,n)$ are independent from each other, then the output $f(\boldsymbol{w},\boldsymbol{D}^{(i)})$ $(i=1,\cdots,n)$ are also independent. Meanwhile, when the loss is the square loss, we can easily bound $0\leq f(\boldsymbol{w},\boldsymbol{D}^{(i)})=\frac{1}{2}\|\boldsymbol{v}-\boldsymbol{y}\|_2^2\leq 1$, since the value of entries in $\boldsymbol{v}$ belongs to $[0,1]$ and $\boldsymbol{y}$ is a one-hot vector label of $\boldsymbol{v}$.

Besides, for arbitrary random variable $x$, $|x-\mathbb{E}x|\leq |x|$. So by Hoeffding's inequality in Lemma 6, we have

$$\mathbb{P}\left(\frac{1}{n}\sum_{i=1}^{n}\Big(f(\boldsymbol{w},\boldsymbol{D}^{(i)})-\mathbb{E}(f(\boldsymbol{w},\boldsymbol{D}^{(i)}))\Big)>t\right) \leq \exp\left(-\frac{2nt^2}{\alpha^2}\right),$$

where $\alpha=1$. This means that $\frac{1}{n}\sum_{i=1}^{n}\big(f(\boldsymbol{w},\boldsymbol{D}^{(i)})-\mathbb{E}(f(\boldsymbol{w},\boldsymbol{D}^{(i)}))\big)$ has exponential tails. □

**Lemma 16.** *Suppose that the activation function $\sigma_1$ is sigmoid and $\sigma_2$ is softmax, and the loss function $f(\boldsymbol{w},\boldsymbol{D})$ is squared loss. Suppose Assumption 1 on the input data $\boldsymbol{D}$ holds. Then for any $t>0$ and arbitrary unit vector $\boldsymbol{\lambda}\in\mathbb{S}^{d-1}$, the gradient $\nabla f(\boldsymbol{w},\boldsymbol{x})$ obeys*

$$\mathbb{P}\left(\frac{1}{n}\sum_{i=1}^{n}\left(\left\langle\boldsymbol{\lambda},\nabla_{\boldsymbol{w}} f(\boldsymbol{w},\boldsymbol{D}^{(i)})-\mathbb{E}_{\boldsymbol{D}\sim\boldsymbol{\mathcal{D}}}\nabla_{\boldsymbol{w}} f(\boldsymbol{w},\boldsymbol{D}^{(i)})\right\rangle\right)>t\right) \leq \exp\left(-\frac{nt^2}{2\beta^2}\right).$$

*where $\beta\triangleq\left[\vartheta\tilde{r}_l\tilde{c}_l d_l+\sum_{i=1}^{l}\frac{\vartheta {b_{l+1}}^2 d_{i-1}}{p^2{b_i}^2 d_i} r_{i-1}c_{i-1}\prod_{s=i}^{l}\frac{d_s{b_s}^2(k_s-s_s+1)^2}{16p^2}\right]^{1/2}$ in which $\vartheta=1/8$.*

*Proof.* Since the input $\boldsymbol{D}^{(i)}$ $(i=1,\cdots,n)$ are independent from each other, then the output $\nabla_{\boldsymbol{w}} f(\boldsymbol{w},\boldsymbol{D}^{(i)})$ $(i=1,\cdots,n)$ are also independent. Furthermore, for arbitrary vector $\boldsymbol{x}$, $\|\boldsymbol{x}-\mathbb{E}\boldsymbol{x}\|_2^2\leq\|\boldsymbol{x}\|_2^2$. Hence, for an arbitrary unit vector $\boldsymbol{\lambda}\in\mathbb{S}^{d-1}$ where $d=\tilde{r}_l\tilde{c}_l d_l d_{l+1}+\sum_{i=1}^{l}{k_i}^2 d_{i-1}d_i$, we have

$$\begin{aligned}\langle\boldsymbol{\lambda},\nabla_{\boldsymbol{w}} f(\boldsymbol{w},\boldsymbol{D}^{(i)})-\mathbb{E}_{\boldsymbol{D}\sim\boldsymbol{\mathcal{D}}}\nabla_{\boldsymbol{w}} f(\boldsymbol{w},\boldsymbol{D}^{(i)})\rangle \leq&\|\boldsymbol{\lambda}\|_2\|\nabla_{\boldsymbol{w}} f(\boldsymbol{w},\boldsymbol{D}^{(i)})-\mathbb{E}_{\boldsymbol{D}\sim\boldsymbol{\mathcal{D}}}\nabla_{\boldsymbol{w}} f(\boldsymbol{w},\boldsymbol{D}^{(i)})\|_2\\ \leq&\|\boldsymbol{\lambda}\|_2\|\nabla_{\boldsymbol{w}} f(\boldsymbol{w},\boldsymbol{D}^{(i)})\|_2\overset{①}{\leq}\beta,\end{aligned}$$

where ① holds since $\|\boldsymbol{\lambda}\|_2=1$ $(\boldsymbol{\lambda}\in\mathbb{S}^{d-1})$ and by Lemma 11, we have $\|\nabla_{\boldsymbol{w}} f(\boldsymbol{w},\boldsymbol{D}^{(i)})\|\leq\beta$ where $\beta\triangleq\left[\vartheta\tilde{r}_l\tilde{c}_l d_l+\sum_{i=1}^{l}\frac{\vartheta {b_{l+1}}^2 d_{i-1}}{p^2{b_i}^2 d_i} r_{i-1}c_{i-1}\prod_{s=i}^{l}\frac{d_s{b_s}^2(k_s-s_s+1)^2}{16p^2}\right]^{1/2}$ in which $\vartheta=1/8$.

Thus, we can use Hoeffding's inequality in Lemma 6 to bound

$$\mathbb{P}\left(\frac{1}{n}\sum_{i=1}^{n}\left(\left\langle\boldsymbol{\lambda},\nabla_{\boldsymbol{w}} f(\boldsymbol{w},\boldsymbol{D}^{(i)})-\mathbb{E}_{\boldsymbol{D}\sim\boldsymbol{\mathcal{D}}}\nabla_{\boldsymbol{w}} f(\boldsymbol{w},\boldsymbol{D}^{(i)})\right\rangle\right)>t\right) \leq \exp\left(-\frac{nt^2}{2\beta^2}\right).$$

The proof is completed. □

**Lemma 17.** *Suppose that the activation function $\sigma_1$ is sigmoid and $\sigma_2$ is softmax, and the loss function $f(\boldsymbol{w},\boldsymbol{D})$ is squared loss. Suppose that Assumption 1 on the input data $\boldsymbol{D}$ and the parameter $\boldsymbol{w}$ holds. Then for any $t>0$ and arbitrary unit vector $\boldsymbol{\lambda}\in\mathbb{S}^{d-1}$, the Hessian $\nabla^2 f(\boldsymbol{w},\boldsymbol{D})$ obeys*

$$\mathbb{P}\left(\frac{1}{n}\sum_{i=1}^{n}\left(\left\langle \boldsymbol{\lambda},(\nabla^2_{\boldsymbol{w}} f(\boldsymbol{w},\boldsymbol{D}^{(i)})-\mathbb{E}_{\boldsymbol{D}\sim\boldsymbol{\mathcal{D}}}\nabla^2_{\boldsymbol{w}} f(\boldsymbol{w},\boldsymbol{D}^{(i)}))\boldsymbol{\lambda}\right\rangle\right)>t\right)\leq 2\exp\left(-\frac{nt^2}{2\gamma^2}\right).$$

*where* $\gamma=\left(\frac{\vartheta {b_{l+1}}^2 d_0^2}{{b_1}^4 d_1^2} l^2 r_0^2 c_0^2\left[\prod_{s=1}^{l}\frac{d_s {b_s}^2(k_s-s_s+1)^2}{8\sqrt{2}p^2}\right]^2\right)^{1/2}$.

*Proof.* Since the input $\boldsymbol{D}^{(i)}$ $(i=1,\cdots,n)$ are independent from each other, then the output $\nabla_{\boldsymbol{w}} f(\boldsymbol{w},\boldsymbol{D}^{(i)})$ $(i=1,\cdots,n)$ are also independent. On the other hand, for arbitrary random matrix $\boldsymbol{X}$, $\|\boldsymbol{X}-\mathbb{E}\boldsymbol{X}\|_F^2\leq\|\boldsymbol{X}\|_F^2$. Thus, for an arbitrary unit vector $\boldsymbol{\lambda}\in\mathbb{S}^{d-1}$ where $d=\tilde{r}_l\tilde{c}_l d_l d_{l+1}+\sum_{i=1}^{l}{k_i}^2 d_{i-1}d_i$, we have

$$\begin{aligned}\langle\boldsymbol{\lambda},(\nabla^2_{\boldsymbol{w}} f(\boldsymbol{w},\boldsymbol{D}^{(i)})-\mathbb{E}_{\boldsymbol{D}\sim\boldsymbol{\mathcal{D}}}\nabla^2_{\boldsymbol{w}} f(\boldsymbol{w},\boldsymbol{D}^{(i)}))\boldsymbol{\lambda}\rangle \leq&\|\boldsymbol{\lambda}\|_2\|(\nabla^2_{\boldsymbol{w}} f(\boldsymbol{w},\boldsymbol{D}^{(i)})-\mathbb{E}_{\boldsymbol{D}\sim\boldsymbol{\mathcal{D}}}\nabla^2_{\boldsymbol{w}} f(\boldsymbol{w},\boldsymbol{D}^{(i)}))\boldsymbol{\lambda}\|_2\\ \leq&\|\nabla^2_{\boldsymbol{w}} f(\boldsymbol{w},\boldsymbol{D}^{(i)})-\mathbb{E}_{\boldsymbol{D}\sim\boldsymbol{\mathcal{D}}}\nabla^2_{\boldsymbol{w}} f(\boldsymbol{w},\boldsymbol{D}^{(i)})\|_{\text{op}}\|\boldsymbol{\lambda}\|_2\\ \leq&\|\nabla^2_{\boldsymbol{w}} f(\boldsymbol{w},\boldsymbol{D}^{(i)})-\mathbb{E}_{\boldsymbol{D}\sim\boldsymbol{\mathcal{D}}}\nabla^2_{\boldsymbol{w}} f(\boldsymbol{w},\boldsymbol{D}^{(i)})\|_F\\ \leq&\|\nabla^2_{\boldsymbol{w}} f(\boldsymbol{w},\boldsymbol{D}^{(i)})\|_F\\ \overset{①}{\leq}&\gamma,\end{aligned}$$

where ① holds since $\|\boldsymbol{\lambda}\|_2=1$ $(\boldsymbol{\lambda}\in\mathbb{S}^{d-1})$ and by Lemma 14, we have $\|\nabla^2_{\boldsymbol{w}} f(\boldsymbol{w},\boldsymbol{D}^{(i)})\|\leq\gamma$ where $\gamma=\left(\frac{\vartheta {b_{l+1}}^2 d_0^2}{{b_1}^4 d_1^2} l^2 r_0^2 c_0^2\left[\prod_{s=1}^{l}\frac{d_s {b_s}^2(k_s-s_s+1)^2}{8\sqrt{2}p^2}\right]^2\right)^{1/2}$.

Thus, we can use Hoeffding's inequality in Lemma 6 to bound

$$\mathbb{P}\left(\frac{1}{n}\sum_{i=1}^{n}\left(\left\langle \boldsymbol{\lambda},(\nabla^2_{\boldsymbol{w}} f(\boldsymbol{w},\boldsymbol{D}^{(i)})-\mathbb{E}_{\boldsymbol{D}\sim\boldsymbol{\mathcal{D}}}\nabla^2_{\boldsymbol{w}} f(\boldsymbol{w},\boldsymbol{D}^{(i)}))\boldsymbol{\lambda}\right\rangle\right)>t\right)\leq \exp\left(-\frac{nt^2}{2\gamma^2}\right).$$

The proof is completed. □

**Lemma 18.** *Suppose that the activation function $\sigma_1$ is sigmoid and $\sigma_2$ is softmax, and the loss function $f(\boldsymbol{w},\boldsymbol{D})$ is squared loss. Suppose that Assumption 1 on the input data $\boldsymbol{D}$ and the parameter $\boldsymbol{w}$ holds. Then the empirical Hessian converges uniformly to the population Hessian in operator norm. Specifically, there exit two universal constants $c_{v'}$ and $c_v$ such that if $n\geq c_{v'}\frac{\nu^2}{d\varrho\varepsilon^2}\left[\prod_{s=1}^{l}\frac{d_s {b_s}^2(k_s-s_s+1)^2}{8\sqrt{2}p^2}\right]^{-1}$, then with probability at least $1-\epsilon$*

$$\sup_{\boldsymbol{w}\in\Omega}\left\|\nabla^2\widetilde{\boldsymbol{Q}}_n(\boldsymbol{w})-\nabla^2\boldsymbol{Q}(\boldsymbol{w})\right\|_{op}\leq c_v\gamma\sqrt{\frac{2d+\theta\varrho+\log\left(\frac{4}{\varepsilon}\right)}{2n}},$$

*holds with probability at least $1-\varepsilon$, where $d=\tilde{r}_l\tilde{c}_l d_l d_{l+1}+\sum_{i=1}^{l}{k_i}^2 d_{i-1}d_i$, $\theta=a_{l+1}(d_{l+1}+\tilde{r}_l\tilde{c}_l d_l-2a_{l+1}+1)+\sum_{i=1}^{l}a_i({k_i}^2 d_i+d_{i-1}-2a_i+1)$, $\varrho=\sum_{i=1}^{l}\log\left(\frac{\sqrt{d_i}b_i(k_i-s_i+1)}{4p}\right)+\log(b_{l+1})+\log\left(\frac{n}{128p^2}\right)$, and $\gamma=\left(\frac{\vartheta {b_{l+1}}^2 d_0^2}{{b_1}^4 d_1^2} l^2 r_0^2 c_0^2\left[\prod_{s=1}^{l}\frac{d_s {b_s}^2(k_s-s_s+1)^2}{8\sqrt{2}p^2}\right]^2\right)^{1/2}$.*

*Proof.* Recall that the weight of each kernel and the feature maps has magnitude bound separately, *i.e.* $\boldsymbol{w}_{(i)}^k\in\mathsf{B}^{{k_i}^2 d_{i-1}}(r_w)$ $(i=1,\cdots,l;k=1,\cdots,d_i)$ and $\boldsymbol{w}_{(l+1)}\in\mathsf{B}^{\tilde{r}_l\tilde{c}_l d_l d_{l+1}}(b_{l+1})$. Since $\widetilde{\boldsymbol{W}}_{(i)}=[\mathsf{vec}(\boldsymbol{W}_{(i)}^1),\mathsf{vec}(\boldsymbol{W}_{(i)}^2),\cdots,\mathsf{vec}(\boldsymbol{W}_{(i)}^{d_{i-1}})]\in\mathbb{R}^{k_i^2 d_i\times d_{i-1}}$, we have $\|\widetilde{\boldsymbol{W}}_{(i)}\|_F\leq d_i b_i$.

So here we assume $\widetilde{\boldsymbol{W}}_{(i,\epsilon)}$ is the $d_i b_i \epsilon/(b_{l+1}+\sum_{i=1}^{l} d_i b_i)$-covering net of the matrix $\widetilde{\boldsymbol{W}}_{(i)}$ which is the set of all parameters in the $i$-th layer. Then by Lemma 7, we have the covering number

$$n_\epsilon{}^i \leq \left(\frac{9(b_{l+1}+\sum_{i=1}^{l} d_i b_i)}{\epsilon}\right)^{a_i(k_i{}^2 d_i + d_{i-1} - 2a_i + 1)},$$

since the rank of $\widetilde{\boldsymbol{W}}_{(i)}$ obeys $\mathsf{rank}(\widetilde{\boldsymbol{W}}_{(i)}) \leq a_i$ for $1 \leq i \leq l$. For the last layer, we also can construct an $b_{l+1}\epsilon/(b_{l+1} + \sum_{i=1}^{l} d_i b_i)$-covering net for the weight matrix $\boldsymbol{W}_{(l+1)}$. Here we have

$$n_\epsilon{}^{l+1} \leq \left(\frac{9(b_{l+1}+\sum_{i=1}^{l} d_i b_i)}{\epsilon}\right)^{a_{l+1}(d_{l+1} + \tilde{r}_l \tilde{c}_l d_l - 2a_{l+1} + 1)},$$

since the rank of $\boldsymbol{W}_{(l+1)}$ obeys $\mathsf{rank}(\boldsymbol{W}_{(l+1)}) \leq a_{l+1}$. Finally, we arrange them together to construct a set $\Theta$ and claim that there is always an $\epsilon$-covering net $\boldsymbol{w}_\epsilon$ in $\Theta$ for any parameter $\boldsymbol{w}$. Accordingly, we have

$$|\Theta| \leq \prod_{i=1}^{l+1} n_\epsilon{}^i = \left(\frac{9(b_{l+1}+\sum_{i=1}^{l} d_i b_i)}{\epsilon}\right)^{a_{l+1}(d_{l+1} + \tilde{r}_l \tilde{c}_l d_l - 2a_{l+1} + 1) + \sum_{i=1}^{l} a_i(k_i{}^2 d_i + d_{i-1} - 2a_i + 1)} = \left(\frac{9(b_{l+1}+\sum_{i=1}^{l} d_i b_i)}{\epsilon}\right)^{\theta},$$

where $\theta = a_{l+1}(d_{l+1} + \tilde{r}_l \tilde{c}_l d_l - 2a_{l+1} + 1) + \sum_{i=1}^{l} a_i(k_i{}^2 d_i + d_{i-1} - 2a_i + 1)$ which is the total freedom degree of the network. So we can always find a vector $\boldsymbol{w}_{k_{\boldsymbol{w}}} \in \Theta$ such that $\|\boldsymbol{w} - \boldsymbol{w}_{k_{\boldsymbol{w}}}\|_2 \leq \epsilon$. Now we use the decomposition strategy to bound our goal:

$$\begin{aligned}
&\left\|\nabla^2 \widetilde{\boldsymbol{Q}}_n(\boldsymbol{w}) - \nabla^2 \boldsymbol{Q}(\boldsymbol{w})\right\|_{\text{op}} \\
=&\left\|\frac{1}{n}\sum_{i=1}^{n} \nabla^2 f(\boldsymbol{w}, \boldsymbol{D}^{(i)}) - \mathbb{E}_{\boldsymbol{D}\sim\boldsymbol{\mathcal{D}}}(\nabla^2 f(\boldsymbol{w}, \boldsymbol{D}))\right\|_{\text{op}} \\
=&\left\|\frac{1}{n}\sum_{i=1}^{n}\left(\nabla^2 f(\boldsymbol{w}, \boldsymbol{D}^{(i)}) - \nabla^2 f(\boldsymbol{w}_{k_{\boldsymbol{w}}}, \boldsymbol{D}^{(i)})\right) + \frac{1}{n}\sum_{i=1}^{n} \nabla^2 f(\boldsymbol{w}_{k_{\boldsymbol{w}}}, \boldsymbol{D}^{(i)}) - \mathbb{E}(\nabla^2 f(\boldsymbol{w}_{k_{\boldsymbol{w}}}, \boldsymbol{D}))\right. \\
&\left. + \mathbb{E}_{\boldsymbol{D}\sim\boldsymbol{\mathcal{D}}}(\nabla^2 f(\boldsymbol{w}_{k_{\boldsymbol{w}}}, \boldsymbol{D})) - \mathbb{E}_{\boldsymbol{D}\sim\boldsymbol{\mathcal{D}}}(\nabla^2 f(\boldsymbol{w}, \boldsymbol{D}))\right\|_{\text{op}} \\
\leq&\left\|\frac{1}{n}\sum_{i=1}^{n}\left(\nabla^2 f(\boldsymbol{w}, \boldsymbol{D}^{(i)}) - \nabla^2 f(\boldsymbol{w}_{k_{\boldsymbol{w}}}, \boldsymbol{D}^{(i)})\right)\right\|_{\text{op}} + \left\|\frac{1}{n}\sum_{i=1}^{n} \nabla^2 f(\boldsymbol{w}_{k_{\boldsymbol{w}}}, \boldsymbol{D}^{(i)}) - \mathbb{E}_{\boldsymbol{D}\sim\boldsymbol{\mathcal{D}}}(\nabla^2 f(\boldsymbol{w}_{k_{\boldsymbol{w}}}, \boldsymbol{D}))\right\|_{\text{op}} \\
&+ \left\|\mathbb{E}_{\boldsymbol{D}\sim\boldsymbol{\mathcal{D}}}(\nabla^2 f(\boldsymbol{w}_{k_{\boldsymbol{w}}}, \boldsymbol{D})) - \mathbb{E}_{\boldsymbol{D}\sim\boldsymbol{\mathcal{D}}}(\nabla^2 f(\boldsymbol{w}, \boldsymbol{D}))\right\|_{\text{op}}.
\end{aligned}$$

Here we also define four events $\boldsymbol{E}_0$, $\boldsymbol{E}_1$, $\boldsymbol{E}_2$ and $\boldsymbol{E}_3$ as

$$\begin{aligned}
\boldsymbol{E}_0 &= \left\{\sup_{\boldsymbol{w}\in\Omega} \left\|\nabla^2 \widetilde{\boldsymbol{Q}}_n(\boldsymbol{w}) - \nabla^2 \boldsymbol{Q}(\boldsymbol{w})\right\|_{\text{op}} \geq t\right\}, \\
\boldsymbol{E}_1 &= \left\{\sup_{\boldsymbol{w}\in\Omega} \left\|\frac{1}{n}\sum_{i=1}^{n}\left(\nabla^2 f(\boldsymbol{w}, \boldsymbol{D}^{(i)}) - \nabla^2 f(\boldsymbol{w}_{k_{\boldsymbol{w}}}, \boldsymbol{D}^{(i)})\right)\right\|_{\text{op}} \geq \frac{t}{3}\right\}, \\
\boldsymbol{E}_2 &= \left\{\sup_{\boldsymbol{w}_{k_{\boldsymbol{w}}}\in\Theta} \left\|\frac{1}{n}\sum_{i=1}^{n} \nabla^2 f(\boldsymbol{w}_{k_{\boldsymbol{w}}}, \boldsymbol{D}^{(i)}) - \mathbb{E}_{\boldsymbol{D}\sim\boldsymbol{\mathcal{D}}}(\nabla^2 f(\boldsymbol{w}_{k_{\boldsymbol{w}}}, \boldsymbol{D}))\right\|_{\text{op}} \geq \frac{t}{3}\right\}, \\
\boldsymbol{E}_3 &= \left\{\sup_{\boldsymbol{w}\in\Omega} \left\|\mathbb{E}_{\boldsymbol{D}\sim\boldsymbol{\mathcal{D}}}(\nabla^2 f(\boldsymbol{w}_{k_{\boldsymbol{w}}}, \boldsymbol{D})) - \mathbb{E}_{\boldsymbol{D}\sim\boldsymbol{\mathcal{D}}}(\nabla^2 f(\boldsymbol{w}, \boldsymbol{D}))\right\|_{\text{op}} \geq \frac{t}{3}\right\}.
\end{aligned}$$

Accordingly, we have

$$\mathbb{P}\left(\boldsymbol{E}_0\right) \leq \mathbb{P}\left(\boldsymbol{E}_1\right) + \mathbb{P}\left(\boldsymbol{E}_2\right) + \mathbb{P}\left(\boldsymbol{E}_3\right).$$

So we can respectively bound $\mathbb{P}\left(\boldsymbol{E}_1\right)$, $\mathbb{P}\left(\boldsymbol{E}_2\right)$ and $\mathbb{P}\left(\boldsymbol{E}_3\right)$ to bound $\mathbb{P}\left(\boldsymbol{E}_0\right)$.

**Step 1. Bound** $\mathbb{P}\left(\boldsymbol{E}_1\right)$: We first bound $\mathbb{P}\left(\boldsymbol{E}_1\right)$ as follows:

$$\begin{aligned}
\mathbb{P}\left(\boldsymbol{E}_1\right) =& \mathbb{P}\left(\sup_{\boldsymbol{w}\in\Omega}\left\|\frac{1}{n}\sum_{i=1}^{n}\left(\nabla^2 f(\boldsymbol{w},\boldsymbol{D}^{(i)}) - \nabla^2 f(\boldsymbol{w}_{k_{\boldsymbol{w}}},\boldsymbol{D}^{(i)})\right)\right\|_2 \geq \frac{t}{3}\right)\\
\overset{①}{\leq}& \frac{3}{t}\mathbb{E}_{\boldsymbol{D}\sim\boldsymbol{\mathcal{D}}}\left(\sup_{\boldsymbol{w}\in\Omega}\left\|\frac{1}{n}\sum_{i=1}^{n}\left(\nabla^2 f(\boldsymbol{w},\boldsymbol{D}^{(i)}) - \nabla^2 f(\boldsymbol{w}_{k_{\boldsymbol{w}}},\boldsymbol{D}^{(i)})\right)\right\|_2\right)\\
\leq& \frac{3}{t}\mathbb{E}_{\boldsymbol{D}\sim\boldsymbol{\mathcal{D}}}\left(\sup_{\boldsymbol{w}\in\Omega}\left\|\nabla^2 f(\boldsymbol{w},\boldsymbol{D}) - \nabla^2 f(\boldsymbol{w}_{k_{\boldsymbol{w}}},\boldsymbol{D})\right\|_2\right)\\
\leq& \frac{3}{t}\mathbb{E}_{\boldsymbol{D}\sim\boldsymbol{\mathcal{D}}}\left(\sup_{\boldsymbol{w}\in\Omega}\frac{\left\|\frac{1}{n}\sum_{i=1}^{n}\left(\nabla^2 f(\boldsymbol{w},\boldsymbol{D}^{(i)}) - \nabla^2 f(\boldsymbol{w}_{k_{\boldsymbol{w}}},\boldsymbol{D}^{(i)})\right)\right\|_2}{\left\|\boldsymbol{w}-\boldsymbol{w}_{k_{\boldsymbol{w}}}\right\|_2}\sup_{\boldsymbol{w}\in\Omega}\left\|\boldsymbol{w}-\boldsymbol{w}_{k_{\boldsymbol{w}}}\right\|_2\right)\\
\overset{②}{\leq}& \frac{3\nu\epsilon}{t},
\end{aligned}$$

where ① holds since by Markov inequality and ② holds because of Lemma 14. Therefore, we can set

$$t \geq \frac{6\nu\epsilon}{\varepsilon}.$$

Then we can bound $\mathbb{P}(\boldsymbol{E}_1)$:

$$\mathbb{P}(\boldsymbol{E}_1) \leq \frac{\varepsilon}{2}.$$

**Step 2. Bound** $\mathbb{P}\left(\boldsymbol{E}_2\right)$: By Lemma 3, we know that for any matrix $\boldsymbol{X}\in\mathbb{R}^{d\times d}$, its operator norm can be computed as

$$\|\boldsymbol{X}\|_{\text{op}} \leq \frac{1}{1-2\epsilon}\sup_{\boldsymbol{\lambda}\in\boldsymbol{\lambda}_\epsilon}\left|\langle\boldsymbol{\lambda},\boldsymbol{X}\boldsymbol{\lambda}\rangle\right|.$$

where $\boldsymbol{\lambda}_\epsilon = \{\boldsymbol{\lambda}_1,\ldots,\boldsymbol{\lambda}_{k_{\boldsymbol{w}}}\}$ be an $\epsilon$-covering net of $\mathsf{B}^d(1)$.

Let $\boldsymbol{\lambda}_{1/4}$ be the $\frac{1}{4}$-covering net of $\mathsf{B}^d(1)$, where $d = \tilde{r}_l\tilde{c}_l d_l d_{l+1} + \sum_{i=1}^{l} {k_i}^2 d_{i-1}d_i$. Recall that we use $\Theta$ to denote the $\epsilon$-net of $\boldsymbol{w}_{k_{\boldsymbol{w}}}$ and we have $|\Theta| \leq \prod_{i=1}^{l+1} {n_\epsilon}^i = \left(\frac{3(b_{l+1}+\sum_{i=1}^{l}d_i b_i)}{\epsilon}\right)^\theta$. Then we can bound $\mathbb{P}\left(\boldsymbol{E}_2\right)$ as follows:

$$\begin{aligned}
\mathbb{P}\left(\boldsymbol{E}_2\right) =& \mathbb{P}\left(\sup_{\boldsymbol{w}_{k_{\boldsymbol{w}}}\in\Theta}\left\|\frac{1}{n}\sum_{i=1}^{n}\nabla^2 f(\boldsymbol{w}_{k_{\boldsymbol{w}}},\boldsymbol{D}^{(i)}) - \mathbb{E}_{\boldsymbol{D}\sim\boldsymbol{\mathcal{D}}}(\nabla^2 f(\boldsymbol{w}_{k_{\boldsymbol{w}}},\boldsymbol{D}))\right\|_2 \geq \frac{t}{3}\right)\\
\leq& \mathbb{P}\left(\sup_{\boldsymbol{w}_{k_{\boldsymbol{w}}}\in\Theta,\boldsymbol{\lambda}\in\boldsymbol{\lambda}_{1/4}} 2\left|\left\langle\boldsymbol{\lambda},\left(\frac{1}{n}\sum_{i=1}^{n}\nabla^2 f(\boldsymbol{w}_{k_{\boldsymbol{w}}},\boldsymbol{D}^{(i)}) - \mathbb{E}_{\boldsymbol{D}\sim\boldsymbol{\mathcal{D}}}\left(\nabla^2 f(\boldsymbol{w}_{k_{\boldsymbol{w}}},\boldsymbol{D})\right)\right)\boldsymbol{\lambda}\right\rangle\right| \geq \frac{t}{3}\right)\\
\leq& 12^d\left(\frac{3(b_{l+1}+\sum_{i=1}^{l}d_i b_i)}{\epsilon}\right)^\theta \sup_{\boldsymbol{w}_{k_{\boldsymbol{w}}}\in\Theta,\boldsymbol{\lambda}\in\boldsymbol{\lambda}_{1/4}}\mathbb{P}\left(\left|\frac{1}{n}\sum_{i=1}^{n}\left\langle\boldsymbol{\lambda},\left(\nabla^2 f(\boldsymbol{w}_{k_{\boldsymbol{w}}},\boldsymbol{D}^{(i)}) - \mathbb{E}_{\boldsymbol{D}\sim\boldsymbol{\mathcal{D}}}\left(\nabla^2 f(\boldsymbol{w}_{k_{\boldsymbol{w}}},\boldsymbol{D})\right)\right)\boldsymbol{\lambda}\right\rangle\right| \geq \frac{t}{6}\right)\\
\overset{①}{\leq}& 12^d\left(\frac{3(b_{l+1}+\sum_{i=1}^{l}d_i b_i)}{\epsilon}\right)^\theta 2\exp\left(-\frac{nt^2}{72\gamma^2}\right),
\end{aligned}$$

where ① holds since by Lemma 17, we have

$$\mathbb{P}\left(\frac{1}{n}\sum_{i=1}^{n}\left(\left\langle\boldsymbol{\lambda},(\nabla^2_{\boldsymbol{w}} f(\boldsymbol{w},\boldsymbol{D}^{(i)}) - \mathbb{E}_{\boldsymbol{D}\sim\boldsymbol{\mathcal{D}}}\nabla^2_{\boldsymbol{w}} f(\boldsymbol{w},\boldsymbol{D}^{(i)}))\boldsymbol{\lambda}\right\rangle\right) > t\right) \leq \exp\left(-\frac{nt^2}{2\gamma^2}\right).$$

where $\gamma = \left( \frac{\vartheta {b_{l+1}}^2 d_0^2}{{b_1}^4 d_1^2} l^2 r_0^2 c_0^2 \left[ \prod_{s=1}^{l} \frac{d_s {b_s}^2 (k_s - s_s + 1)^2}{8\sqrt{2}p^2} \right]^2 \right)^{1/2}$.

Thus, if we set

$$t \geq \sqrt{\frac{72\gamma^2 \left( d\log(12) + \theta \log\left( \frac{3(b_{l+1} + \sum_{i=1}^{l} d_i b_i)}{\epsilon} \right) + \log\left(\frac{4}{\varepsilon}\right) \right)}{n}},$$

then we have

$$\mathbb{P}\left(\boldsymbol{E}_2\right) \leq \frac{\varepsilon}{2}.$$

**Step 3. Bound** $\mathbb{P}\left(\boldsymbol{E}_3\right)$: We first bound $\mathbb{P}\left(\boldsymbol{E}_3\right)$ as follows:

$$\begin{aligned}
\mathbb{P}\left(\boldsymbol{E}_3\right) =& \mathbb{P}\left( \sup_{\boldsymbol{w}\in\Omega} \left\| \mathbb{E}_{\boldsymbol{D}\sim\boldsymbol{\mathcal{D}}}(\nabla^2 f(\boldsymbol{w}_{k_{\boldsymbol{w}}}, \boldsymbol{D})) - \mathbb{E}_{\boldsymbol{D}\sim\boldsymbol{\mathcal{D}}}(\nabla^2 f(\boldsymbol{w}, \boldsymbol{D})) \right\|_2 \geq \frac{t}{3} \right) \\
\leq& \mathbb{P}\left( \mathbb{E}_{\boldsymbol{D}\sim\boldsymbol{\mathcal{D}}} \sup_{\boldsymbol{w}\in\Omega} \left\| (\nabla^2 f(\boldsymbol{w}_{k_{\boldsymbol{w}}}, \boldsymbol{D}) - \nabla^2 f(\boldsymbol{w}, \boldsymbol{D}) \right\|_2 \geq \frac{t}{3} \right) \\
\leq& \mathbb{P}\left( \sup_{\boldsymbol{w}\in\Omega} \frac{\left| \frac{1}{n}\sum_{i=1}^{n} \left( \nabla^2 f(\boldsymbol{w}, \boldsymbol{D}^{(i)}) - \nabla^2 f(\boldsymbol{w}_{k_{\boldsymbol{w}}}, \boldsymbol{D}^{(i)}) \right) \right|}{\|\boldsymbol{w} - \boldsymbol{w}_{k_{\boldsymbol{w}}}\|_2} \sup_{\boldsymbol{w}\in\Omega} \|\boldsymbol{w} - \boldsymbol{w}_{k_{\boldsymbol{w}}}\|_2 \geq \frac{t}{3} \right) \\
\overset{①}{\leq}& \mathbb{P}\left( \nu\epsilon \geq \frac{t}{3} \right),
\end{aligned}$$

where ① holds because of Lemma 14. We set $\epsilon$ enough small such that $\nu\epsilon < t/3$ always holds. Then it yields $\mathbb{P}\left(\boldsymbol{E}_3\right) = 0$.

**Step 4. Final result**: To ensure $\mathbb{P}(\boldsymbol{E}_0) \leq \varepsilon$, we just set $\epsilon = \frac{36(b_{l+1} + \sum_{i=1}^{l} d_i b_i)}{\vartheta^2 n b_{l+1}} \left[ \prod_{s=1}^{l} \frac{d_s {b_s}^2 (k_s - s_s + 1)^2}{8\sqrt{2}p^2} \right]^{-\frac{1}{2}}$. Note that $\frac{6\epsilon\nu}{\varepsilon} > 3\epsilon\nu$. Thus we can obtain

$$t \geq \max\left( \frac{6\nu\epsilon}{\varepsilon}, \sqrt{\frac{72\gamma^2 \left( d\log(12) + \theta \log\left( \frac{3(b_{l+1} + \sum_{i=1}^{l} d_i b_i)}{\epsilon} \right) + \log\left(\frac{4}{\varepsilon}\right) \right)}{n}} \right).$$

Thus, if $n \geq c_{v'} \frac{\nu^2}{d\varrho\varepsilon^2} \left[ \prod_{s=1}^{l} \frac{d_s {b_s}^2 (k_s - s_s + 1)^2}{8\sqrt{2}p^2} \right]^{-1}$ where $c_{v'}$ is a constant, there exists a universal constant $c_v$ such that

$$\begin{aligned}
\sup_{\boldsymbol{w}\in\Omega} \left\| \nabla^2 \widetilde{\boldsymbol{Q}}_n(\boldsymbol{w}) - \nabla^2 \boldsymbol{Q}(\boldsymbol{w}) \right\|_{\text{op}} \leq& \hat{c}_v \gamma \sqrt{\frac{d\log(12) + \theta\left( \sum_{i=1}^{l} \log\left( \frac{\sqrt{d_s} b_s (k_s - s_s + 1)}{4p} \right) + \log(b_{l+1}) + \log\left( \frac{n}{128p^2} \right) \right) + \log\left(\frac{4}{\varepsilon}\right)}{n}} \\
=& c_v \gamma \sqrt{\frac{2d + \theta\varrho + \log\left(\frac{4}{\varepsilon}\right)}{2n}}
\end{aligned}$$

holds with probability at least $1 - \varepsilon$, where $d = \tilde{r}_l \tilde{c}_l d_l d_{l+1} + \sum_{i=1}^{l} {k_i}^2 d_{i-1} d_i$, $\theta = a_{l+1}(d_{l+1} + \tilde{r}_l \tilde{c}_l d_l - 2a_{l+1} + 1) + \sum_{i=1}^{l} a_i ({k_i}^2 d_i + d_{i-1} - 2a_i + 1)$, $\varrho = \sum_{i=1}^{l} \log\left( \frac{\sqrt{d_i} b_i (k_i - s_i + 1)}{4p} \right) + \log(b_{l+1}) + \log\left( \frac{n}{128p^2} \right)$, and $\gamma = \left( \frac{\vartheta {b_{l+1}}^2 d_0^2}{{b_1}^4 d_1^2} l^2 r_0^2 c_0^2 \left[ \prod_{s=1}^{l} \frac{d_s {b_s}^2 (k_s - s_s + 1)^2}{8\sqrt{2}p^2} \right]^2 \right)^{1/2}$. The proof is completed. □

# D Proofs of Main Theorems

## D.1 Proof of Lemma 1

*Proof.* Recall that the weight of each kernel and the feature maps has magnitude bound separately, *i.e.* $\boldsymbol{w}_{(i)}^{k} \in \mathsf{B}^{{k_i}^2 d_{i-1}}(r_w)$ $(i = 1, \cdots, l; k = 1, \cdots, d_i)$ and $\boldsymbol{w}_{(l+1)} \in \mathsf{B}^{\tilde{r}_l \tilde{c}_l d_l d_{l+1}}(b_{l+1})$. Since $\widetilde{\boldsymbol{W}}_{(i)} = [\mathsf{vec}(\boldsymbol{W}_{(i)}^1), \mathsf{vec}(\boldsymbol{W}_{(i)}^2), \cdots, \mathsf{vec}(\boldsymbol{W}_{(i)}^{d_{i-1}})] \in \mathbb{R}^{k_i^2 d_i \times d_{i-1}}$, we have $\|\widetilde{\boldsymbol{W}}_{(i)}\|_F \leq d_i b_i$.

So here we assume $\widetilde{\boldsymbol{W}}_{(i,\epsilon)}$ is the $d_i b_i \epsilon/(b_{l+1}+\sum_{i=1}^{l} d_i b_i)$-covering net of the matrix $\widetilde{\boldsymbol{W}}_{(i)}$ which is the set of all parameters in the $i$-th layer. Then by Lemma 7, we have the covering number

$$n_{\epsilon}{}^{i} \leq \left(\frac{9(b_{l+1}+\sum_{i=1}^{l} d_i b_i)}{\epsilon}\right)^{a_i({k_i}^2 d_i + d_{i-1} - 2a_i + 1)},$$

since the rank of $\widetilde{\boldsymbol{W}}_{(i)}$ obeys $\mathsf{rank}(\widetilde{\boldsymbol{W}}_{(i)}) \leq a_i$ for $1 \leq i \leq l$. For the last layer, we also can construct an $b_{l+1}\epsilon/(b_{l+1} + \sum_{i=1}^{l} d_i b_i)$-covering net for the weight matrix $\boldsymbol{W}_{(l+1)}$. Here we have

$$n_{\epsilon}{}^{l+1} \leq \left(\frac{9(b_{l+1}+\sum_{i=1}^{l} d_i b_i)}{\epsilon}\right)^{a_{l+1}(d_{l+1}+\tilde{r}_l \tilde{c}_l d_l - 2a_{l+1}+1)},$$

since the rank of $\boldsymbol{W}_{(l+1)}$ obeys $\mathsf{rank}(\boldsymbol{W}_{(l+1)}) \leq a_{l+1}$. Finally, we arrange them together to construct a set $\Theta$ and claim that there is always an $\epsilon$-covering net $\boldsymbol{w}_\epsilon$ in $\Theta$ for any parameter $\boldsymbol{w}$. Accordingly, we have

$$|\Theta| \leq \prod_{i=1}^{l+1} n_{\epsilon}{}^{i} = \left(\frac{9(b_{l+1}+\sum_{i=1}^{l} d_i b_i)}{\epsilon}\right)^{a_{l+1}(d_{l+1}+\tilde{r}_l \tilde{c}_l d_l - 2a_{l+1}+1)+\sum_{i=1}^{l} a_i({k_i}^2 d_i + d_{i-1} - 2a_i + 1)} = \left(\frac{9(b_{l+1}+\sum_{i=1}^{l} d_i b_i)}{\epsilon}\right)^{\theta},$$

where $\theta = a_{l+1}(d_{l+1}+\tilde{r}_l \tilde{c}_l d_l - 2a_{l+1}+1)+\sum_{i=1}^{l} a_i({k_i}^2 d_i + d_{i-1} - 2a_i + 1)$ which is the total freedom degree of the network. So we can always find a vector $\boldsymbol{w}_{k_{\boldsymbol{w}}} \in \Theta$ such that $\|\boldsymbol{w} - \boldsymbol{w}_{k_{\boldsymbol{w}}}\|_2 \leq \epsilon$. Now we use the decomposition strategy to bound our goal:

$$\begin{aligned}
&\left|\widetilde{\boldsymbol{Q}}_n(\boldsymbol{w}) - \boldsymbol{Q}(\boldsymbol{w})\right| = \left|\frac{1}{n}\sum_{i=1}^{n} f(\boldsymbol{w}, \boldsymbol{D}^{(i)}) - \mathbb{E}_{\boldsymbol{D}\sim\boldsymbol{\mathcal{D}}}(f(\boldsymbol{w}, \boldsymbol{D}))\right| \\
=&\left|\frac{1}{n}\sum_{i=1}^{n}\Big(f(\boldsymbol{w}, \boldsymbol{D}^{(i)}) - f(\boldsymbol{w}_{k_{\boldsymbol{w}}}, \boldsymbol{D}^{(i)})\Big) + \frac{1}{n}\sum_{i=1}^{n} f(\boldsymbol{w}_{k_{\boldsymbol{w}}}, \boldsymbol{D}^{(i)}) - \mathbb{E} f(\boldsymbol{w}_{k_{\boldsymbol{w}}}, \boldsymbol{D}) + \mathbb{E}_{\boldsymbol{D}\sim\boldsymbol{\mathcal{D}}} f(\boldsymbol{w}_{k_{\boldsymbol{w}}}, \boldsymbol{D}) - \mathbb{E}_{\boldsymbol{D}\sim\boldsymbol{\mathcal{D}}} f(\boldsymbol{w}, \boldsymbol{D})\right| \\
\leq&\left|\frac{1}{n}\sum_{i=1}^{n}\Big(f(\boldsymbol{w}, \boldsymbol{D}^{(i)}) - f(\boldsymbol{w}_{k_{\boldsymbol{w}}}, \boldsymbol{D}^{(i)})\Big)\right| + \left|\frac{1}{n}\sum_{i=1}^{n} f(\boldsymbol{w}_{k_{\boldsymbol{w}}}, \boldsymbol{D}^{(i)}) - \mathbb{E}_{\boldsymbol{D}\sim\boldsymbol{\mathcal{D}}} f(\boldsymbol{w}_{k_{\boldsymbol{w}}}, \boldsymbol{D})\right| \\
&+ \left|\mathbb{E}_{\boldsymbol{D}\sim\boldsymbol{\mathcal{D}}} f(\boldsymbol{w}_{k_{\boldsymbol{w}}}, \boldsymbol{D}) - \mathbb{E}_{\boldsymbol{D}\sim\boldsymbol{\mathcal{D}}} f(\boldsymbol{w}, \boldsymbol{D})\right|.
\end{aligned}$$

Then, we define four events $\boldsymbol{E}_0$, $\boldsymbol{E}_1$, $\boldsymbol{E}_2$ and $\boldsymbol{E}_3$ as

$$\begin{aligned}
\boldsymbol{E}_0 &= \left\{\sup_{\boldsymbol{w}\in\Omega} \left|\widetilde{\boldsymbol{Q}}_n(\boldsymbol{w}) - \boldsymbol{Q}(\boldsymbol{w})\right| \geq t\right\}, \\
\boldsymbol{E}_1 &= \left\{\sup_{\boldsymbol{w}\in\Omega} \left|\frac{1}{n}\sum_{i=1}^{n}\Big(f(\boldsymbol{w}, \boldsymbol{D}^{(i)}) - f(\boldsymbol{w}_{k_{\boldsymbol{w}}}, \boldsymbol{x}_{(i)})\Big)\right| \geq \frac{t}{3}\right\}, \\
\boldsymbol{E}_2 &= \left\{\sup_{\boldsymbol{w}_{k_{\boldsymbol{w}}}\in\Theta} \left|\frac{1}{n}\sum_{i=1}^{n} f(\boldsymbol{w}_{k_{\boldsymbol{w}}}, \boldsymbol{D}^{(i)}) - \mathbb{E}_{\boldsymbol{D}\sim\boldsymbol{\mathcal{D}}}(f(\boldsymbol{w}_{k_{\boldsymbol{w}}}, \boldsymbol{D}))\right| \geq \frac{t}{3}\right\}, \\
\boldsymbol{E}_3 &= \left\{\sup_{\boldsymbol{w}\in\Omega} \left|\mathbb{E}_{\boldsymbol{D}\sim\boldsymbol{\mathcal{D}}}(f(\boldsymbol{w}_{k_{\boldsymbol{w}}}, \boldsymbol{D})) - \mathbb{E}_{\boldsymbol{D}\sim\boldsymbol{\mathcal{D}}}(f(\boldsymbol{w}, \boldsymbol{D}))\right| \geq \frac{t}{3}\right\}.
\end{aligned}$$

Accordingly, we have

$$\mathbb{P}(\boldsymbol{E}_0) \leq \mathbb{P}(\boldsymbol{E}_1) + \mathbb{P}(\boldsymbol{E}_2) + \mathbb{P}(\boldsymbol{E}_3).$$

So we can respectively bound $\mathbb{P}(\boldsymbol{E}_1)$, $\mathbb{P}(\boldsymbol{E}_2)$ and $\mathbb{P}(\boldsymbol{E}_3)$ to bound $\mathbb{P}(\boldsymbol{E}_0)$.

**Step 1. Bound $\mathbb{P}(\boldsymbol{E}_1)$**: We first bound $\mathbb{P}(\boldsymbol{E}_1)$ as follows:

$$\begin{aligned}\mathbb{P}(\boldsymbol{E}_1) =&\mathbb{P}\left(\sup_{\boldsymbol{w}\in\Omega}\left|\frac{1}{n}\sum_{i=1}^{n}\left(f(\boldsymbol{w},\boldsymbol{D}^{(i)})-f(\boldsymbol{w}_{k_{\boldsymbol{w}}},\boldsymbol{D}^{(i)})\right)\right|\geq\frac{t}{3}\right)\\ \overset{①}{\leq}&\frac{3}{t}\mathbb{E}_{\boldsymbol{D}\sim\boldsymbol{\mathcal{D}}}\left(\sup_{\boldsymbol{w}\in\Omega}\left|\frac{1}{n}\sum_{i=1}^{n}\left(f(\boldsymbol{w},\boldsymbol{D}^{(i)})-f(\boldsymbol{w}_{k_{\boldsymbol{w}}},\boldsymbol{D}^{(i)})\right)\right|\right)\\ \leq&\frac{3}{t}\mathbb{E}_{\boldsymbol{D}\sim\boldsymbol{\mathcal{D}}}\left(\sup_{\boldsymbol{w}\in\Omega}\frac{\left|\frac{1}{n}\sum_{i=1}^{n}\left(f(\boldsymbol{w},\boldsymbol{D}^{(i)})-f(\boldsymbol{w}_{k_{\boldsymbol{w}}},\boldsymbol{D}^{(i)})\right)\right|}{\|\boldsymbol{w}-\boldsymbol{w}_{k_{\boldsymbol{w}}}\|_2}\sup_{\boldsymbol{w}\in\Omega}\|\boldsymbol{w}-\boldsymbol{w}_{k_{\boldsymbol{w}}}\|_2\right)\\ \leq&\frac{3\epsilon}{t}\mathbb{E}_{\boldsymbol{D}\sim\boldsymbol{\mathcal{D}}}\left(\sup_{\boldsymbol{w}\in\Omega}\left\|\nabla\widetilde{\boldsymbol{Q}}_n(\boldsymbol{w},\boldsymbol{D})\right\|_2\right),\end{aligned}$$

where ① holds since by Markov inequality, we have that for an arbitrary nonnegative random variable $x$, then

$$\mathbb{P}(x\geq t)\leq\frac{\mathbb{E}(x)}{t}.$$

Now we only need to bound $\mathbb{E}_{\boldsymbol{D}\sim\boldsymbol{\mathcal{D}}}\left(\sup_{\boldsymbol{w}\in\Omega}\left\|\nabla\widetilde{\boldsymbol{Q}}_n(\boldsymbol{w},\boldsymbol{D})\right\|_2\right)$. Therefore, by Lemma 11, we have

$$\mathbb{E}_{\boldsymbol{D}\sim\boldsymbol{\mathcal{D}}}\left(\sup_{\boldsymbol{w}\in\Omega}\left\|\nabla\widetilde{\boldsymbol{Q}}_n(\boldsymbol{w},\boldsymbol{D})\right\|_2\right)=\mathbb{E}_{\boldsymbol{D}\sim\boldsymbol{\mathcal{D}}}\left(\sup_{\boldsymbol{w}\in\Omega}\left\|\frac{1}{n}\sum_{i=1}^{n}\nabla f(\boldsymbol{w},\boldsymbol{D}^{(i)})\right\|_2\right)\leq\mathbb{E}_{\boldsymbol{D}\sim\boldsymbol{\mathcal{D}}}\left(\sup_{\boldsymbol{w}\in\Omega}\|\nabla f(\boldsymbol{w},\boldsymbol{D})\|_2\right)\leq\beta.$$

where $\beta\triangleq\left[\vartheta\tilde{r}_l\tilde{c}_l d_l+\sum_{i=1}^{l}\frac{\vartheta {b_{l+1}}^2 d_{i-1}}{p^2 {b_i}^2 d_i}r_{i-1}c_{i-1}\prod_{s=i}^{l}\frac{d_s {b_s}^2(k_s-s_s+1)^2}{16p^2}\right]^{1/2}$ in which $\vartheta=1/8$. Therefore, we have

$$\mathbb{P}(\boldsymbol{E}_1)\leq\frac{3\epsilon\beta}{t}.$$

We further let

$$t\geq\frac{6\epsilon\beta}{\varepsilon}.$$

Then we can bound $\mathbb{P}(\boldsymbol{E}_1)$:

$$\mathbb{P}(\boldsymbol{E}_1)\leq\frac{\varepsilon}{2}.$$

**Step 2. Bound $\mathbb{P}(\boldsymbol{E}_2)$**: Recall that we use $\Theta$ to denote the index of $\boldsymbol{w}_{k_{\boldsymbol{w}}}$ and we have $|\Theta|\leq\prod_{i=1}^{l+1}{n_\epsilon}^i=\left(\frac{9(b_{l+1}+\sum_{i=1}^{l}d_i b_i)}{\epsilon}\right)^\theta$. We can bound $\mathbb{P}(\boldsymbol{E}_2)$ as follows:

$$\begin{aligned}\mathbb{P}(\boldsymbol{E}_2)=&\mathbb{P}\left(\sup_{\boldsymbol{w}_{k_{\boldsymbol{w}}}\in\Theta}\left|\frac{1}{n}\sum_{i=1}^{n}f(\boldsymbol{w}_{k_{\boldsymbol{w}}},\boldsymbol{D}^{(i)})-\mathbb{E}_{\boldsymbol{D}\sim\boldsymbol{\mathcal{D}}}(f(\boldsymbol{w}_{k_{\boldsymbol{w}}},\boldsymbol{D}))\right|\geq\frac{t}{3}\right)\\ \leq&\left(\frac{9(b_{l+1}+\sum_{i=1}^{l}d_i b_i)}{\epsilon}\right)^\theta\sup_{\boldsymbol{w}_{k_{\boldsymbol{w}}}\in\Theta}\mathbb{P}\left(\left|\frac{1}{n}\sum_{i=1}^{n}f(\boldsymbol{w}_{k_{\boldsymbol{w}}},\boldsymbol{D}^{(i)})-\mathbb{E}_{\boldsymbol{D}\sim\boldsymbol{\mathcal{D}}}(f(\boldsymbol{w}_{k_{\boldsymbol{w}}},\boldsymbol{D}))\right|\geq\frac{t}{3}\right)\\ \overset{①}{\leq}&\left(\frac{9(b_{l+1}+\sum_{i=1}^{l}d_i b_i)}{\epsilon}\right)^\theta 2\exp\left(-\frac{2nt^2}{\alpha^2}\right),\end{aligned}$$

where ① holds because in Lemma 15, we have

$$\mathbb{P}\left(\frac{1}{n}\sum_{i=1}^{n}\left(f(\boldsymbol{w},\boldsymbol{D}^{(i)})-\mathbb{E}(f(\boldsymbol{w},\boldsymbol{D}^{(i)}))\right)>t\right)\leq\exp\left(-\frac{2nt^2}{\alpha^2}\right),$$

where $\alpha = 1$. Thus, if we set

$$t \geq \sqrt{\frac{\alpha^2\left(\theta \log\left(\frac{9(b_{l+1}+\sum_{i=1}^{l} d_i b_i)}{\epsilon}\right) + \log\left(\frac{4}{\varepsilon}\right)\right)}{2n}},$$

then we have

$$\mathbb{P}\left(\boldsymbol{E}_2\right) \leq \frac{\varepsilon}{2}.$$

**Step 3. Bound** $\mathbb{P}\left(\boldsymbol{E}_3\right)$: We first bound $\mathbb{P}\left(\boldsymbol{E}_3\right)$ as follows:

$$\begin{aligned}
\mathbb{P}\left(\boldsymbol{E}_3\right) =& \mathbb{P}\left(\sup_{\boldsymbol{w}\in\Omega} \left\|\mathbb{E}_{\boldsymbol{D}\sim\boldsymbol{\mathcal{D}}}(f(\boldsymbol{w}_{k_{\boldsymbol{w}}}, \boldsymbol{D})) - \mathbb{E}_{\boldsymbol{D}\sim\boldsymbol{\mathcal{D}}}(f(\boldsymbol{w}, \boldsymbol{D}))\right\|_2 \geq \frac{t}{3}\right)\\
=& \mathbb{P}\left(\sup_{\boldsymbol{w}\in\Omega} \frac{\left\|\mathbb{E}_{\boldsymbol{D}\sim\boldsymbol{\mathcal{D}}}\left(f(\boldsymbol{w}_{k_{\boldsymbol{w}}}, \boldsymbol{D}) - f(\boldsymbol{w}, \boldsymbol{D})\right\|_2\right)}{\|\boldsymbol{w} - \boldsymbol{w}_{k_{\boldsymbol{w}}}\|_2} \sup_{\boldsymbol{w}\in\Omega} \|\boldsymbol{w} - \boldsymbol{w}_{k_{\boldsymbol{w}}}\|_2 \geq \frac{t}{3}\right)\\
\leq& \mathbb{P}\left(\epsilon \mathbb{E}_{\boldsymbol{D}\sim\boldsymbol{\mathcal{D}}} \sup_{\boldsymbol{w}\in\Omega} \|\nabla \boldsymbol{Q}_{\boldsymbol{w}}(\boldsymbol{w}, \boldsymbol{D})\|_2 \geq \frac{t}{3}\right)\\
\overset{①}{\leq}& \mathbb{P}\left(\beta\epsilon \geq \frac{t}{3}\right),
\end{aligned}$$

where ① holds since we utilize Lemma 11. We set $\epsilon$ enough small such that $\beta\epsilon < t/3$ always holds. Then it yields $\mathbb{P}\left(\boldsymbol{E}_3\right) = 0$.

**Step 4. Final result**: To ensure $\mathbb{P}(\boldsymbol{E}_0) \leq \varepsilon$, we just set $\epsilon = \frac{18p^2(b_{l+1}+\sum_{i=1}^{l} d_i b_i)}{\vartheta^2 n b_{l+1}} \left[\prod_{s=1}^{l} \frac{d_s {b_s}^2 (k_s - s_s + 1)^2}{16p^2}\right]^{-\frac{1}{2}}$. Note that $\frac{6\epsilon\beta}{\varepsilon} > 3\epsilon\beta$ due to $\varepsilon \leq 1$. Thus we can obtain

$$t \geq \max\left(\frac{6\epsilon\beta}{\varepsilon}, \sqrt{\frac{\alpha^2\left(\theta \log\left(\frac{9(b_{l+1}+\sum_{i=1}^{l} d_i b_i)}{\epsilon}\right) + \log\left(\frac{4}{\varepsilon}\right)\right)}{2n}}\right).$$

By comparing the values of $\alpha$, we can observe that if $n \geq c_{f'} \frac{l^2(b_{l+1}+\sum_{i=1}^{l} d_i b_i)^2 \max_i \sqrt{r_i c_i}}{\theta \varrho \varepsilon^2}$ where $c_{f'}$ is a constant, there exists such a universal constant $c_f$ such that

$$\sup_{\boldsymbol{w}\in\Omega} \left|\widetilde{\boldsymbol{Q}}_n(\boldsymbol{w}) - \boldsymbol{Q}(\boldsymbol{w})\right| \leq \alpha \sqrt{\frac{\theta\left(\sum_{i=1}^{l} \log\left(\frac{\sqrt{d_s} b_s (k_s - s_s + 1)}{4p}\right) + \log(b_{l+1}) + \log\left(\frac{n}{128p^2}\right)\right) + \log\left(\frac{4}{\varepsilon}\right)}{2n}} = \sqrt{\frac{\theta\varrho + \log\left(\frac{4}{\varepsilon}\right)}{2n}}$$

holds with probability at least $1 - \varepsilon$, where $\theta = a_{l+1}(d_{l+1} + \tilde{r}_l \tilde{c}_l d_l - 2a_{l+1} + 1) + \sum_{i=1}^{l} a_i({k_i}^2 d_i + d_{i-1} - 2a_i + 1)$, $\varrho = \sum_{i=1}^{l} \log\left(\frac{\sqrt{d_i} b_i (k_i - s_i + 1)}{4p}\right) + \log(b_{l+1}) + \log\left(\frac{n}{128p^2}\right)$, and $\alpha = 1$. The proof is completed. □

### D.2 Proof of Theorem 1

*Proof.* By Lemma 1 in the manuscript, we know that if $n \geq c_{f'} l^2(b_{l+1} + \sum_{i=1}^{l} d_i b_i)^2 \max_i \sqrt{r_i c_i}/(\theta\varrho\varepsilon^2)$ where $c_{f'}$ is a universal constant, then with probability at least $1 - \varepsilon$, we have

$$\sup_{\boldsymbol{w}\in\Omega} \left|\widetilde{\boldsymbol{Q}}_n(\boldsymbol{w}) - \boldsymbol{Q}(\boldsymbol{w})\right| \leq \sqrt{\frac{\theta\varrho + \log\left(\frac{4}{\varepsilon}\right)}{2n}},$$

where the total freedom degree $\theta$ of the network is $\theta = a_{l+1}(d_{l+1} + \tilde{r}_l \tilde{c}_l d_l + 1) + \sum_{i=1}^{l} a_i({k_i}^2 d_{i-1} + d_i + 1)$ and $\varrho = \sum_{i=1}^{l} \log\left(\frac{\sqrt{d_i} b_i (k_i - s_i + 1)}{4p}\right) + \log(b_{l+1}) + \log\left(\frac{n}{128p^2}\right)$.

Thus based on such a result, we can derive the following generalization bound:

$$\mathbb{E}_{\boldsymbol{\mathcal{S}}\sim\boldsymbol{\mathcal{D}}}\left|\mathbb{E}_{\mathcal{A}}(\boldsymbol{Q}(\widetilde{\boldsymbol{w}})-\widetilde{\boldsymbol{Q}}_n(\widetilde{\boldsymbol{w}}))\right| \leq \mathbb{E}_{\boldsymbol{\mathcal{S}}\sim\boldsymbol{\mathcal{D}}}\left(\sup_{\boldsymbol{w}\in\Omega}\left|\widetilde{\boldsymbol{Q}}_n(\boldsymbol{w})-\boldsymbol{Q}(\boldsymbol{w})\right|\right) \leq \sup_{\boldsymbol{w}\in\Omega}\left|\widetilde{\boldsymbol{Q}}_n(\boldsymbol{w})-\boldsymbol{Q}(\boldsymbol{w})\right| \leq \sqrt{\frac{\theta\varrho+\log\left(\frac{4}{\varepsilon}\right)}{2n}}.$$

Thus, the conclusion holds. The proof is completed. □

### D.3 Proof of Theorem 2

*Proof.* Recall that the weight of each kernel and the feature maps has magnitude bound separately, *i.e.* $\boldsymbol{w}_{(i)}^k \in \mathsf{B}^{k_i{}^2 d_{i-1}}(r_w)$ $(i=1,\cdots,l; k=1,\cdots,d_i)$ and $\boldsymbol{w}_{(l+1)} \in \mathsf{B}^{\tilde{r}_l\tilde{c}_l d_l d_{l+1}}(b_{l+1})$. Since $\widetilde{\boldsymbol{W}}_{(i)} = [\mathsf{vec}(\boldsymbol{W}_{(i)}^1), \mathsf{vec}(\boldsymbol{W}_{(i)}^2), \cdots, \mathsf{vec}(\boldsymbol{W}_{(i)}^{d_{i-1}})] \in \mathbb{R}^{k_i^2 d_{i-1}\times d_i}$, we have $\|\widetilde{\boldsymbol{W}}_{(i)}\|_F \leq d_i b_i$.

So here we assume $\widetilde{\boldsymbol{W}}_{(i,\epsilon)}$ is the $d_i b_i\epsilon/(b_{l+1}+\sum_{i=1}^l d_i b_i)$-covering net of the matrix $\widetilde{\boldsymbol{W}}_{(i)}$ which is the set of all parameters in the $i$-th layer. Then by Lemma 7, we have the covering number

$$n_\epsilon{}^i \leq \left(\frac{9(b_{l+1}+\sum_{i=1}^l d_i b_i)}{\epsilon}\right)^{a_i(k_i{}^2 d_{i-1}+d_i-2a_i+1)},$$

since the rank of $\widetilde{\boldsymbol{W}}_{(i)}$ obeys $\mathsf{rank}(\widetilde{\boldsymbol{W}}_{(i)}) \leq a_i$ for $1\leq i\leq l$. For the last layer, we also can construct an $b_{l+1}\epsilon/(b_{l+1}+\sum_{i=1}^l d_i b_i)$-covering net for the weight matrix $\boldsymbol{W}_{(l+1)}$. Here we have

$$n_\epsilon{}^{l+1} \leq \left(\frac{9(b_{l+1}+\sum_{i=1}^l d_i b_i)}{\epsilon}\right)^{a_{l+1}(d_{l+1}+\tilde{r}_l\tilde{c}_l d_l-2a_{l+1}+1)},$$

since the rank of $\boldsymbol{W}_{(l+1)}$ obeys $\mathsf{rank}(\boldsymbol{W}_{(l+1)}) \leq a_{l+1}$. Finally, we arrange them together to construct a set $\Theta$ and claim that there is always an $\epsilon$-covering net $\boldsymbol{w}_\epsilon$ in $\Theta$ for any parameter $\boldsymbol{w}$. Accordingly, we have

$$|\Theta| \leq \prod_{i=1}^{l+1} n_\epsilon{}^i = \left(\frac{9(b_{l+1}+\sum_{i=1}^l d_i b_i)}{\epsilon}\right)^{a_{l+1}(d_{l+1}+\tilde{r}_l\tilde{c}_l d_l-2a_{l+1}+1)+\sum_{i=1}^l a_i(k_i{}^2 d_{i-1}+d_i-2a_i+1)} = \left(\frac{9(b_{l+1}+\sum_{i=1}^l d_i b_i)}{\epsilon}\right)^{\theta},$$

where $\theta = a_{l+1}(d_{l+1}+\tilde{r}_l\tilde{c}_l d_l - 2a_{l+1}+1)+\sum_{i=1}^l a_i(k_i{}^2 d_{i-1}+d_i-2a_i+1)$ which is the total freedom degree of the network. So we can always find a vector $\boldsymbol{w}_{k_{\boldsymbol{w}}} \in \Theta$ such that $\|\boldsymbol{w}-\boldsymbol{w}_{k_{\boldsymbol{w}}}\|_2 \leq \epsilon$. Accordingly, we can decompose $\left\|\nabla\widetilde{\boldsymbol{Q}}_n(\boldsymbol{w})-\nabla\boldsymbol{Q}(\boldsymbol{w})\right\|_2$ as

$$\begin{aligned}
&\left\|\nabla\widetilde{\boldsymbol{Q}}_n(\boldsymbol{w})-\nabla\boldsymbol{Q}(\boldsymbol{w})\right\|_2\\
=&\left\|\frac{1}{n}\sum_{i=1}^n \nabla f(\boldsymbol{w},\boldsymbol{D}^{(i)})-\mathbb{E}_{\boldsymbol{D}\sim\boldsymbol{\mathcal{D}}}(\nabla f(\boldsymbol{w},\boldsymbol{D}))\right\|_2\\
=&\Bigg\|\frac{1}{n}\sum_{i=1}^n\left(\nabla f(\boldsymbol{w},\boldsymbol{D}^{(i)})-\nabla f(\boldsymbol{w}_{k_{\boldsymbol{w}}},\boldsymbol{D}^{(i)})\right)+\frac{1}{n}\sum_{i=1}^n\nabla f(\boldsymbol{w}_{k_{\boldsymbol{w}}},\boldsymbol{D}^{(i)})-\mathbb{E}_{\boldsymbol{D}\sim\boldsymbol{\mathcal{D}}}(\nabla f(\boldsymbol{w}_{k_{\boldsymbol{w}}},\boldsymbol{D}))\\
&\quad+\mathbb{E}_{\boldsymbol{D}\sim\boldsymbol{\mathcal{D}}}(\nabla f(\boldsymbol{w}_{k_{\boldsymbol{w}}},\boldsymbol{D}))-\mathbb{E}_{\boldsymbol{D}\sim\boldsymbol{\mathcal{D}}}(\nabla f(\boldsymbol{w},\boldsymbol{D}))\Bigg\|_2\\
\leq&\left\|\frac{1}{n}\sum_{i=1}^n\left(\nabla f(\boldsymbol{w},\boldsymbol{D}^{(i)})-\nabla f(\boldsymbol{w}_{k_{\boldsymbol{w}}},\boldsymbol{D}^{(i)})\right)\right\|_2+\left\|\frac{1}{n}\sum_{i=1}^n\nabla f(\boldsymbol{w}_{k_{\boldsymbol{w}}},\boldsymbol{D}^{(i)})-\mathbb{E}_{\boldsymbol{D}\sim\boldsymbol{\mathcal{D}}}(\nabla f(\boldsymbol{w}_{k_{\boldsymbol{w}}},\boldsymbol{D}))\right\|_2\\
&\quad+\left\|\mathbb{E}_{\boldsymbol{D}\sim\boldsymbol{\mathcal{D}}}(\nabla f(\boldsymbol{w}_{k_{\boldsymbol{w}}},\boldsymbol{D}))-\mathbb{E}_{\boldsymbol{D}\sim\boldsymbol{\mathcal{D}}}(\nabla f(\boldsymbol{w},\boldsymbol{D}))\right\|_2.
\end{aligned}$$

Here we also define four events $\boldsymbol{E}_0$, $\boldsymbol{E}_1$, $\boldsymbol{E}_2$ and $\boldsymbol{E}_3$ as

$$\begin{aligned}
\boldsymbol{E}_0 &= \left\{ \sup_{\boldsymbol{w}\in\Omega} \left\| \nabla \widetilde{\boldsymbol{Q}}_n(\boldsymbol{w}) - \nabla \boldsymbol{Q}(\boldsymbol{w}) \right\|_2 \geq t \right\}, \\
\boldsymbol{E}_1 &= \left\{ \sup_{\boldsymbol{w}\in\Omega} \left\| \frac{1}{n}\sum_{i=1}^{n} \left( \nabla f(\boldsymbol{w}, \boldsymbol{D}^{(i)}) - \nabla f(\boldsymbol{w}_{k_{\boldsymbol{w}}}, \boldsymbol{D}^{(i)}) \right) \right\|_2 \geq \frac{t}{3} \right\}, \\
\boldsymbol{E}_2 &= \left\{ \sup_{\boldsymbol{w}_{k_{\boldsymbol{w}}}\in\Theta} \left\| \frac{1}{n}\sum_{i=1}^{n} \nabla f(\boldsymbol{w}_{k_{\boldsymbol{w}}}, \boldsymbol{D}^{(i)}) - \mathbb{E}_{\boldsymbol{D}\sim\boldsymbol{\mathcal{D}}}(\nabla f(\boldsymbol{w}_{k_{\boldsymbol{w}}}, \boldsymbol{D})) \right\|_2 \geq \frac{t}{3} \right\}, \\
\boldsymbol{E}_3 &= \left\{ \sup_{\boldsymbol{w}\in\Omega} \left\| \mathbb{E}_{\boldsymbol{D}\sim\boldsymbol{\mathcal{D}}}(\nabla f(\boldsymbol{w}_{k_{\boldsymbol{w}}}, \boldsymbol{D})) - \mathbb{E}_{\boldsymbol{D}\sim\boldsymbol{\mathcal{D}}}(\nabla f(\boldsymbol{w}, \boldsymbol{D})) \right\|_2 \geq \frac{t}{3} \right\}.
\end{aligned}$$

Accordingly, we have

$$\mathbb{P}\left(\boldsymbol{E}_0\right) \leq \mathbb{P}\left(\boldsymbol{E}_1\right) + \mathbb{P}\left(\boldsymbol{E}_2\right) + \mathbb{P}\left(\boldsymbol{E}_3\right).$$

So we can respectively bound $\mathbb{P}\left(\boldsymbol{E}_1\right)$, $\mathbb{P}\left(\boldsymbol{E}_2\right)$ and $\mathbb{P}\left(\boldsymbol{E}_3\right)$ to bound $\mathbb{P}\left(\boldsymbol{E}_0\right)$.

**Step 1. Bound $\mathbb{P}\left(\boldsymbol{E}_1\right)$:** We first bound $\mathbb{P}\left(\boldsymbol{E}_1\right)$ as follows:

$$\begin{aligned}
\mathbb{P}\left(\boldsymbol{E}_1\right) =& \mathbb{P}\left( \sup_{\boldsymbol{w}\in\Omega} \left\| \frac{1}{n}\sum_{i=1}^{n} \left( \nabla f(\boldsymbol{w}, \boldsymbol{D}^{(i)}) - \nabla f(\boldsymbol{w}_{k_{\boldsymbol{w}}}, \boldsymbol{D}^{(i)}) \right) \right\|_2 \geq \frac{t}{3} \right) \\
\overset{①}{\leq}& \frac{3}{t}\mathbb{E}_{\boldsymbol{D}\sim\boldsymbol{\mathcal{D}}}\left( \sup_{\boldsymbol{w}\in\Omega} \left\| \frac{1}{n}\sum_{i=1}^{n} \left( \nabla f(\boldsymbol{w}, \boldsymbol{D}^{(i)}) - \nabla f(\boldsymbol{w}_{k_{\boldsymbol{w}}}, \boldsymbol{D}^{(i)}) \right) \right\|_2 \right) \\
\leq& \frac{3}{t}\mathbb{E}_{\boldsymbol{D}\sim\boldsymbol{\mathcal{D}}}\left( \sup_{\boldsymbol{w}\in\Omega} \frac{\left\| \frac{1}{n}\sum_{i=1}^{n} \left( \nabla f(\boldsymbol{w}, \boldsymbol{D}^{(i)}) - \nabla f(\boldsymbol{w}_{k_{\boldsymbol{w}}}, \boldsymbol{D}^{(i)}) \right) \right\|_2}{\|\boldsymbol{w}-\boldsymbol{w}_{k_{\boldsymbol{w}}}\|_2} \sup_{\boldsymbol{w}\in\Omega} \|\boldsymbol{w}-\boldsymbol{w}_{k_{\boldsymbol{w}}}\|_2 \right) \\
\leq& \frac{3\epsilon}{t}\mathbb{E}_{\boldsymbol{D}\sim\boldsymbol{\mathcal{D}}}\left( \sup_{\boldsymbol{w}\in\Omega} \left\| \nabla^2 \widetilde{\boldsymbol{Q}}_n(\boldsymbol{w}, \boldsymbol{D}) \right\|_2 \right),
\end{aligned}$$

where ① holds since by Markov inequality, we have that for an arbitrary nonnegative random variable $x$, then $\mathbb{P}(x \geq t) \leq \frac{\mathbb{E}(x)}{t}$.

Now we only need to bound $\mathbb{E}_{\boldsymbol{D}\sim\boldsymbol{\mathcal{D}}}\left( \sup_{\boldsymbol{w}\in\Omega} \left\| \nabla^2 \widetilde{\boldsymbol{Q}}_n(\boldsymbol{w}, \boldsymbol{D}) \right\|_2 \right)$. Here we utilize Lemma 14 to achieve this goal:

$$\mathbb{E}_{\boldsymbol{D}\sim\boldsymbol{\mathcal{D}}}\left( \sup_{\boldsymbol{w}\in\Omega} \left\| \nabla^2 \widetilde{\boldsymbol{Q}}_n(\boldsymbol{w}, \boldsymbol{D}) \right\|_2 \right) \leq \mathbb{E}_{\boldsymbol{D}\sim\boldsymbol{\mathcal{D}}}\left( \sup_{\boldsymbol{w}\in\Omega} \left\| \nabla^2 f(\boldsymbol{w}, \boldsymbol{D}) - \nabla^2 f(\boldsymbol{w}^*, \boldsymbol{D}) \right\|_2 \right) \leq \gamma.$$

where $\gamma = \left( \frac{\vartheta {b_{l+1}}^2 d_0^2}{{b_1}^4 d_1^2} l^2 r_0^2 c_0^2 \left[ \prod_{s=1}^{l} \frac{d_s {b_s}^2 (k_s - s_s + 1)^2}{8\sqrt{2} p^2} \right]^2 \right)^{1/2}$. Therefore, we have

$$\mathbb{P}\left(\boldsymbol{E}_1\right) \leq \frac{3\gamma\epsilon}{t}.$$

We further let

$$t \geq \frac{6\gamma\epsilon}{\varepsilon}.$$

Then we can bound $\mathbb{P}(\boldsymbol{E}_1)$:

$$\mathbb{P}(\boldsymbol{E}_1) \leq \frac{\varepsilon}{2}.$$

**Step 2. Bound $\mathbb{P}\left(\boldsymbol{E}_2\right)$:** By Lemma 2, we know that for any vector $\boldsymbol{x} \in \mathbb{R}^d$, its $\ell_2$-norm can be computed as

$$\|\boldsymbol{x}\|_2 \leq \frac{1}{1-\epsilon} \sup_{\boldsymbol{\lambda}\in\boldsymbol{\lambda}_\epsilon} \left\langle \boldsymbol{\lambda}, \boldsymbol{x} \right\rangle.$$

where $\boldsymbol{\lambda}_\epsilon = \{\boldsymbol{\lambda}_1, \ldots, \boldsymbol{\lambda}_{k_w}\}$ be an $\epsilon$-covering net of $\mathsf{B}^d(1)$.

Let $\boldsymbol{\lambda}$ be the $\frac{1}{2}$-covering net of $\mathsf{B}^d(1)$, where $d = \tilde{r}_l \tilde{c}_l d_l d_{l+1} + \sum_{i=1}^l {k_i}^2 d_{i-1} d_i$. Recall that we use $\Theta$ to denote the index of $\boldsymbol{w}_{k_{\boldsymbol{w}}}$ so that $\|\boldsymbol{w} - \boldsymbol{w}_{k_{\boldsymbol{w}}}\| \leq \epsilon$. Besides, $|\Theta| \leq \prod_{i=1}^{l+1} {n_\epsilon}^i = \left(\frac{3(b_{l+1} + \sum_{i=1}^l d_i b_i)}{\epsilon}\right)^\theta$. Then we can bound $\mathbb{P}(\boldsymbol{E}_2)$ as follows:

$$
\begin{aligned}
\mathbb{P}(\boldsymbol{E}_2) =& \mathbb{P}\left(\sup_{\boldsymbol{w}_{k_{\boldsymbol{w}}} \in \Theta} \left\| \frac{1}{n} \sum_{i=1}^n \nabla f(\boldsymbol{w}_{k_{\boldsymbol{w}}}, \boldsymbol{D}^{(i)}) - \mathbb{E}_{\boldsymbol{D} \sim \boldsymbol{\mathcal{D}}}(\nabla f(\boldsymbol{w}_{k_{\boldsymbol{w}}}, \boldsymbol{D})) \right\|_2 \geq \frac{t}{3}\right) \\
=& \mathbb{P}\left(\sup_{\boldsymbol{w}_{k_{\boldsymbol{w}}} \in \Theta, \boldsymbol{\lambda} \in \boldsymbol{\lambda}_{1/2}} 2 \left\langle \boldsymbol{\lambda}, \frac{1}{n} \sum_{i=1}^n \nabla f(\boldsymbol{w}_{k_{\boldsymbol{w}}}, \boldsymbol{D}^{(i)}) - \mathbb{E}_{\boldsymbol{D} \sim \boldsymbol{\mathcal{D}}} \left(\nabla f(\boldsymbol{w}_{k_{\boldsymbol{w}}}, \boldsymbol{D})\right) \right\rangle \geq \frac{t}{3}\right) \\
\leq& 6^d \left(\frac{9(b_{l+1} + \sum_{i=1}^l d_i b_i)}{\epsilon}\right)^\theta \sup_{\boldsymbol{w}_{k_{\boldsymbol{w}}} \in \Theta, \boldsymbol{\lambda} \in \boldsymbol{\lambda}_{1/2}} \mathbb{P}\left(\frac{1}{n} \sum_{i=1}^n \left\langle \boldsymbol{\lambda}, \nabla f(\boldsymbol{w}_{k_{\boldsymbol{w}}}, \boldsymbol{D}^{(i)}) - \mathbb{E}_{\boldsymbol{D} \sim \boldsymbol{\mathcal{D}}} \left(\nabla f(\boldsymbol{w}_{k_{\boldsymbol{w}}}, \boldsymbol{D})\right) \right\rangle \geq \frac{t}{6}\right) \\
\overset{①}{\leq}& 6^d \left(\frac{9(b_{l+1} + \sum_{i=1}^l d_i b_i)}{\epsilon}\right)^\theta 2 \exp\left(-\frac{nt^2}{72\beta^2}\right),
\end{aligned}
$$

where ① holds since by Lemma 16, we have

$$
\mathbb{P}\left(\frac{1}{n} \sum_{i=1}^n \left(\left\langle \boldsymbol{\lambda}, \nabla_{\boldsymbol{w}} f(\boldsymbol{w}, \boldsymbol{D}^{(i)}) - \mathbb{E}_{\boldsymbol{D} \sim \boldsymbol{\mathcal{D}}} \nabla_{\boldsymbol{w}} f(\boldsymbol{w}, \boldsymbol{D}^{(i)}) \right\rangle\right) > t\right) \leq \exp\left(-\frac{nt^2}{2\beta^2}\right).
$$

where $\beta \triangleq \left[\vartheta \tilde{r}_l \tilde{c}_l d_l + \sum_{i=1}^l \frac{\vartheta {b_{l+1}}^2 d_{i-1}}{p^2 {b_i}^2 d_i} r_{i-1} c_{i-1} \prod_{s=i}^l \frac{d_s {b_s}^2 (k_s - s_s + 1)^2}{16 p^2}\right]^{1/2}$ in which $\vartheta = 1/8$.

Thus, if we set

$$
t \geq \sqrt{\frac{72\beta^2 \left(d \log(6) + \theta \log\left(\frac{9(b_{l+1} + \sum_{i=1}^l d_i b_i)}{\epsilon}\right) + \log\left(\frac{4}{\varepsilon}\right)\right)}{n}},
$$

then we have

$$
\mathbb{P}(\boldsymbol{E}_2) \leq \frac{\varepsilon}{2}.
$$

**Step 3. Bound $\mathbb{P}(\boldsymbol{E}_3)$**: We first bound $\mathbb{P}(\boldsymbol{E}_3)$ as follows:

$$
\begin{aligned}
\mathbb{P}(\boldsymbol{E}_3) =& \mathbb{P}\left(\sup_{\boldsymbol{w} \in \Omega} \|\mathbb{E}(\nabla f(\boldsymbol{w}_{k_{\boldsymbol{w}}}, \boldsymbol{x})) - \mathbb{E}_{\boldsymbol{D} \sim \boldsymbol{\mathcal{D}}}(\nabla f(\boldsymbol{w}, \boldsymbol{x}))\|_2 \geq \frac{t}{3}\right) \\
=& \mathbb{P}\left(\sup_{\boldsymbol{w} \in \Omega} \frac{\|\mathbb{E}_{\boldsymbol{D} \sim \boldsymbol{\mathcal{D}}} \left(\nabla f(\boldsymbol{w}_{k_{\boldsymbol{w}}}, \boldsymbol{x}) - \nabla f(\boldsymbol{w}, \boldsymbol{x})\|_2\right)}{\|\boldsymbol{w} - \boldsymbol{w}_{k_{\boldsymbol{w}}}\|_2} \sup_{\boldsymbol{w} \in \Omega} \|\boldsymbol{w} - \boldsymbol{w}_{k_{\boldsymbol{w}}}\|_2 \geq \frac{t}{3}\right) \\
\leq& \mathbb{P}\left(\epsilon \mathbb{E}_{\boldsymbol{D} \sim \boldsymbol{\mathcal{D}}} \sup_{\boldsymbol{w} \in \Omega} \left\|\nabla^2 \widetilde{\boldsymbol{Q}}_n(\boldsymbol{w}, \boldsymbol{x})\right\|_2 \geq \frac{t}{3}\right) \\
\leq& \mathbb{P}\left(\gamma \epsilon \geq \frac{t}{3}\right).
\end{aligned}
$$

We set $\epsilon$ enough small such that $\gamma\epsilon < t/3$ always holds. Then it yields $\mathbb{P}(\boldsymbol{E}_3) = 0$.

**Step 4. Final result**: Note that $\frac{6\beta\epsilon}{\varepsilon} \geq 3\beta\epsilon$. Finally, to ensure $\mathbb{P}(\boldsymbol{E}_0) \leq \varepsilon$, we just set $\epsilon = \frac{18p^2(b_{l+1} + \sum_{i=1}^l d_i b_i)}{\vartheta^2 n b_{l+1}} \left[\prod_{s=i}^l \frac{d_s {b_s}^2 (k_s - s_s + 1)^2}{16p^2}\right]^{-\frac{1}{2}}$.

$$
t \geq \max\left(\frac{6\gamma\epsilon}{\varepsilon}, \sqrt{\frac{72\beta^2 \left(d \log(6) + \theta \log\left(\frac{9(b_{l+1} + \sum_{i=1}^l d_i b_i)}{\epsilon}\right) + \log\left(\frac{4}{\varepsilon}\right)\right)}{n}}\right).
$$

By comparing the values of $\beta$ and $\gamma$, we have if $n \geq c_{g'} \frac{l^2 {b_{l+1}}^2 (b_{l+1}+\sum_{i=1}^{l} d_i b_i)^2 (r_0 c_0 d_0)^4}{d_0^4 {b_1}^8 (d\log(6)+\theta\varrho)\varepsilon^2 \max_i (r_i c_i)}$ where $c_{g'}$ is a universal constant, then there exists a universal constant $c_g$ such that

$$\sup_{\boldsymbol{w}\in\Omega} \left\|\nabla_{\boldsymbol{w}} \widetilde{\boldsymbol{Q}}_n(\boldsymbol{w}) - \nabla_{\boldsymbol{w}} \boldsymbol{Q}(\boldsymbol{w})\right\|_2 \leq c_g \beta \sqrt{\frac{d + \frac{1}{\log(6)}\theta \left[\left(\sum_{i=1}^{l} \log\left(\frac{\sqrt{d_s} b_s (k_s - s_s + 1)}{4p}\right) + \log(b_{l+1}) + \log\left(\frac{n}{128p^2}\right)\right) + \log\left(\frac{4}{\varepsilon}\right)\right]}{n}}$$
$$\leq c_g \beta \sqrt{\frac{d + \frac{1}{2}\theta\varrho + \frac{1}{2}\log\left(\frac{4}{\varepsilon}\right)}{n}},$$

holds with probability at least $1-\varepsilon$, where $d = \tilde{r}_l \tilde{c}_l d_l d_{l+1} + \sum_{i=1}^{l} {k_i}^2 d_{i-1} d_i$, $\theta = a_{l+1}(d_{l+1} + \tilde{r}_l \tilde{c}_l d_l + 1) + \sum_{i=1}^{l} a_i ({k_i}^2 d_{i-1} + d_i + 1)$, $\varrho = \sum_{i=1}^{l} \log\left(\frac{\sqrt{d_i} b_i (k_i - s_i + 1)}{4p}\right) + \log(b_{l+1}) + \log\left(\frac{n}{128p^2}\right)$, and $\beta \triangleq \left[\vartheta \tilde{r}_l \tilde{c}_l d_l + \sum_{i=1}^{l} \frac{\vartheta {b_{l+1}}^2 d_{i-1}}{p^2 {b_i}^2 d_i} r_{i-1} c_{i-1} \prod_{s=i}^{l} \frac{d_s {b_s}^2 (k_s - s_s + 1)^2}{16p^2}\right]^{1/2}$ in which $\vartheta = 1/8$. The proof is completed. □

## D.4 Proof of Corollary 1

*Proof.* By Theorem 2, we know that there exist universal constants $c_{g'}$ and $c_g$ such that if $n \geq c_{g'} \frac{l^2 {b_{l+1}}^2 (b_{l+1}+\sum_{i=1}^{l} d_i b_i)^2 (r_0 c_0 d_0)^4}{d_0^4 {b_1}^8 (d\log(6)+\theta\varrho)\varepsilon^2 \max_i (r_i c_i)}$, then

$$\sup_{\boldsymbol{w}\in\Omega} \left\|\nabla_{\boldsymbol{w}} \widetilde{\boldsymbol{Q}}_n(\boldsymbol{w}) - \nabla_{\boldsymbol{w}} \boldsymbol{Q}(\boldsymbol{w})\right\|_2 \leq c_g \beta \sqrt{\frac{2d + \theta\varrho + \log\left(\frac{4}{\varepsilon}\right)}{2n}}$$

holds with probability at least $1-\varepsilon$, where $\varrho$ is provided in Lemma 1. Here $\beta$ and $d$ are defined as $\beta = \left[\frac{r_l c_l d_l}{8p^2} + \sum_{i=1}^{l} \frac{{b_{l+1}}^2 d_{i-1}}{8p^2 {b_i}^2 d_i} r_{i-1} c_{i-1} \prod_{j=i}^{l} \frac{d_j {b_j}^2 (k_j - s_j + 1)^2}{16p^2}\right]^{1/2}$ and $d = \tilde{r}_l \tilde{c}_l d_l d_{l+1} + \sum_{i=1}^{l} {k_i}^2 d_{i-1} d_i$, respectively.

So based on such a result, we can derive that if $n \geq c_g^2 (2d + \theta\varrho + \log(4/\varepsilon))\beta^2/(2\epsilon)$, then we have

$$\|\nabla \boldsymbol{Q}(\widetilde{\boldsymbol{w}})\|_2 \leq \left\|\nabla_{\boldsymbol{w}} \widetilde{\boldsymbol{Q}}_n(\widetilde{\boldsymbol{w}})\right\|_2 + \left\|\nabla_{\boldsymbol{w}} \widetilde{\boldsymbol{Q}}_n(\widetilde{\boldsymbol{w}}) - \nabla_{\boldsymbol{w}} \boldsymbol{Q}(\widetilde{\boldsymbol{w}})\right\|_2 \leq \sqrt{\epsilon} + c_g \beta \sqrt{\frac{2d + \theta\varrho + \log\left(\frac{4}{\varepsilon}\right)}{2n}} \leq 2\sqrt{\epsilon}.$$

Thus, we have $\|\nabla \boldsymbol{Q}(\widetilde{\boldsymbol{w}})\|_2^2 \leq 4\epsilon$, which means that $\widetilde{\boldsymbol{w}}$ is a $4\epsilon$-approximate stationary point in population risk with probability at least $1-\varepsilon$. The proof is completed. □

## D.5 Proof of Theorem 3

*Proof.* Suppose that $\{\boldsymbol{w}_{(1)}, \boldsymbol{w}_{(2)}, \cdots, \boldsymbol{w}_{(m)}\}$ are the non-degenerate critical points of $\boldsymbol{Q}(\boldsymbol{w})$. So for any $\boldsymbol{w}_{(k)}$, it obeys

$$\inf_i \left|\lambda_i^k \left(\nabla^2 \boldsymbol{Q}(\boldsymbol{w}_{(k)})\right)\right| \geq \zeta,$$

where $\lambda_i^k \left(\nabla^2 \boldsymbol{Q}(\boldsymbol{w}_{(k)})\right)$ denotes the $i$-th eigenvalue of the Hessian $\nabla^2 \boldsymbol{Q}(\boldsymbol{w}_{(k)})$ and $\zeta$ is a constant. We further define a set $D = \{\boldsymbol{w} \in \mathbb{R}^d \,|\, \|\nabla \boldsymbol{Q}(\boldsymbol{w})\|_2 \leq \epsilon \text{ and } \inf_i |\lambda_i \left(\nabla^2 \boldsymbol{Q}(\boldsymbol{w}_{(k)})\right)| \geq \zeta\}$. According to Lemma 5, $D = \cup_{k=1}^{\infty} D_k$ where each $D_k$ is a disjoint component with $\boldsymbol{w}_{(k)} \in D_k$ for $k \leq m$ and $D_k$ does not contain any critical point of $\boldsymbol{Q}(\boldsymbol{w})$ for $k \geq m+1$. On the other hand, by the continuity of $\nabla \boldsymbol{Q}(\boldsymbol{w})$, it yields $\|\nabla \boldsymbol{Q}(\boldsymbol{w})\|_2 = \epsilon$ for $\boldsymbol{w} \in \partial D_k$. Notice, we set the value of $\epsilon$ blow which is actually a function related to $n$.

Then by utilizing Theorem 2, we let sample number $n$ sufficient large such that

$$\sup_{\boldsymbol{w}\in\Omega} \left\|\nabla \widetilde{\boldsymbol{Q}}_n(\boldsymbol{w}) - \nabla \boldsymbol{Q}(\boldsymbol{w})\right\|_2 \leq \frac{\epsilon}{2}$$

holds with probability at least $1-\varepsilon$, where $\epsilon$ is defined as

$$\frac{\epsilon}{2} \triangleq c_g \beta \sqrt{\frac{d\log(6) + \theta\left(\sum_{i=1}^{l} \log\left(\frac{\sqrt{d_s} b_s (k_s - s_s + 1)}{4p}\right) + \log(b_{l+1}) + \log\left(\frac{n}{128p^2}\right)\right) + \log\left(\frac{4}{\varepsilon}\right)}{n}}.$$

This further gives that for arbitrary $\boldsymbol{w} \in D_k$, we have

$$\begin{aligned}\inf_{\boldsymbol{w}\in D_k} \left\|t\nabla \widetilde{\boldsymbol{Q}}_n(\boldsymbol{w}) + (1-t)\nabla \boldsymbol{Q}(\boldsymbol{w})\right\|_2 =& \inf_{\boldsymbol{w}\in D_k} \left\|t\left(\nabla \widetilde{\boldsymbol{Q}}_n(\boldsymbol{w}) - \nabla \boldsymbol{Q}(\boldsymbol{w})\right) + \nabla \boldsymbol{Q}(\boldsymbol{w})\right\|_2 \\ \geq& \inf_{\boldsymbol{w}\in D_k} \|\nabla \boldsymbol{Q}(\boldsymbol{w})\|_2 - \sup_{\boldsymbol{w}\in D_k} t\left\|\nabla \widetilde{\boldsymbol{Q}}_n(\boldsymbol{w}) - \nabla \boldsymbol{Q}(\boldsymbol{w})\right\|_2 \\ \geq& \frac{\epsilon}{2}. \end{aligned} \tag{6}$$

Similarly, by utilizing Lemma 18, let $n$ be sufficient large such that

$$\sup_{\boldsymbol{w}\in\Omega} \left\|\nabla^2 \widetilde{\boldsymbol{Q}}_n(\boldsymbol{w}) - \nabla^2 \boldsymbol{Q}(\boldsymbol{w})\right\|_{\text{op}} \leq \frac{\zeta}{2}$$

holds with probability at least $1-\varepsilon$, where $\zeta$ satisfies

$$\frac{\zeta}{2} \geq c_v \gamma \sqrt{\frac{d + \theta\varrho + \log\left(\frac{4}{\varepsilon}\right)}{n}}.$$

Assume that $\boldsymbol{b} \in \mathbb{R}^d$ is a vector and satisfies $\boldsymbol{b}^T\boldsymbol{b} = 1$. In this case, we can bound $\lambda_i^k\left(\nabla^2 \widetilde{\boldsymbol{Q}}_n(\boldsymbol{w})\right)$ for arbitrary $\boldsymbol{w} \in D_k$ as follows:

$$\begin{aligned}\inf_{\boldsymbol{w}\in D_k} \left|\lambda_i^k\left(\nabla^2 \widetilde{\boldsymbol{Q}}_n(\boldsymbol{w})\right)\right| =& \inf_{\boldsymbol{w}\in D_k} \min_{\boldsymbol{b}^T\boldsymbol{b}=1} \left|\boldsymbol{b}^T \nabla^2 \widetilde{\boldsymbol{Q}}_n(\boldsymbol{w}) \boldsymbol{b}\right| \\ =& \inf_{\boldsymbol{w}\in D_k} \min_{\boldsymbol{b}^T\boldsymbol{b}=1} \left|\boldsymbol{b}^T\left(\nabla^2 \widetilde{\boldsymbol{Q}}_n(\boldsymbol{w}) - \nabla^2 \boldsymbol{Q}(\boldsymbol{w})\right)\boldsymbol{b} + \boldsymbol{b}^T \nabla^2 \boldsymbol{Q}(\boldsymbol{w})\boldsymbol{b}\right| \\ \geq& \inf_{\boldsymbol{w}\in D_k} \min_{\boldsymbol{b}^T\boldsymbol{b}=1} \left|\boldsymbol{b}^T \nabla^2 \boldsymbol{Q}(\boldsymbol{w})\boldsymbol{b}\right| - \min_{\boldsymbol{b}^T\boldsymbol{b}=1}\left|\boldsymbol{b}^T\left(\nabla^2 \widetilde{\boldsymbol{Q}}_n(\boldsymbol{w}) - \nabla^2 \boldsymbol{Q}(\boldsymbol{w})\right)\boldsymbol{b}\right| \\ \geq& \inf_{\boldsymbol{w}\in D_k} \min_{\boldsymbol{b}^T\boldsymbol{b}=1} \left|\boldsymbol{b}^T \nabla^2 \boldsymbol{Q}(\boldsymbol{w})\boldsymbol{b}\right| - \max_{\boldsymbol{b}^T\boldsymbol{b}=1}\left|\boldsymbol{b}^T\left(\nabla^2 \widetilde{\boldsymbol{Q}}_n(\boldsymbol{w}) - \nabla^2 \boldsymbol{Q}(\boldsymbol{w})\right)\boldsymbol{b}\right| \\ =& \inf_{\boldsymbol{w}\in D_k} \inf_i \left|\lambda_i^k\left(\nabla^2 f(\boldsymbol{w}_{(k)}, \boldsymbol{x})\right)\right| - \left\|\nabla^2 \widetilde{\boldsymbol{Q}}_n(\boldsymbol{w}) - \nabla^2 \boldsymbol{Q}(\boldsymbol{w})\right\|_{\text{op}} \\ \geq& \frac{\zeta}{2}. \end{aligned} \tag{7}$$

This means that in each set $D_k$, $\nabla^2 \widetilde{\boldsymbol{Q}}_n(\boldsymbol{w})$ has no zero eigenvalues. Then, combine this and Eqn. (6), by Lemma 4 we know that if the population risk $\boldsymbol{Q}(\boldsymbol{w})$ has no critical point in $D_k$, then the empirical risk $\widetilde{\boldsymbol{Q}}_n(\boldsymbol{w})$ has also no critical point in $D_k$; otherwise it also holds.

Now we bound the distance between the corresponding critical points of $\boldsymbol{Q}(\boldsymbol{w})$ and $\widetilde{\boldsymbol{Q}}_n(\boldsymbol{w})$. Assume that in $D_k$, $\boldsymbol{Q}(\boldsymbol{w})$ has a unique critical point $\boldsymbol{w}_{(k)}$ and $\widetilde{\boldsymbol{Q}}_n(\boldsymbol{w})$ also has a unique critical point $\boldsymbol{w}_n^{(k)}$. Then, there exists $t \in [0,1]$ such that for any $\boldsymbol{z} \in \partial\mathsf{B}^d(1)$, we have

$$\begin{aligned}\epsilon \geq& \|\nabla \boldsymbol{Q}(\boldsymbol{w}_n^{(k)})\|_2 \\ =& \max_{\boldsymbol{z}^T\boldsymbol{z}=1} \langle \nabla \boldsymbol{Q}(\boldsymbol{w}_n^{(k)}), \boldsymbol{z}\rangle \\ =& \max_{\boldsymbol{z}^T\boldsymbol{z}=1} \langle \nabla \boldsymbol{Q}(\boldsymbol{w}_{(k)}), \boldsymbol{z}\rangle + \langle \nabla^2 \boldsymbol{Q}(\boldsymbol{w}_{(k)} + t(\boldsymbol{w}_n^{(k)} - \boldsymbol{w}_{(k)}))(\boldsymbol{w}_n^{(k)} - \boldsymbol{w}_{(k)}), \boldsymbol{z}\rangle \\ \overset{①}{\geq}& \left\langle \left(\nabla^2 \boldsymbol{Q}(\boldsymbol{w}_{(k)})\right)^2 (\boldsymbol{w}_n^{(k)} - \boldsymbol{w}_{(k)}), (\boldsymbol{w}_n^{(k)} - \boldsymbol{w}_{(k)})\right\rangle^{1/2} \\ \overset{②}{\geq}& \zeta\|\boldsymbol{w}_n^{(k)} - \boldsymbol{w}_{(k)}\|_2, \end{aligned}$$

where ① holds since $\nabla \boldsymbol{Q}(\boldsymbol{w}_{(k)}) = \boldsymbol{0}$ and ② holds since $\boldsymbol{w}_{(k)} + t(\boldsymbol{w}_n^{(k)} - \boldsymbol{w}_{(k)})$ is in $D_k$ and for any $\boldsymbol{w} \in D_k$ we have $\inf_i \left|\lambda_i\left(\nabla^2 \boldsymbol{Q}(\boldsymbol{w})\right)\right| \geq \zeta$. So if $n \geq c_h \max\left(\frac{l^2 {b_{l+1}}^2 (b_{l+1} + \sum_{i=1}^l d_i b_i)^2 (r_0 c_0 d_0)^4}{d_0^4 {b_1}^8 d\varrho\varepsilon^2 \max_i(r_i c_i)}, \frac{d+\theta\varrho}{\zeta^2}\right)$ where $c_h$ is a constant, then

$$\|\boldsymbol{w}_n^{(k)} - \boldsymbol{w}_{(k)}\|_2 \leq \frac{2c_g\beta}{\zeta}\sqrt{\frac{d\log(6) + \theta\left(\sum_{i=1}^l \log\left(\frac{\sqrt{d_s} b_s (k_s - s_s + 1)}{4p}\right) + \log(b_{l+1}) + \log\left(\frac{n}{128p^2}\right)\right) + \log\left(\frac{4}{\varepsilon}\right)}{n}}$$

holds with probability at least $1-\varepsilon$. □

### D.6 Proof of Corollary 2

*Proof.* By Theorem 3, we know that the non-degenerate stationary point $\boldsymbol{w}_{(k)}$ in the $m$ non-degenerate stationary points in population risk, denoted by $\{\boldsymbol{w}_{(1)}, \boldsymbol{w}_{(2)}, \cdots, \boldsymbol{w}_{(m)}\}$ uniquely corresponding to a non-degenerate stationary point $\boldsymbol{w}_n^{(k)}$ in the empirical risk.

On the other hand, for any $\boldsymbol{w}_{(k)}$, it obeys

$$\inf_i \left|\lambda_i^k\left(\nabla^2 \boldsymbol{Q}(\boldsymbol{w}_{(k)})\right)\right| \geq \zeta,$$

where $\lambda_i^k\left(\nabla^2 \boldsymbol{Q}(\boldsymbol{w}_{(k)})\right)$ denotes the $i$-th eigenvalue of the Hessian $\nabla^2 \boldsymbol{Q}(\boldsymbol{w}_{(k)})$ and $\zeta$ is a constant. We further define a set $D=\{\boldsymbol{w}\in\mathbb{R}^d \mid \|\nabla \boldsymbol{Q}(\boldsymbol{w})\|_2 \leq \epsilon \text{ and } \inf_i|\lambda_i\left(\nabla^2 \boldsymbol{Q}(\boldsymbol{w}_{(k)})\right)| \geq \zeta\}$. According to Lemma 5, $D=\cup_{k=1}^{\infty} D_k$ where each $D_k$ is a disjoint component with $\boldsymbol{w}_{(k)}\in D_k$ for $k\leq m$ and $D_k$ does not contain any critical point of $\boldsymbol{Q}(\boldsymbol{w})$ for $k\geq m+1$. Then $\boldsymbol{w}_n^{(k)}$ also belong to the component $D_k$ due to the unique corresponding relation between $\boldsymbol{w}_{(k)}$ and $\boldsymbol{w}_n^{(k)}$. Then from Eqn. (6) and (7), we know that if the assumptions in Theorem 3 hold, then for arbitrary $\boldsymbol{w}\in D_k$ and $t\in(0,1)$,

$$\inf_{\boldsymbol{w}\in D_k}\left\|t\nabla\widetilde{\boldsymbol{Q}}_n(\boldsymbol{w})+(1-t)\nabla \boldsymbol{Q}(\boldsymbol{w})\right\|_2 \geq \frac{\epsilon}{2} \quad \text{and} \quad \inf_{\boldsymbol{w}\in D_k}\left|\lambda_i^k\left(\nabla^2\widetilde{\boldsymbol{Q}}_n(\boldsymbol{w})\right)\right| \geq \frac{\zeta}{2},$$

where $\epsilon$ and $\zeta$ are constants. This means that in each set $D_k$, $\nabla^2\widetilde{\boldsymbol{Q}}_n(\boldsymbol{w})$ has no zero eigenvalues. Then, combine this and Eqn. (6), we can obtain that in $D_k$, if $\boldsymbol{Q}(\boldsymbol{w})$ has a unique critical point $\boldsymbol{w}_{(k)}$ with non-degenerate index $s_k$, then $\widetilde{\boldsymbol{Q}}_n(\boldsymbol{w})$ also has a unique critical point $\boldsymbol{w}_{(k)}^n$ in $D_k$ with the same non-degenerate index $s_k$. Namely, the number of negative eigenvalues of the Hessian matrices $\nabla^2 \boldsymbol{Q}(\boldsymbol{w}_{(k)})$ and $\nabla^2 \boldsymbol{Q}(\boldsymbol{w}_n^{(k)})$ are the same. This further gives that if one of the pair $(\boldsymbol{w}_{(k)}, \boldsymbol{w}_n^{(k)})$ is a local minimum or saddle point, then another one is also a local minimum or a saddle point. The proof is completed. □

## E Proof of Auxiliary Lemmas

### E.1 Proof of Lemma 8

*Proof.* (1) Since $\mathsf{G}(\boldsymbol{z})$ is a diagonal matrix and its diagonal values are upper bounded by $\sigma_1(\boldsymbol{z}_i)(1-\sigma_1(\boldsymbol{z}))\leq 1/4$ where $\boldsymbol{z}_i$ denotes the $i$-th entry of $\boldsymbol{z}_i$, we can conclude

$$\|\mathsf{G}(\boldsymbol{z})\boldsymbol{M}\|_F^2 \leq \frac{1}{16}\|\boldsymbol{M}\|_F^2 \quad \text{and} \quad \|\boldsymbol{N}\mathsf{G}(\boldsymbol{z})\|_F^2 \leq \frac{1}{16}\|\boldsymbol{N}\|_F^2.$$

(2) The operator $\mathsf{Q}(\cdot)$ maps a vector $\boldsymbol{z}\in\mathbb{R}^d$ into a matrix of size $d^2\times d$ whose $((i-1)d+i, i)$ $(i=1,\cdots,d)$ entry equal to $\sigma_1(\boldsymbol{z}_i)(1-\sigma_1(\boldsymbol{z}_i))(1-2\sigma_1(\boldsymbol{z}_i))$ and rest entries are all 0. This gives

$$\begin{aligned}\sigma_1(\boldsymbol{z}_i)(1-\sigma_1(\boldsymbol{z}_i))(1-2\sigma_1(\boldsymbol{z}_i)) =& \frac{1}{3}(3\sigma_1(\boldsymbol{z}_i))(1-\sigma_1(\boldsymbol{z}_i))(1-2\sigma_1(\boldsymbol{z}_i))\\ \leq& \frac{1}{3}\left(\frac{3\sigma_1(\boldsymbol{z}_i)+1-\sigma_1(\boldsymbol{z}_i)+1-2\sigma_1(\boldsymbol{z}_i)}{3}\right)^3\\ \leq& \frac{2^3}{3^4}.\end{aligned}$$

This means the maximal value in $\mathsf{Q}(\boldsymbol{z})$ is at most $\frac{2^3}{3^4}$. Consider the structure in $\mathsf{Q}(\boldsymbol{z})$, we can obtain

$$\|\mathsf{Q}(\boldsymbol{z})\boldsymbol{M}\|_F^2 \leq \frac{2^6}{3^8}\|\boldsymbol{M}\|_F^2 \quad \text{and} \quad \|\boldsymbol{N}\mathsf{Q}(\boldsymbol{z})\|_F^2 \leq \frac{2^6}{3^8}\|\boldsymbol{N}\|_F^2.$$

(3) $\mathsf{up}(\boldsymbol{M})$ represents conducting upsampling on $\boldsymbol{M}\in\mathbb{R}^{s\times t\times q}$. Let $\boldsymbol{N}=\mathsf{up}(\boldsymbol{M})\in\mathbb{R}^{ps\times pt\times q}$. Specifically, for each slice $\boldsymbol{N}(:,:,i)$ $(i=1,\cdots,q)$, we have $\boldsymbol{N}(:,:,i)=\mathsf{up}(\boldsymbol{M}(:,:,i))$. It actually upsamples each entry $\boldsymbol{M}(g,h,i)$ into a matrix of $p^2$ same entries $\frac{1}{p^2}\boldsymbol{M}(g,h,i)$. So it is easy to obtain

$$\|\mathsf{up}(\boldsymbol{M})\|_F^2 \leq \frac{1}{p^2}\|\boldsymbol{M}\|_F^2.$$

(4) Let $\boldsymbol{M} = \boldsymbol{W}(:,:,i)$ and $\boldsymbol{N} = \widetilde{\boldsymbol{\delta}}_{i+1}(:,:i)$. Assume that $\boldsymbol{H} = \boldsymbol{M} \circledast \boldsymbol{N} \in \mathbb{R}^{m_1 \times m_2}$, where $m_1 = \tilde{r}_{i-1} - 2k_i + 2$ and $m_2 = \tilde{c}_{i-1} - 2k_i + 2$. Then we have

$$\|\boldsymbol{H}\|_F^2 = \sum_{i=1}^{m_1}\sum_{j=1}^{m_2} |\boldsymbol{H}(i,j)|^2 = \sum_{i=1}^{m_1}\sum_{j=1}^{m_2} \langle \boldsymbol{M}_{\Omega_{i,j}}, \boldsymbol{N}\rangle^2 \leq \sum_{i=1}^{m_1}\sum_{j=1}^{m_2} \|\boldsymbol{M}_{\Omega_{i,j}}\|_F^2 \|\boldsymbol{N}\|_F^2,$$

where $\Omega_{i,j}$ denotes the entry index of $\boldsymbol{M}$ for the $(i,j)$-th convolution operation (*i.e.* computing the $\boldsymbol{H}(i,j)$).

Since for each convolution computing, each element in $\boldsymbol{M}$ is involved at most one time, we can claim that any element in $\boldsymbol{M}$ in $\sum_{i=1}^{m_1}\sum_{j=1}^{m_2} \|\boldsymbol{M}_{\Omega_{i,j}}\|_F^2$ occurs at most $(k_i - s_i + 1)^2$ since there are $s_i - 1$ rows and columns between each neighboring nonzero entries in $\boldsymbol{N}$ which is decided by the definition of $\widetilde{\boldsymbol{\delta}}_{i+1}$ in Sec. B.1. Therefore, we have

$$\sum_{i=1}^{m_1}\sum_{j=1}^{m_2} \|\boldsymbol{M}_{\Omega_{i,j}}\|_F^2 \leq (k_i - s_i + 1)^2 \|\boldsymbol{M}\|_F^2,$$

which further gives

$$\|\boldsymbol{M} \widetilde{\circledast} \boldsymbol{N}\|_F^2 \leq (k_i - s_i + 1)^2 \|\boldsymbol{M}\|_F^2 \|\boldsymbol{N}\|_F^2.$$

Consider all the slices in $\widetilde{\boldsymbol{\delta}}_{i+1}$, we can obtain

$$\|\widetilde{\boldsymbol{\delta}}_{i+1} \widetilde{\circledast} \boldsymbol{W}\|_F^2 \leq (k_i - s_i + 1)^2 \|\boldsymbol{W}\|_F^2 \|\widetilde{\boldsymbol{\delta}}_{i+1}\|_F^2.$$

(5) Since for softmax activation function $\sigma_2$, we have $\sum_{i=1}^{d_{l+1}} \boldsymbol{v}_i = 1$ $(\boldsymbol{v}_i \geq 0)$ and there is only one nonzero entry (*i.e.* 1) in $\boldsymbol{y}$, we can obtain

$$0 \leq \|\boldsymbol{v} - \boldsymbol{y}\|_2^2 = \|\boldsymbol{v}\|_2^2 + \|\boldsymbol{y}\|_2^2 - 2\langle \boldsymbol{v}, \boldsymbol{y}\rangle = 2 - 2\langle \boldsymbol{v}, \boldsymbol{y}\rangle \leq 2.$$

The proof is completed. □